\documentclass[a4paper,fleqn]{cas-sc}

\usepackage[authoryear]{natbib}
\usepackage{flafter}
\usepackage{capt-of} %
\usepackage{array}  
\usepackage{placeins} 
\usepackage{nicefrac}
\usepackage{lipsum}  
\usepackage{graphicx}
\usepackage{subcaption}
\usepackage{float}
\usepackage{placeins}
\usepackage{lipsum}
\usepackage{soul,color}
\usepackage{pdflscape}
\usepackage{rotating}
\usepackage{makecell}
\usepackage{placeins}
\usepackage{float}
\usepackage{booktabs}
\usepackage{xcolor}
\usepackage{caption}
\usepackage{graphicx} 
\usepackage{subcaption} 
\usepackage{lscape}
\usepackage{pdflscape}
\usepackage{pifont} 

\usepackage{amsmath}
\usepackage{amsfonts}
\usepackage{amssymb}

\usepackage{bbm}
\usepackage{ltablex}        
\keepXColumns              
\usepackage{array}         
\usepackage{tikz}
\usepackage{titlesec}
\usepackage{listings}
\usepackage{makecell}
\usepackage{tabularx,ragged2e}
\newcolumntype{C}{>{\arraybackslash}X} 

\usepackage{xcolor}

\definecolor{codegreen}{rgb}{0,0.6,0}
\definecolor{codegray}{rgb}{0.5,0.5,0.5}
\definecolor{codepurple}{rgb}{0.58,0,0.82}
\definecolor{backcolour}{rgb}{0.95,0.95,0.92}

\lstdefinestyle{mystyle}{
    backgroundcolor=\color{backcolour},   
    commentstyle=\color{codegreen},
    keywordstyle=\color{magenta},
    numberstyle=\tiny\color{codegray},
    stringstyle=\color{codepurple},
    basicstyle=\ttfamily\footnotesize,
    breakatwhitespace=false,         
    breaklines=true,                 
    captionpos=b,                    
    keepspaces=true,                 
    numbers=left,                    
    numbersep=5pt,                  
    showspaces=false,                
    showstringspaces=false,
    showtabs=false,                  
    tabsize=2
}

\usetikzlibrary{shapes.geometric, arrows.meta, positioning}
\definecolor{myLightBlue}{HTML}{DDEBF7}
\definecolor{myRed}{HTML}{FFC7CE}
\definecolor{myGreen}{HTML}{C6E0B4}

\tikzstyle{startstop} = [rectangle, rounded corners, minimum width=2.7cm, minimum height=1.2cm, text centered, draw=black, fill=myLightBlue]
\tikzstyle{process} = [rectangle, minimum width=2.7cm, minimum height=1.2cm, text centered, draw=black, fill=orange!30]
\tikzstyle{subprocess} = [rectangle, minimum width=2.7cm, minimum height=1.2cm, text centered, draw=black, fill=green!30]
\tikzstyle{arrow} = [thick,-{Stealth[scale=1.2]},>=stealth]

\titleformat*{\section}{\large\bfseries}
\titleformat{\subsubsection}{\normalfont\normalsize\bfseries}{\thesubsubsection}{1em}{\normalfont\normalsize\bfseries}

\titleformat{\paragraph}
{\normalfont\normalsize\bfseries}{\theparagraph}{1em}{}
\titlespacing*{\paragraph}
{0pt}{3.25ex plus 1ex minus .2ex}{1.5ex plus .2ex}

\numberwithin{equation}{section}
\newcommand{\cmark}{\ding{51}}%
\newcommand{\xmark}{\ding{55}}%
\DeclareMathOperator{\diag}{diag}
\usepackage[pagewise]{lineno}
\newcommand{\norm}[1]{\left\lVert#1\right\rVert}

\def\tsc#1{\csdef{#1}{\textsc{\lowercase{#1}}\xspace}}
\tsc{WGM}
\tsc{QE}
\tsc{EP}
\tsc{PMS}
\tsc{BEC}
\tsc{DE}

\newcount\Comments  
\usepackage{color}
\definecolor{darkgreen}{rgb}{0,0.5,0}
\definecolor{purple}{rgb}{1,0,1}
\newcommand{\kibitz}[2]{\ifnum\Comments=0\textcolor{#1}{#2}\fi}

\begin{document}
\let\WriteBookmarks\relax
\def\floatpagepagefraction{1}
\def\textpagefraction{.001}
\shorttitle{A Gentle Introduction and Tutorial on Deep Generative Models in Transportation Research}
\shortauthors{S. Choi, Z. Jin, S. Ham, J. Kim, and L. Sun}

\title [mode = title]{A Gentle Introduction and Tutorial on Deep Generative Models in Transportation Research}



\author[1]{Seongjin Choi}[orcid=0000-0001-7140-537X]
\ead{chois@umn.edu}
\cormark[1]
\credit{Conceptualization of this study, Data Curation, Investigation, Methodology, Software, Validation, Formal analysis, Supervision, Project administration, Writing - original draft, Writing - review and editing}

\author[2,3]{Zhixiong Jin}[orcid=0000-0002-1370-781X]
\ead{zhixiong.jin@univ-eiffel.fr}
\cormark[1]
\credit{Conceptualization of this study, Investigation, Methodology, Software, Validation, Formal analysis,  Visualization, Writing - original draft, Writing - review and editing}

\author[4]{Seung Woo Ham}[orcid=0000-0001-7531-9217]
\ead{sw.ham@nus.edu.sg}
\credit{Software, Data Curation, Visualization, Writing - original draft, Writing - review and editing}

\author[5]{Jiwon Kim}[orcid=0000-0001-6380-3001]
\ead{jiwon.kim@uq.edu.au}
\credit{Conceptualization of this study, Supervision, Writing - review and editing}

\author[6]{Lijun Sun}[orcid=0000-0001-9488-0712]
\ead{lijun.sun@mcgill.ca}
\credit{Conceptualization of this study, Supervision, Writing - review and editing}

\address[1]{Department of Civil, Environmental, and Geo- Engineering, University of Minnesota, 500 Pillsbury Dr. SE, Minneapolis, MN 55455, USA}
\address[2]{Univ. Guatave Eiffel, ENTPE, LICIT-ECO7, Lyon, France}
\address[3]{Urban Transport Systems Laboratory (LUTS), École Polytechnique Fédérale de Lausanne (EPFL), Lausanne, CH 1015, Switzerland}
\address[4]{Department of Civil and Environmental Engineering, National University of Singapore, Singapore}
\address[5]{School of Civil Engineering, The University of Queensland, Brisbane St Lucia, Queensland, Australia}
\address[6]{Department of Civil Engineering, McGill University, 817 Sherbrooke Street West, Montreal, Quebec H3A 0C3, Canada}

\cortext[cor1]{Equal Contribution}
\cortext[cor2]{This survey is based on literature up to October 2024.}










\begin{abstract}
Deep Generative Models (DGMs) have rapidly advanced in recent years, becoming essential tools in various fields due to their ability to learn data distributions and generate synthetic data. Their importance in transportation research is increasingly recognized, particularly for applications like traffic data generation, prediction, and feature extraction. This paper offers a comprehensive introduction and tutorial on DGMs, with a focus on their applications in transportation. It begins with an overview of generative models, followed by detailed explanations of fundamental models, a systematic review of the literature, and practical tutorial code to aid implementation. The paper also discusses current challenges and opportunities, highlighting how these models can be effectively utilized and further developed in transportation research. This paper serves as a valuable reference, guiding researchers and practitioners from foundational knowledge to advanced applications of DGMs in transportation research.
%
\end{abstract}





\patchcmd{\maketitle}
  {\finalMaketitle}
  {\finalMaketitle\tableofcontents\vspace{\baselineskip}}
  {}{}
  
\maketitle

\setcounter{tocdepth}{2}
\tableofcontents


\clearpage
\section{Introduction}
The rapid growth of artificial intelligence in transportation research over recent years has provided innovative solutions to address both longstanding and emerging transportation issues. Among the spectrum of artificial intelligence techniques, Deep Generative Models (DGMs) are increasingly gaining attention in transportation research due to their ability to learn the underlying patterns in large datasets and model the data distribution. Using the learned distributional characteristics, DGMs can generate synthetic data that accurately replicate real-world scenarios, estimate and predict agent-level trajectories, link-level and network-level traffic states, and provide latent representations of complex transportation systems. 
The primary purpose of this paper is to review the state-of-the-art theories and investigate the applications, potential benefits, and challenges of using DGMs in transportation research. Additionally, we aim to offer a comprehensive tutorial for researchers in transportation engineering who are interested in incorporating DGMs into their works. By providing a detailed overview of the current state-of-the-art models and literature with practical tutorials, this paper seeks to facilitate the adoption of DGMs and inspire further innovations in transportation research.


\subsection{Background}
Conventional deep learning models in transportation research have predominantly utilized discriminative modeling approaches. Discriminative models, as the name suggests, are designed to discriminate or differentiate patterns in data. These models learn a direct mapping from input variables ($\mathbf{x}$) to output variables ($\mathbf{y}$), effectively estimating the conditional probability $p(\mathbf{y}|\mathbf{x})$. In practice, this translates into the model's ability to classify input data into predefined categories or predict target variables based on input features. For example, Convolutional Neural Networks (CNNs) and their variants are widely used to analyze spatial traffic patterns for classifying different traffic conditions; Recurrent Neural Networks (RNNs), including Long Short-Term Memory (LSTM) networks, model temporal dependencies for forecasting traffic variables over time; and Graph Neural Networks (GNNs) capture complex spatial relationships within transportation networks represented as graphs, facilitating tasks such as traffic prediction and network anomaly detection. 


Even though discriminative models have achieved great success in transportation research, inherent limitations constrain their effectiveness. They require a structured form of extensive labeled (paired $\mathbf{x}$ and $\mathbf{y}$ data) datasets and are confined to the patterns present within the training data. Additionally, these models are unable to capture the underlying data distribution $p(\mathbf{x})$, which restricts their ability to understand complex patterns in the dataset fully. Consequently, these limitations hinder their capacity to generalize beyond the observed scenarios in the training dataset, i.e., the predictive ability is limited within what was given in the training dataset. Moreover, transportation data is inherently dynamic and complex, often characterized by uncertainty and rare events. Relying solely on discriminative models can thus impede the development of robust solutions capable of adapting to such complexities and predicting or generating rare events with non-zero probability.


%

Deep Generative Models (DGMs), on the other hand, are a class of machine learning models with the core objective of learning the joint probability $p(\mathbf{x}, \mathbf{y})$ or the underlying data distribution $p(\mathbf{x})$. This capability enables DGMs to handle both discriminative tasks (such as classification and prediction) and generative tasks (such as data generation) within a unified framework. In contrast to traditional discriminative models, which are often limited by their reliance on extensive labeled datasets, DGMs can learn from the inherent patterns in data to generate realistic synthetic examples. In transportation research, DGMs can be used to create realistic traffic scenarios, such as time-space speed contour diagrams, simulate traveler behaviors under different conditions while accounting for the characteristics of individual agents, and generate synthetic datasets to augment rare events. Their ability to capture the complexities of traffic dynamics and traffic agent behavior directly addresses data scarcity issues and enhances the robustness and generalization of predictive models. This versatility makes DGMs an essential tool for managing the dynamic and uncertain nature of transportation systems.


From our viewpoint, the development of effective DGMs represents one of the most promising opportunities in modern transportation research. DGMs have the potential to revolutionize how we simulate, analyze, and optimize transportation systems by learning complex patterns from vast amounts of data without relying on strict assumptions or extensive model-based parameterization. While traditional theory-based models have been invaluable in understanding transportation dynamics, they often face challenges such as oversimplified assumptions and a narrow focus on specific scenarios. Despite these limitations, theory-based models remain essential, as they provide a foundation upon which data-driven methods like DGMs can build. Specifically, instead of replacing traditional approaches, DGMs serve as complementary and supplementary tools, enhancing and expanding upon established models. Their flexibility allows for the modeling of intricate dynamics and the generation of realistic, unseen scenarios, ultimately improving traffic simulations, scenario planning, and the implementation of digital twins in the transportation area.

\begin{table}[b]
    \centering
    \caption{Comparison of previous survey papers}
    \label{tab:table1}
    \begin{tabularx}{\textwidth}
    {l||C|p{1.6cm}|p{1.9cm}|p{1cm}}
\toprule
\toprule
Reference & Contribution & Wide range of DGMs & Transportation Application & Tutorial Code \\
\midrule\midrule
\cite{harshvardhan2020comprehensive} &  provide a comprehensive survey and implementation guide for machine and deep learning-based DGMs. &\cmark & \xmark & \cmark \\ 
\midrule
\cite{bond2021deep} &  provide a comparative review of DGMs, analyzing their strengths, weaknesses, and applications &\cmark & \xmark & \xmark \\ 
\midrule
\cite{de2022deep} & provide a comprehensive review of DGMs in Industrial Internet of Things (IIoT) applications &\cmark & \xmark & \xmark \\ 
\midrule
\cite{guo2022systematic} & provide systematic surveys of DGMS for graph generation with their applications, taxonomy, and analysis.&\cmark & \xmark & \xmark \\ 
\midrule
\cite{lopez2020enhancing} & provide a comprehensive review in the application of DGMs in molecular biology, highlighting their role in advancing scientific discoveries &\cmark & \xmark & \xmark \\ 
\midrule
\cite{anstine2023generative} & provide a comprehensive review in the application of chemical sciences with discussing their application challenges &\cmark & \xmark & \xmark \\
\midrule\midrule
\cite{boquet2020variational} &review the Variational Autoencoders (VAE) models to address key challenges in transportation &\cmark \newline(partially)  & \cmark & \xmark \\
\midrule
\cite{lin2023generative} & provide comprehensive applications of Generative Adversarial Networks (GANs) in Intelligent Transportation Systems (ITSs) &\cmark \newline (partially)  & \cmark & \xmark \\ 
\midrule
\cite{yan2023survey} & provide a comprehensive reviews of DGMs in the ITS applications &\cmark & \cmark & \xmark \\ 
\midrule\midrule
Current Work & provide a comprehensive review and implementation tutorial on DGMs focusing on their applications in transportation research & \cmark & \cmark & \cmark \\
\bottomrule
\bottomrule
\end{tabularx}
\end{table}

\subsection{Relationship with Existing Surveys}
%


DGMs have played a pivotal role in advancing machine learning over the past decade by enabling the generation and approximation of joint distributions for targets and training data. Several notable surveys have highlighted the significance and versatility of DGMs across various fields. For example, \cite{harshvardhan2020comprehensive} provided a comprehensive guide tracing the evolution from traditional machine learning-based generative models to advanced DGMs. Similarly, \cite{bond2021deep} reviewed a wide range of DGMs, including energy-based models, Variational Autoencoders (VAEs), autoregressive models, and normalizing flows, discussing their strengths, weaknesses, and applications. Area-specific surveys further illustrate the adaptability and impact of DGMs. For instance, \cite{de2022deep} explored DGMs' applications in the Industrial Internet of Things (IIoT), while \cite{guo2022systematic} systematically reviewed DGMs for graph generation, focusing on their applications, taxonomy, and analysis. Additionally, \cite{lopez2020enhancing} examined the role of DGMs in molecular biology, highlighting their utility in advancing scientific discoveries. Similarly, \cite{anstine2023generative} provided a comprehensive review in the application of chemical sciences with discussing application challenges.

In transportation research, several review papers have similarly explored the potential of DGMs and highlighted their extensive applications in the field. For instance, \cite{boquet2020variational} discussed the use of VAEs to address key transportation challenges, including traffic imputation, dimensionality reduction, and anomaly detection Similarly, \cite{lin2023generative} categorized the use of Generative Adversarial Networks (GANs) in autonomous driving, traffic flow research, and anomaly detection, while also identifying challenges and future research directions for integrating GANs into transportation research. However, these survey primarily focuses on typical models, with less coverage of other DGMs. \cite{yan2023survey} systematically investigated the role of DGMs, or generative AI, in addressing key issues such as traffic perception, prediction, simulation, and decision-making within Intelligent Transportation Systems (ITS), while discussing existing challenges and future research directions. Even though this survey provides valuable insights, it would benefit from a more comprehensive explanation of fundamental DGMs and a deeper analysis of challenges and opportunities within the transportation context. Such enhancements could make DGMs more accessible and easier to adopt for a broader range of researchers in the field.

To address these gaps, this paper offers a comprehensive introduction and tutorial on DGMs in transportation applications. First, we provide a clear explanation of fundamental DGMs to ensure readers gain a solid understanding of core concepts. This is followed by a systematic review of state-of-the-art DGMs, with particular attention to their application in transportation research. We also include practical tutorial codes to guide researchers and practitioners in effectively applying these models to real-world transportation problems. Lastly, the paper concludes by emphasizing the importance of DGMs, discussing current research challenges, and exploring potential solutions, making it a valuable resource for those interested in exploring the use of DGMs in transportation.





\subsection{Objectives of the Paper}

This paper aims to provide a comprehensive survey of the application and potential of DGMs in the field of transportation research. The objectives of this survey paper are as follows:

\begin{itemize}
\item To provide an accessible and comprehensive introduction to DGMs for transportation researchers. We aim to establish this paper as a starting point for those interested in exploring the potential of DGMs in transportation research.
\item To review the state-of-the-art in transportation research using DGMs, offering insights into the current landscape and practices across various transportation domains.
\item To contribute practical value by offering a tutorial section, providing hands-on guidance and resources for implementing DGMs in transportation research.
\item To identify and discuss the challenges, limitations, and opportunities of employing DGMs in transportation research. 
\end{itemize}

With these objectives, we hope to contribute to the wider acceptance and understanding of DGMs within the field of transportation research. It is our belief that the adoption of these models can revolutionize the way we approach and solve transportation problems, leading to more robust, efficient, and sustainable systems.

\subsection{Structure of Paper}
{

In Section~\ref{sec:intro_to_dgm}, we introduce the capabilities and applications of several key DGMs in the literature, providing a structured overview along with detailed mathematical formulations to enhance understanding. 
}
Specifically, we introduce Variational Autoencoders (VAEs) in Section~\ref{sec:vae}, Generative Adversarial Networks (GANs) in Section~\ref{sec:gan}, Normalizing Flows (or flow-based generative models) in Section~\ref{sec:nf}, 
Score-based generative models in Section~\ref{sec:score} and
Diffusion models in Section~\ref{sec:diff}.
In Section~\ref{sec:dgmreview}, we introduce state-of-the-art transportation research using DGMs.
Particularly, we introduce applications of DGM 1) for generating realistic new data samples that can be applied in data synthesis, trajectory generation, and missing data imputation in Section~\ref{sec:3.1}, 2) for estimating and predicting distributions at three different levels of analyses in transportation research (agent-level, link-level, and region-level) in Section~\ref{sec:3.2}, and 3) for understanding underlying dynamics and learning unsupervised representations of data for applications like anomaly detection and mode choice analysis in Section~\ref{sec:3.3}. 
\textbf{Readers primarily interested in the current practices can begin with Section~\ref{sec:dgmreview} for a review of the latest literature in transportation research using DGMs, and then refer back to Section~\ref{sec:intro_to_dgm} for an introduction to the models themselves.}
In Section~\ref{sec:tutorial}, we provide a tutorial that offers practical guidance on implementing DGMs in transportation research. We introduce two examples: 1) generating travel survey data in Section~\ref{sec:tutorial_survey}, and 2) generating highway traffic speed contour data in Section~\ref{sec:tutorial_speed}. 
In Section~\ref{sec:5}, we identify and discuss the challenges and opportunities associated with using DGMs in the transportation domain, emphasizing the importance of addressing these challenges for the successful adoption of DGMs. Finally, in Section~\ref{sec:conclusion}, we summarize and conclude the paper.

\section{Introduction to Deep Generative Models}\label{sec:intro_to_dgm}
\subsection{Overview of Deep Generative Models}

\subsubsection{Background}
Generative models are a class of machine learning models that aim to understand and capture the underlying probability distribution of a dataset. By learning this distribution, these models can generate new data points that are similar to those in the original dataset. 
Unlike discriminative models, which focus on modeling the conditional probability of the output given the input, $p(\mathbf{y}|\mathbf{x})$, generative models aim to learn the joint distribution $p(\mathbf{x}, \mathbf{y})$ or $p(\mathbf{x})$. By understanding the joint distribution, generative models can also derive the conditional probability $p(\mathbf{y}|\mathbf{x})$ as $p(\mathbf{y}|\mathbf{x}) = \frac{p(\mathbf{x},\mathbf{y})}{p(\mathbf{x})}$. This means generative models can perform discriminative tasks effectively while also being capable of generating new data that aligns with the learned distribution. Therefore, generative models provide a broader framework that not only includes the functionalities of discriminative models but also extends beyond them to include data generation capabilities and model data generation process. Examples of earlier generative models include Gaussian Mixture Models, Hidden Markov Model, and  Naive Bayes Classifier. However, they often rely on simplifying assumptions and may struggle with high-dimensional or complex data patterns.

Deep Generative Models (DGMs) build upon these concepts by leveraging deep learning architectures to model more complex, high-dimensional data without relying on such strong assumptions. By utilizing neural networks, DGMs can learn these complex, nonlinear relationships within data, allowing for more accurate modeling and generation of realistic data samples.

A key advantage of DGMs is their ability to generate data, which is essential in applications such as data augmentation, synthetic data creation, and data privacy enhancement. This capability allows the creation of new, synthetic datasets that can enhance the robustness and performance of machine-learning models, especially when real-world data is limited or sensitive.
Another significant benefit of using DGMs is their capacity for probabilistic inference. It involves using the probability distributions learned by DGMs to do predictions and estimate uncertainties. The probabilistic inference is vital for understanding and modeling the uncertainties inherent in transportation systems, as it provides a range of possible outcomes along with their associated probabilities.
Furthermore, DGMs are notable for their ability to extract deep insights from data and enable a wide range of applications through the manipulation and analysis of latent vectors. The latent vectors represent compressed, lower-dimensional forms of the input data that retain its essential features. This capability makes DGMs powerful tools for capturing and analyzing complex data relationships in transportation applications.

{

Some fundamental concepts and terminologies to understand the following sections are explained in Appendix \ref{sec:prelim}.
}

\subsubsection{Model Training Objective}
The objective of training DGMs is to minimize the discrepancy between the true data distribution $p_{\text{data}} (\mathbf{x})$ and the model's estimated distribution $p_{\text{model}} (\mathbf{x;\theta})$. This discrepancy can be expressed using the KL divergence. The goal is to adjust the model parameters $\theta$ such that the model distribution closely approximates the true data distribution. This is expressed in the optimization problem:
\begin{equation}
\begin{split}
\theta^* = \arg \min_{\theta} D_{KL} (p_{\text{data}}  (\mathbf{x} ) || p_{\text{model}} (\mathbf{x} ; \theta)).
\end{split}
\end{equation}

However, in practice, we do not have direct access to $p_{\text{data}} (\mathbf{x})$, and hence we rely on an empirical dataset (or training dataset) to represent the true data distribution.
Given that direct computation of KL divergence is not feasible without knowledge of $p_{\text{data}} (\mathbf{x})$, DGMs typically rely on maximizing the likelihood of the observed data under the model. This approach aligns with the principle of maximum likelihood estimation (MLE), which posits that the optimal model parameters $\theta$ are those that maximize the likelihood of the training data appearing under the model's assumed probability distribution. Mathematically, this is represented as:
\begin{equation}
\begin{split}
\theta^* = \arg \max_{\theta} \prod_{i=1}^N p_{\text{model}} (\mathbf{x}_i ; \theta ),
\end{split}
\end{equation}
where $\mathbf{x}_i$ denotes an instance of the training data.
To simplify calculations and enhance numerical stability, we typically work with the logarithm of the likelihood function, transforming the product into a summation. This is referred to as the log-likelihood, and it modifies our optimization problem to:
\begin{equation}
\begin{split}
\theta^* = \arg \max_{\theta} \sum_{i=1}^N \log p_{\text{model}} (\mathbf{x}_i ; \theta ),
\end{split}
\end{equation}
which is essentially the mean log-likelihood over all $N$ data points in the training dataset. This is the empirical estimate of the expected log-likelihood under the data distribution $p_{\text{data}} (x)$:
\begin{equation}
\begin{split}
\theta^* = \arg \max_{\theta} \mathbbm{E} \left[ \log p_{\text{model}} (\mathbf{x} ; \theta) \right]= \arg \max_{\theta} \mathbbm{E} \left[ \log p_{\theta} (\mathbf{x} ) \right].
\end{split}
\end{equation}
\subsection{DGM Classification}
DGMs can be classified into different categories based on their approach to maximizing data likelihood. The first one is \textbf{explicit density models} that directly model the probability distribution $p_{\theta} (\mathbf{x})$. These models use the likelihood function during training to align the generated data distribution with the true data distribution.
In contrast, \textbf{implicit density models} do not explicitly use $p_{\theta} (\mathbf{x})$ during training. Instead, they employ alternative methods that theoretically achieve the goal of maximizing data likelihood by implicitly capturing the data distribution. Explicit density models can be further categorized into \textbf{tractable density models} and \textbf{intractable density models} based on whether the likelihood computation during training is directly calculable or must be approximated.

\subsubsection{Explicit Tractable Density Models}
Explicit Tractable Density Models perform direct and explicit computations of the likelihood, making them highly interpretable and theoretically robust. Examples include:
\begin{itemize}
\item Autoregressive Models: Examples include PixelRNN \citep{van2016pixel}, PixelCNN \citep{van2016conditional}, and NADE \citep{uria2016neural}. In these models, each output is generated sequentially with each output conditioned on previous outputs. This sequential generation ensures that the likelihood of each output can be directly calculated.
\item Normalizing Flows: These models utilize invertible transformations to map complex data distributions to simpler, known parametric distributions (such as Gaussian distribution). The invertibility of these transformations allows for the exact computation of the likelihood of any given input, making these models particularly powerful for tasks requiring detailed density estimation.
\end{itemize}

\subsubsection{Explicit Intractable Density Models}
Explicit Intractable Density Models aim to directly model $p_{\theta} (\mathbf{x})$, but they rely on approximation methods due to the infeasibility of exact computation in complex data scenarios. These models include:
\begin{itemize}
\item Variational Autoencoders (VAEs):  VAEs use an encoder-decoder architecture where the encoder approximates the posterior distribution of latent variables, and the decoder approximates the data distribution conditioned on these latent variables. The likelihood is indirectly maximized through a lower bound estimate, known as the Evidence Lower Bound (ELBO).
\item Diffusion Models: These models gradually convert data into a known noise distribution and learn to reverse this process. Training involves approximating the reverse of this diffusion process, which indirectly maximizes the data likelihood.
\item Score-Based Generative Models: These models learn the score (gradient of log probability) of the data distribution and use it to generate samples by iteratively refining noise samples. The model indirectly maximizes the likelihood through the score-matching technique.
\end{itemize}

\subsubsection{Implicit Density Models}
Implicit Density Models do not explicitly use $p_{\theta} (\mathbf{x})$ during training. Instead, they employ alternative approaches that can theoretically align with the goal of maximizing data likelihood by implicitly capturing the data distribution. 
\begin{itemize}
\item Generative Adversarial Networks (GANs): GANs consist of two competing networks—a generator and a discriminator. The generator produces samples aimed at being indistinguishable from real data, while the discriminator evaluates their authenticity. Even though $p_{\theta} (\mathbf{x})$ is not explicitly computed, the adversarial training process theoretically leads to the generator capturing the true data distribution, resulting in implicitly maximizing the data likelihood.
\end{itemize}








\begin{figure}[b]
    \centering
    \begin{subfigure}{0.8\textwidth}
        \centering
        \includegraphics[width=0.9\textwidth]{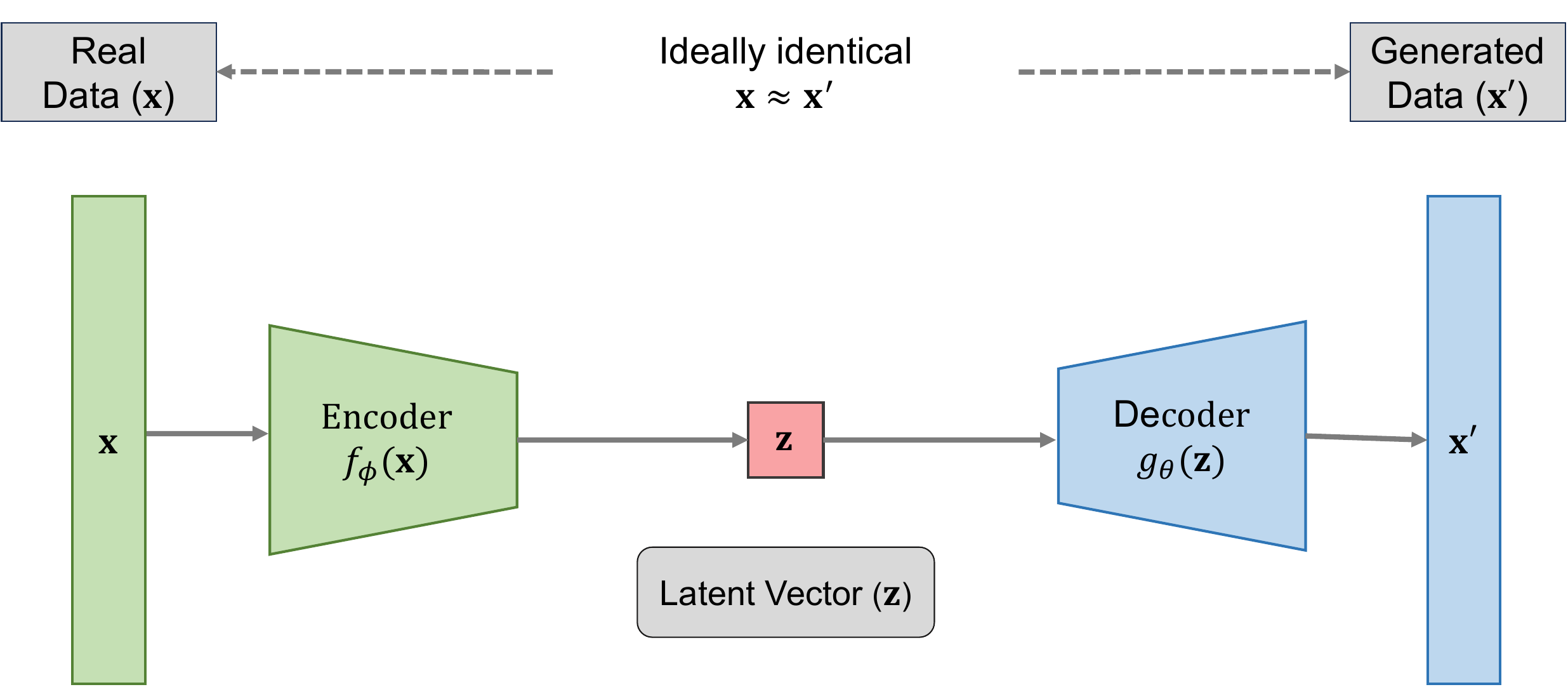} 
        \label{fig:AE}
        \caption{Model architecture of Autoencoder (AE)}
    \end{subfigure}\hfill
    \begin{subfigure}{0.9\textwidth}
        \centering
        \includegraphics[width=0.8\textwidth]{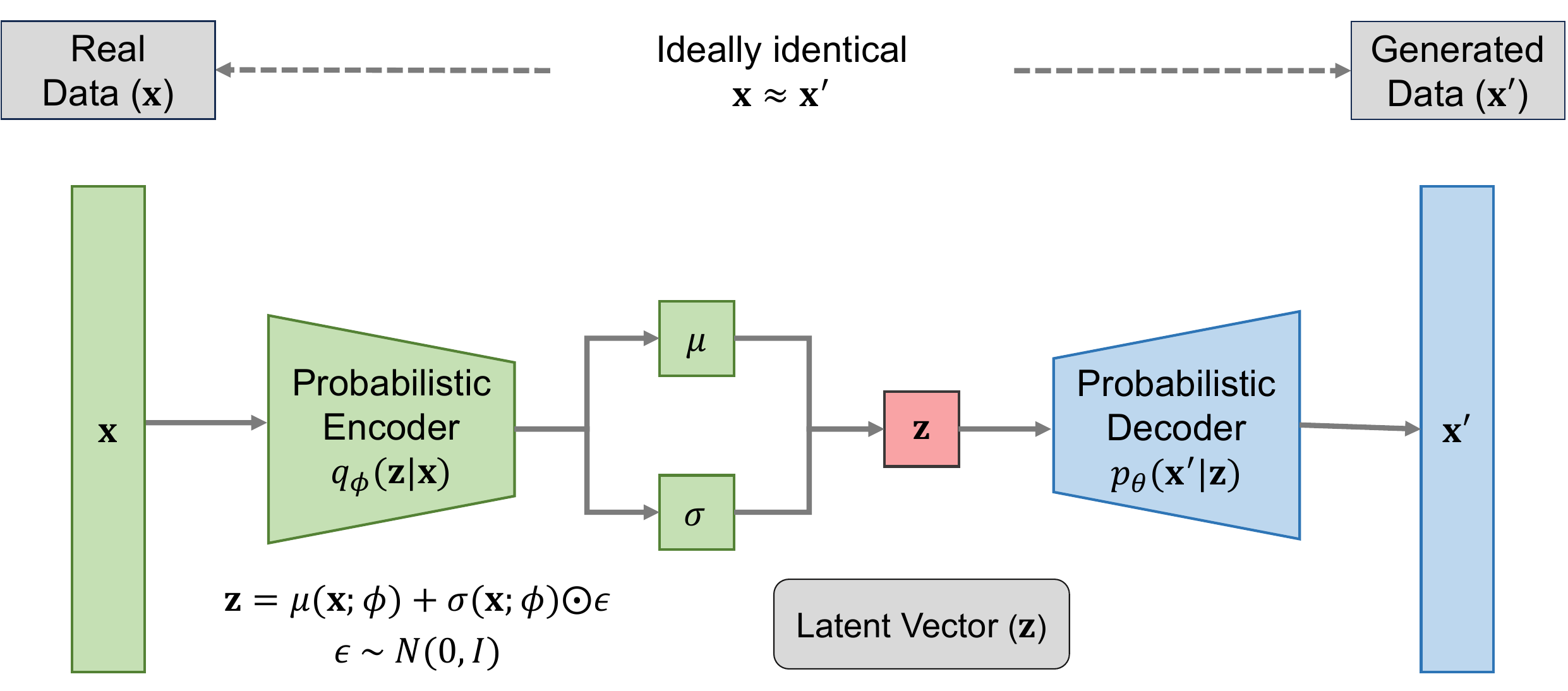} 
        \caption{Model architecture of Variational Autoencoder (VAE)}
        \label{fig:VAE}
    \end{subfigure}
\caption{{Schematic overview of AE and VAE. Here, $\mathbf{x}$ is the real data and $\mathbf{x}'$ denotes the generated data, and $\mathbf{z}$ represents the latent vector.}}
\label{fig:AEVAE}
\end{figure}

\subsection{Variational Autoencoder}\label{sec:vae}

Variational Autoencoders (VAEs) are a class of DGMs that combine the structure of Autoencoders with variational inference.
Since the introduction in \citet{kingma2013auto}, VAEs have significantly influenced the overall field of generative models.
%
As shown in Figure \ref{fig:AEVAE} (a), Autoencoders (AEs) typically comprise two main components: an encoder and a decoder. AEs were first introduced by \cite{hinton2006reducing} for data compression and dimensionality reduction. Specifically, AEs compress input data $\mathbf{x}$ into a lower-dimensional latent representation $\mathbf{z}$ using the $\phi$-parameterized encoder $f_{\phi} (\mathbf{x}) = \mathbf{z}$ and reconstruct the input data from the latent representation with the $\theta$-parameterized decoder $g_{\theta} (\mathbf{z}) = \mathbf{x}$. The objective function of an AE is typically:
\begin{equation}
    \begin{split}
        \mathcal{L}_{AE} = \| \mathbf{x} -  \mathbf{x'} \|_2^2 
        = \| \mathbf{x} -  g_{\theta} (f_{\phi} (\mathbf{x})) \|_2^2
    \end{split},
\end{equation}
where $\mathbf{x'}$ is the reconstructed data of $\mathbf{x}$.

In contrast, the primary objective of VAEs is data generation, focusing on training a high-performing decoder. The decoder samples from the latent space to generate new data that resemble the original training dataset. Unlike traditional AEs that learn deterministic functions, $f_{\phi}$ and $g_{\theta}$, VAE model probabilistic (or generative) encoder and decoder,  $q_{\phi} (\mathbf{z} | \mathbf{x})$ and $p_{\theta} (\mathbf{x} | \mathbf{z})$, by introducing intermediate latent variable $z$ and using variational inference as shown in Figure \ref{fig:AEVAE} (b).
The $\phi$-parameterized encoder, represented by $q_\phi(\mathbf{z}|\mathbf{x})$, transforms input data $\mathbf{x}$ into a latent representation $\mathbf{z}$ in a lower-dimensional latent space. Notably, instead of learning a deterministic mapping to a single latent vector, VAE's encoder estimates the parameters of a probability distribution in the latent space. 
%
The $\theta$-parameterized decoder, represented by $p_\theta(\mathbf{x}|\mathbf{z})$, takes the encoded latent vector $\mathbf{z}$ and reconstructs the original data point $\mathbf{x}$.

Following the detailed derivations in Appendix \ref{app:vae},
a new objective function called the Evidence Lower BOund (ELBO), $\mathcal{L}(\mathbf{x};\theta,\phi)$, can be defined as:
\begin{equation}
\log p_{\theta}(\mathbf{x}) 
\geq \mathcal{L}(\mathbf{x};\theta,\phi) = \mathbbm{E}_{\mathbf{z}}  \left[ \log p_{\theta} (\mathbf{x} | \mathbf{z}) \right]
- D_{KL} \left( q_{\phi} \left(\mathbf{z}|\mathbf{x} \right) || p (\mathbf{z})  \right).
\label{eq:vae_elbo}\end{equation}

%
Equation~\eqref{eq:vae_elbo} demonstrates that the ELBO is a \textit{`lower bound'} to the log-likelihood of the data. Therefore, maximizing the ELBO with respect to the parameters of the encoder and decoder, $\phi$ and $\theta$, also maximizes the true log-likelihood ($\log p_{\theta} (\mathbf{x}) $). The ELBO can be efficiently estimated using stochastic gradient-based optimization methods without involving the intractable posterior $p_{\theta}(\mathbf{z}|\mathbf{x})$. 
The first term of the right hand side of Equation~\eqref{eq:vae_elbo}, $\mathbbm{E}_{\mathbf{z} \sim q_{\phi} (\mathbf{z} | \mathbf{x})} \left[ \log p_{\theta}(\mathbf{x}|\mathbf{z})  \right]$, is called \textit{reconstruction error}, which measures the VAE's capability to accurately reconstruct the input data from its latent variable generated by the encoder network. The second term, $D_{KL} \left( q_{\phi} \left(\mathbf{z}|\mathbf{x} \right) || p (\mathbf{z})  \right)$, is called \textit{regularization}, which focuses on aligning the distribution of the latent variables with the assumed distribution.

A key challenge in training VAEs is estimating the gradient of the expected value term in the ELBO to the encoder network parameters $\phi$. This involves differentiating through the random sampling operation involved in generating the latent variable $\mathbf{z}$, which is inherently non-deterministic and does not permit direct differentiation.
The \textit{reparameterization trick} addresses this challenge by reparameterizing the random variable $\mathbf{z}$ such that the randomness is independent of the parameters. This enables the model to backpropagate gradients through the deterministic part of the reparameterization, while the stochasticity is handled separately.
Instead of sampling $\mathbf{z}$ directly from $q_{\phi}(\mathbf{z}|\mathbf{x}) = \mathcal{N}(\mu(\mathbf{x};\phi), \sigma(\mathbf{x};\phi)^2)$, the model samples a standard Gaussian noise variable $\epsilon \sim \mathcal{N}(0, I)$ and reparameterizes $\mathbf{z}$ as follows:
\begin{equation}
\mathbf{z} = \mu(\mathbf{x};\phi) + \sigma(\mathbf{x};\phi) \odot \epsilon ,
\end{equation}
where $\odot$ denotes the element-wise multiplication. The gradients with respect to $\phi$ can be computed since the sampling operation is deterministic for $\phi$, and all randomness is relegated to $\epsilon$.


Conditional VAEs (CVAEs) \citep{sohn2015learning} is one of the extensions of the standard VAEs. The core innovation of CVAEs lies in their ability to incorporate additional conditional information, often in the form of labels or related data, directly into the generative process. This conditioning allows the model to generate more targeted and context-specific data. Unlike traditional VAEs, CVAEs leverage the extra conditional information to produce outputs that are not only high in quality but also relevant to the specified conditions. This is achieved by modifying both the encoder and decoder to accept and process the conditional information alongside the input data. As a result, CVAEs have a degree of control and specificity in the generated output, such as generating images of a certain class or style, customizing text generation, and enhancing recommendation systems. They also offer benefits in interpretability and the potential for disentangled representation learning. The CVAE framework has thus emerged as a powerful tool for controlling the generated output and improving the model performance. Applications of CVAE in transportation research varies from traffic data generation, prediction and classification. 



\subsection{Generative Adversarial Networks}\label{sec:gan}

Generative Adversarial Networks (GANs) are another class of generative models, introduced in \citet{goodfellow2014generative}. GANs have attracted considerable attention due to their capability to generate highly realistic data, and they have been widely used in a wide range of applications, including image synthesis, text generation, and data augmentation.
Unlike other types of DGMs, GANs do not \textit{explicitly} model data distribution or data likelihood. Instead, they employ a game-theoretical approach to train a model focused on generating realistic data.

\begin{figure}[!t]
  \centering
  \includegraphics[width=0.8\textwidth]{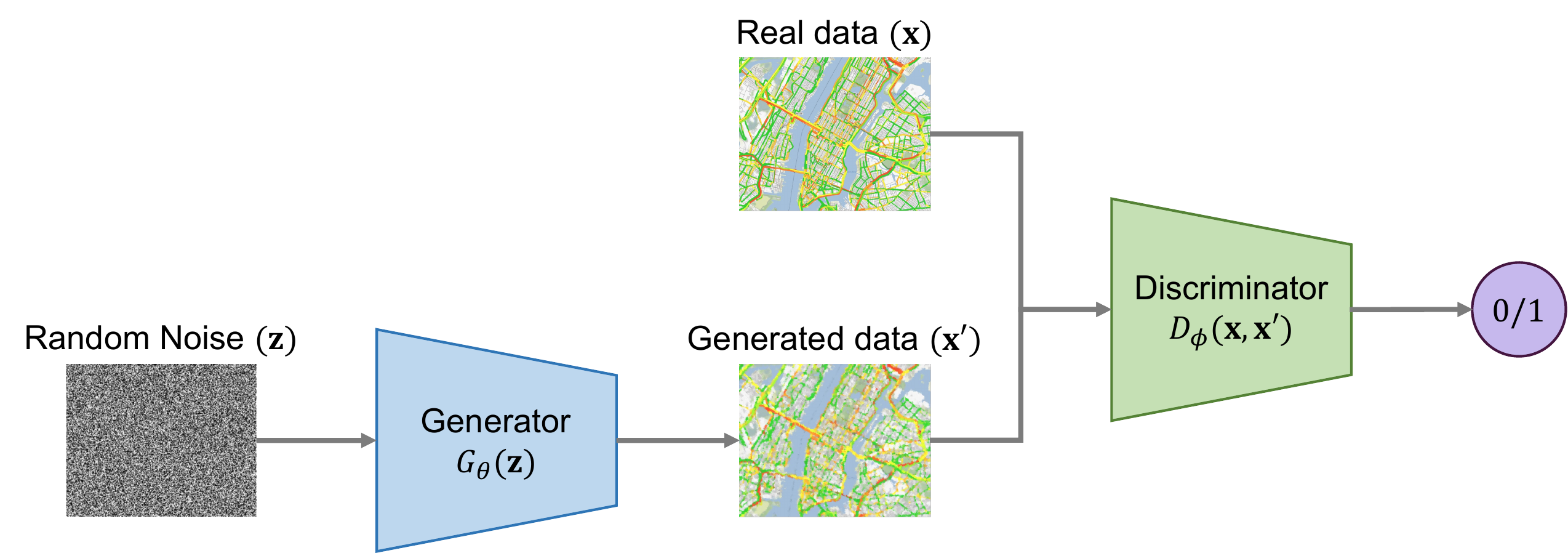}
\caption{{Schematic overview of Generative Adversarial Network (GAN). Here, $\mathbf{x}$ is the real data, $\mathbf{x}'$ denotes the generated data, and $\mathbf{z}$ represents the random noise.}}
\label{fig:GAN}
\end{figure}

As shown in Figure \ref{fig:GAN}, GANs consist of two primary components: a \textit{Generator} (G) and a \textit{Discriminator} (D). The Generator's task is to produce realistic synthetic data, while the Discriminator focuses on distinguishing between real data and synthetic data generated by the Generator. This competitive dynamic resembles a two-player min-max game, where the Generator attempts to deceive the Discriminator, and the Discriminator endeavors to avoid being deceived.  Formally, let the parameters of the Generator $G$ be denoted as $\theta$ and the parameters of the Discriminator $D$ be denoted as $\phi$. The Generator $G$ takes as input a random noise vector $\mathbf{z}$, sampled from a known distribution (often a standard normal distribution), and transforms it into a data instance $\mathbf{x} = G_{\theta}(\mathbf{z})$. The Generator is similar to the decoder of VAE in that both aim to learn the mapping from random noise to the data sample.
The Discriminator $D$ takes as input a data instance, which can either be real data $\mathbf{x}$ or generated data $G_{\theta}(\mathbf{z})$, and outputs a scalar representing the probability that the input data is real, denoted as $D_{\phi}(\mathbf{x})$.
The learning process is guided by a value function $V(\phi, \theta)$, defined as:
\begin{equation}
V(\phi, \theta) = \mathbbm{E}_{\mathbf{x} \sim p_{\text{data}}(\mathbf{x})}[\log D_{\phi}(\mathbf{x})] + \mathbbm{E}_{\mathbf{z} \sim p(\mathbf{z})}[\log(1 - D_{\phi}(G_{\theta}(\mathbf{z})))],
\label{eqn:gan_1}
\end{equation}
where $p_{\text{data}}$ represents the true data distribution, and $p({\mathbf{z}})$ denotes the distribution of the input noise vectors. The first and second terms in the value function correspond to the log probability of the Discriminator correctly identifying real and generated data, respectively. 
The learning process involves finding optimal parameters, $\phi^{*}$ and $\theta^*$, for both $D$ and $G$, through gradient-based methods as shown below. This process alternates between the following two steps:
\begin{equation}
\label{eqn:gan_2}
\min_{{\theta}} \max_{{\phi}} V(\phi, \theta),
\end{equation}
\begin{enumerate}
    \item Maximizing $V(\phi, \theta)$ with respect to $\phi$: This step enhances the Discriminator's capability to distinguish between real and generated data.
    
    \item Minimizing $V(\phi, \theta)$ with respect to $\theta$: This step improves the Generator's ability to generate synthetic data that can deceive the Discriminator.
    
\end{enumerate}

This process does not explicitly involve the evaluation of the likelihood function; however, it implicitly maximizes the data likelihood and minimizes the difference between true data distribution $p_{\text{data}}$ and model distribution $p_{\text{model}}$. 
{

A detailed derivation of how GAN formulation matches with maximizing likelihood is shown in Appendix \ref{sec:discuss_gan}.
}

Conditional GANs \citep{mirza2014conditional}, commonly known as cGANs, are an advanced adaptation of the traditional GAN framework. The primary motivation behind cGANs is to give more control over the data generation process. While traditional GANs are adept at generating data from a learned distribution, cGANs add an additional layer of conditioning, usually based on labels or external information. For example, in the case that the user wants to generate two distinctive classes of outputs, standard GANs would require two separate models to learn each distribution or, at best, generate a mix without explicit control over the output. On the other hand, cGANs allow the model to generate images of a specific category based on a label provided to it. The beauty of cGANs lies in the fact that both the generator and the discriminator are conditioned on this label. The generator utilizes this label to produce data, while the discriminator uses it to discern the authenticity of the generated data. The mathematical representation encapsulates this conditioning through additional input layers for the generator and discriminator. When producing an output sample, the generator does not merely rely on the random noise vector but also factors in the label, ensuring the generated output aligns with the desired category. On the flip side, the discriminator evaluates both the output and its corresponding label, ensuring that fake (generated) data is not only realistic but is also consistent with its associated type. 




Generative Adversarial Imitation Learning \citep{ho2016generative}, or GAIL, is an innovative fusion of imitation learning principles with the GAN framework. Imitation learning, at its core, revolves around the idea of learning a policy (or behavior) by using experts' demonstrations. GAIL captures this essence but employs the adversarial training approach of GANs to achieve it. In the GAIL framework, the generator is no longer producing static data; instead, it produces sequences of actions aiming to imitate an expert's behavior. Meanwhile, the discriminator's role is to distinguish between sequences produced by the generator and those exhibited by the expert. Traditional imitation learning often relies on methods like behavioral cloning, where the agent directly learns from expert trajectories. However, this method is prone to compounding errors and struggles with situations not encountered during training. GAIL, on the other hand, captures the underlying reward structure by using the adversarial training framework. 
{

Similar to the concept of cGANs, which integrate additional conditioning information to guide the data generation process, conditional GAIL (cGAIL) \citep{zhang2020cgail} extends the GAIL framework by conditioning both the generator and the discriminator on extra contextual variables. These variables can include environmental factors, state features, or operational conditions that are critical in accurately capturing expert behavior. By incorporating such information, cGAIL is better equipped to model context-dependent behaviors, leading to improved performance and robustness in dynamic, real-world settings.}

{

Adversarial Autoencoders (AAEs) \citep{makhzani2015adversarial} represent an innovative fusion of VAEs and GANs, designed to enhance the capabilities of generative models. The primary advantage of AAEs lies in their ability to impose arbitrary prior distributions on the latent space, going beyond the Gaussian priors typically used in VAEs. This flexibility allows for the modeling of complex data distributions more effectively. AAEs consist of a standard autoencoder architecture, complemented by an adversarial network that enforces the latent space to conform to the chosen prior distribution. To be more specific, the fundamental condition in VAE loss is that KL divergence between the distribution of encoded input data and prior can be calculated. The function of KL divergence in loss term is making the distribution of encoded input data $q_{\phi} \left(\mathbf{z}|\mathbf{x} \right)$ the same as the prior distribution $p (\mathbf{z})$, which is the same logic of GAN. Therefore, AAE utilizes the discriminator term of GAN to replace the KL divergence of VAE, making it possible to use arbitrary prior distribution. Through adversarial training, AAEs generate sharper, more detailed outputs compared to traditional VAEs, especially noticeable in tasks like image generation. They are versatile in applications, ranging from semi-supervised learning to unsupervised clustering and anomaly detection. Additionally, AAEs offer better control over the generation process, including conditional generation and the learning of disentangled representations, making them suitable for tasks requiring precise control and interpretability. The integration of adversarial principles into autoencoders has thus positioned AAEs as a powerful and flexible tool in the realm of generative modeling, addressing key challenges of traditional VAEs and opening new avenues in machine learning research.}

\subsection{Normalizing Flows (Flow-based Generative Models)}\label{sec:nf}
A flow-based deep generative model learns the likelihood by using \textit{normalizing flows}. A normalizing flow describes a method for constructing a complex distribution by using a series of invertible mapping functions \citep{rezende2015variational}. The idea behind using normalizing flows to learn the likelihood of the real data is that a complex distribution can be learned by sequentially applying multiple normalizing flows to a simple base distribution (e.g., Gaussian distribution).

\begin{figure}[t]
  \centering
  \includegraphics[width=0.9\textwidth]{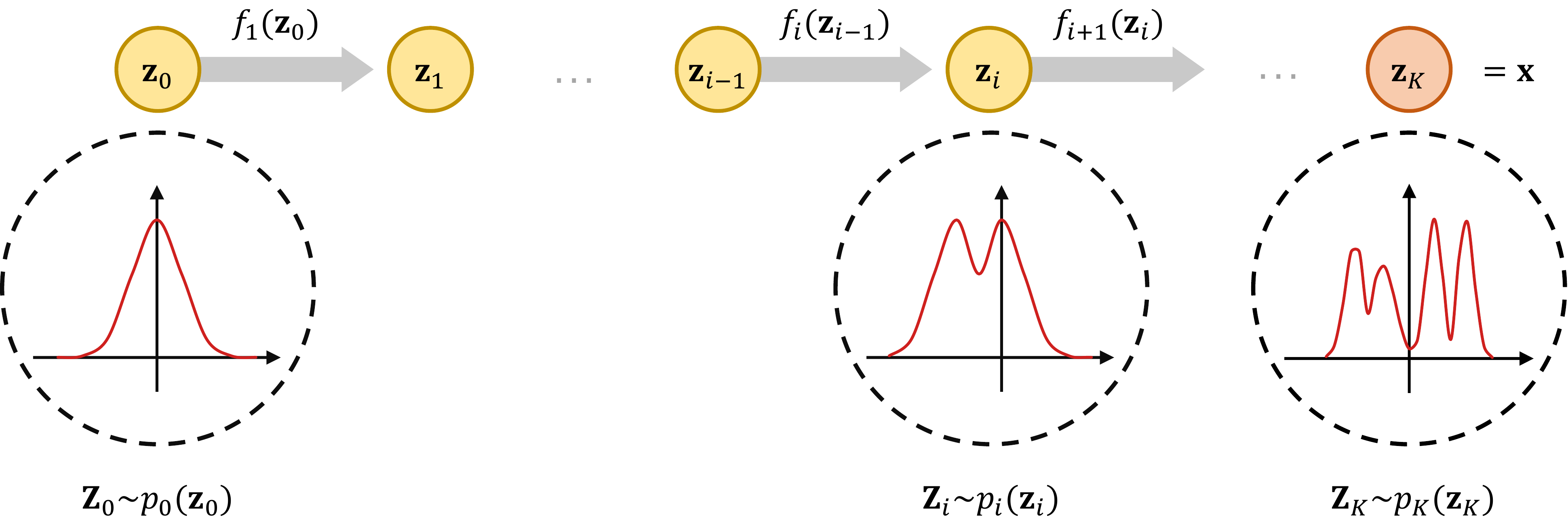}
  \caption{{Schematic overview of normalizing flow. Here, $\mathbf{z}_0$ is a simple, known distribution (such as a standard Gaussian), $\mathbf{z}_i$ represents an intermediate distribution, and $\mathbf{z}_K$ denotes the target distribution.}}

  \label{fig: NF}
\end{figure}

As shown in Figure \ref{fig: NF}, for a multivariate vector $\textbf{x}$, the objective of normalizing flow is to learn the probability density function $p (\textbf{x})$. The density $p (\textbf{x})$ is transformed into a simple distribution $p(\mathbf{z}_0)$ (independent multivariate Gaussian distribution). This transformation $f$ is composed of $K$ invertible (bijective) functions $f_i, i\in \{1,...K\}$. This transformation can be denoted as follows:
\begin{equation}
    \begin{split}
    & \mathbf{x} = \mathbf{z}_K = f_K \circ f_{K-1} \circ \cdots \circ f_1(\mathbf{z}_0) = f(\mathbf{z}_0), \\
    \end{split}
\end{equation}
where we denote $\mathbf{z}_i$ as latent vector after applying $i$ invertible functions ($f_1,\cdots,f_i$). 


%
By definition,
\begin{equation}
    \begin{split}
    & \mathbf{z}_i = f_i(\mathbf{z}_{i-1}), \text{ thus } \mathbf{z}_{i-1} = f_i^{-1} (\mathbf{z}_i) \\
    \end{split}.
\end{equation}

The log-likelihood of probability density function of $i$-th latent vector, $p(\mathbf{z}_i)$, can be calculated as follows:
\begin{equation}
    \begin{split}
    &\int p_{i}(\mathbf{z}_{i}) \, d\mathbf{z}_{i} = \int p_{i-1}(\mathbf{z}_{i-1}) \, d\mathbf{z}_{i-1} =1  && \Rightarrow  
    \textit{ \small integration property of probability distribution} \\ 
    & p_i(\mathbf{z}_i) = p_{i-1} \left(f_i^{-1} (\mathbf{z}_i) \right) \left| \det \left({\frac{df_i^{-1}}{d\mathbf{z}_i} }  \right) \right| && \Rightarrow  
    \textit{ \small Change of Variable Theorem} \\
    & p_i(\mathbf{z}_i) = p_{i-1} \left( \mathbf{z}_{i-1} \right)  \left| \det  \left( {\frac{df_i}{d{\mathbf{z}_{i-1}}}} \right)^{-1} \right| && \Rightarrow  
    \textit{ \small Inverse Function Theorem} \\
    & p_i(\mathbf{z}_i) = p_{i-1} \left( \mathbf{z}_{i-1} \right)  \left|   \det \left( {\frac{df_i}{d{z_{i-1}}}} \right) \right|^{-1} && \Rightarrow   
    \textit{ \small property of Jacobians of invertible function}\\
    & \log p_i(\mathbf{z}_i) = \log p_{i-1} \left( \mathbf{z}_{i-1} \right) - \log \left|  \det \left( {\frac{df_i}{d{z_{i-1}}}} \right) \right|. && \Rightarrow   
    \textit{ \small by applying logarithm to both sides} \\
    \end{split}
\end{equation}

As a results, the probability density of $\mathbf{x}$ can be calculated as follows:
\begin{equation}
\label{eqn:nf_1}
    \begin{split}
    \log p(\mathbf{x}) = \log \pi_K (\mathbf{z}_K) 
    & = \log p_{K-1} (\mathbf{z}_{K-1}) - \log \left| \det \left({\frac{df_K}{dz_{K-1}}} \right) \right|\\
    & = \log p_{K-2} (\mathbf{z}_{K-2}) - \log \left| \det \left( {\frac{df_{K-1}}{dz_{K-2}}} \right) \right| - \log \left| \det \left( {\frac{df_K}{dz_{K-1}}} \right) \right|\\
    & = \cdots \\
    & = \log p_{0} (z_{0}) - \sum_{i=1}^K \log \left| \det \left(  {\frac{df_i}{dz_{i-1}}} \right) \right|.
    \end{split}
\end{equation}

Finally, the objective of training for a generative model is maximizing the likelihood of a training set, $\mathcal{D}$, as follows:
\begin{equation}
\label{eqn:nf_2}
    \mathcal{L} (\mathcal{D}) = - \frac{1}{|\mathcal{D}|} \sum_{x\in\mathcal{D}} \log p(\mathbf{x}).
\end{equation}

In practice, there are two conditions to consider when deciding proper functions for normalizing flows. The first condition is that by definition, the functions should be invertible. Also, computing the Jacobian determinant should be feasible since usually computing the Jacobian of functions and computing the determinant are both computationally expensive. As a result, properly defining the function $f$ is the key to normalizing flows. 

There are several feasible functions from previous studies that satisfy the constraints: linear function \citep{rezende2015variational}, $1\times1$ convolution \citep{kingma2018glow}, and affine coupling layer \citep{dinh2014nice,durkan2019neural}. In this paper, we introduce the affine coupling layer as an example since this is one of the widely used forms due to its ability to deal with high-dimensional data such as image and sound. The affine coupling layer from \cite{dinh2016density} can be formulated as follows:
\begin{equation}
    \label{eqn:affine}
    \begin{split}
    & \mathbf{y}_{1:d} = \mathbf{x}_{1:d}, \\
    & \mathbf{y}_{d+1:D} = \mathbf{x}_{d+1:D} \odot \exp{\left( s(\mathbf{x}_{1:d} ) \right)} + t(\mathbf{x}_{1:d} ), \\    
    \end{split}
\end{equation}
where $\mathbf{y}_{1:d}$ is the first $d$ elements of $\mathbf{y}$, $\mathbf{y}_{d+1:D}$ is the rest of $\mathbf{y}$, $\mathbf{x}_{1:d}$ is the first $d$ elements of $\mathbf{x}$, $\mathbf{x}_{d+1:D}$ is the rest of $\mathbf{x}$, $s$ and $t$ are scale and translation functions, which are usually implemented as multi-layer perceptrons (MLP), and $\odot$ is the element-wise multiplication operator.
As noted in \cite{dinh2016density}, the affine coupling layer is an invertible function and guarantees fast computation of the determinant of the Jacobian matrix. 

First, the inverse function of Equation~\eqref{eqn:affine} is as follows:
\begin{equation}
\label{eqn:nf_4}
    \begin{split}
    & \mathbf{x}_{1:d} = \mathbf{y}_{1:d}, \\
    & \mathbf{x}_{d+1:D} = \left( \mathbf{y}_{d+1:D} - t\left( \mathbf{y}_{1:d}  \right)\right) \odot \exp{\left( -s\left( \mathbf{y}_{1:d} \right) \right)}.\\
    \end{split}
\end{equation}

Also, the Jacobian matrix of this transformation is
\begin{equation}
    \begin{split}
    \frac{\partial \mathbf{y}}{\partial \mathbf{x}^\top} = \begin{bmatrix}
\mathbbm{I}_d & 0 \\
\frac{\partial \mathbf{y}_{d+1:D}}{\partial \mathbf{x}_{1:d}^\top} & \diag\left( \exp{\left[ s(\mathbf{x}_{1:d} ) \right]} \right)
\end{bmatrix},
    \end{split}
    \label{eq:jac}
\end{equation}
\noindent
where $\mathbbm{I}_d$ is a $d\times d$ identity matrix, $\diag(*)$ is the diagonal matrix whose diagonal entries correspond to the given matrix and non-diagonal entries are zero.

As a result, the Jacobian matrix shown in Equation~\eqref{eq:jac} is a lower-triangular matrix. Therefore, the determinant of the Jacobian matrix can be calculated as follows:
\begin{equation}
\label{eqn:nf_6}
    \begin{split}
    \left| \det\left(\frac{\partial \mathbf{y}}{\partial \mathbf{x}^\top} \right) \right| & = \prod_{i=1}^{D} \left[ \frac{\partial \mathbf{y}}{\partial \mathbf{x}^\top} \right]_{i,i} =\prod_{i=1}^{d} \exp{\left[s(\mathbf{x}_{1:d})\right]_{i,i}} =\exp{ \left( \sum_{i=1}^{d} \left[s(\mathbf{x}_{1:d})\right]_{i,i} \right)}, \\
    \end{split}
\end{equation}
and
\begin{equation}
\label{eqn:nf_7}
    \begin{split}
    \log \left| \det\left(\frac{\partial \mathbf{y}}{\partial \mathbf{x}^\top} \right) \right| 
    &= \sum_{i=1}^{d} \left[s(\mathbf{x}_{1:d})\right]_{i,i}, \\
    \end{split}
\end{equation}
\noindent
where $[*]_{i,i}$ is $i$-th diagonal entry of the given matrix.

\subsection{Score-Based Generative Models}\label{sec:score}
\begin{figure}[b]
  \centering
  \includegraphics[width=0.8\textwidth]{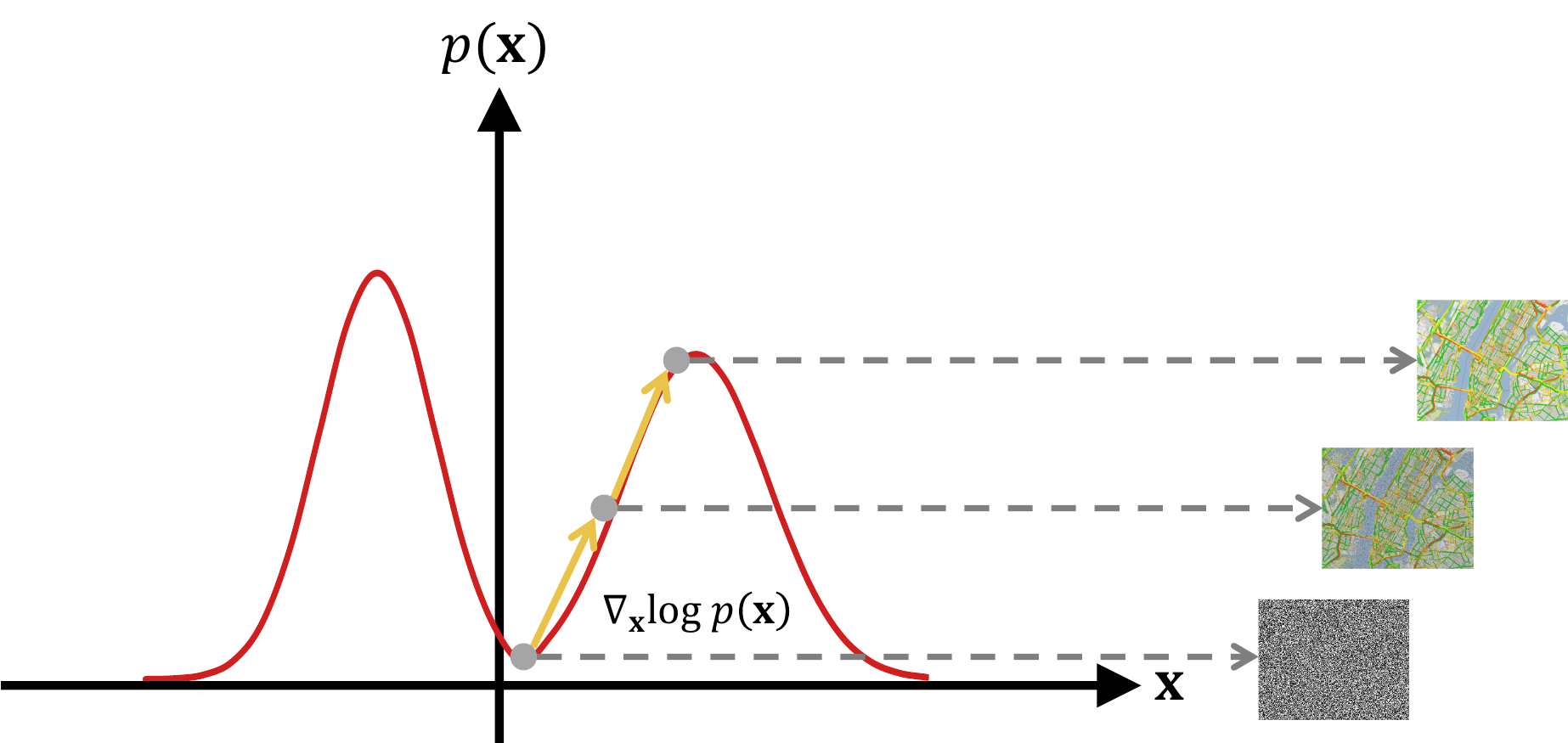}
   \caption{{Schematic overview of score-based generative model. Here, $\mathbf{x}$ denotes a data sample from the underlying distribution, and $\nabla_{\mathbf{x}} \log p(\mathbf{x})$ represents the score function.}}
   \label{fig: Score_generated_model}

\end{figure}

Score-based generative models are a class of generative models that leverage the concept of ``score'' of a probability density function to generate new data samples. The term \textit{score} is used in a statistical context, referring to the gradient (or intuitively, \textit{slope}) of the log-likelihood function with respect to the data, $\mathbf{x}$. Formally, the score is denoted as $\nabla_{\mathbf{x}} \log p(\mathbf{x})$, as shown in Figure \ref{fig: Score_generated_model}. Intuitively, the score tells us how to change $\mathbf{x}$ to increase the probability of $\mathbf{x}$ under the model's current estimate of the probability density function. A higher score indicates a direction in which the data point $\mathbf{x}$ could be adjusted to make it more likely under the model.

In score-based generative models, this concept is used to iteratively refine samples from an initial noise distribution, guiding them towards regions of higher probability under the target distribution (i.e., the distribution of the training data). This is achieved by estimating the score function at various points in the data space and using it to perform gradient ascent on the log-likelihood landscape. Essentially, by following the direction of the score, new data samples can be generated that are likely under the model's learned distribution, thus resembling the characteristics of the training data.

The goal of (deep) generative models is to learn the underlying data distribution $p(\mathbf{x})$ given training data samples drawn from true data distribution, $\{\mathbf{x}_1, \mathbf{x}_2, \cdots, \mathbf{x}_N \}$ with $\mathbf{x}_i \sim p(\mathbf{x}), \forall i \in {1, \cdots, N}$. 
The score-based generative models \citep{song2019generative,song2020score} focuses on the development of a ``score network,'' denoted as $s_\theta (\mathbf{x})$, which is essentially a $\theta$-parameterized neural network designed to approximate the score of the data distribution; that is, $s_\theta (\mathbf{x})$ aims to be a close estimate of $\nabla_{\mathbf{x}} \log p(\mathbf{x})$.
The score network can be trained by minimizing the squared $L2$ distance between the ground-truth score and the estimated score by the score network as follows:
\begin{equation}
\label{eq:score}
    \begin{split}
    \mathbbm{E}_{\mathbf{x} \sim p(\mathbf{x})} \| \nabla_{\mathbf{x}} \log p(\mathbf{x}) - s_{\theta} (\mathbf{x}) \|^2_2.
    \end{split}
\end{equation}

A significant challenge with this approach is that it involves direct access to $\nabla_{\mathbf{x}} \log p(\mathbf{x})$, which is not practically feasible. To address this issue, a substantial body of research has been dedicated to developing alternative methodologies, collectively known as ``score matching'' \citep{hyvarinen2005estimation, vincent2011connection, song2020sliced} that minimizes Equation~\eqref{eq:score} without having to have access to the ground-truth data score.

One of the main concerns in score-based generative models is figuring out how the \textit{score network} can be efficiently and accurately trained. \cite{song2019generative} identified several challenges in training the score network. One of the main challenges is that since most training data is located in a small subspace of the high dimensional data space, training the score network in low data density regions can be challenging. 

Therefore, \cite{song2019generative} proposed to perturb the data space using a scheduled Gaussian noise and learn the score function using the perturbed data following \cite{vincent2011connection}.
This approach is also called ``denoising score matching,'' and as shown in \cite{vincent2011connection}, if the perturbation is small enough, learning the score function of a perturbed data distribution ($q_\sigma \left( \Tilde{\mathbf{x}} | \mathbf{x} \right) $) is almost surely equivalent to learning the actual score function of $p(\mathbf{x})$, i.e., $\nabla_{\mathbf{x}} \log q_\sigma (\Tilde{\mathbf{x}} | \mathbf{x}) \approx \nabla_{\mathbf{x}}$. Therefore, the updated learning objective is as follows:
\begin{equation}
    \label{eqn:score_1}
    \begin{split}
& \mathbbm{E}_{\mathbf{x} \sim p(\mathbf{x})} \left[  \mathbbm{E}_{\Tilde{\mathbf{x}} \sim q_\sigma \left( \Tilde{\mathbf{x}} | \mathbf{x} \right) } \norm{ \nabla_{\mathbf{x}} \log q_\sigma \left( \Tilde{\mathbf{x}} | \mathbf{x} \right) - s_{\theta} (\Tilde{\mathbf{x}})}^2_2 \right] \\
= &  \mathbbm{E}_{\mathbf{x} \sim p(\mathbf{x})} \left[ \mathbbm{E}_{\Tilde{\mathbf{x}} \sim \mathcal{N} \left(\mathbf{x} , \sigma^2 \mathbbm{I} \right)} \norm{ \frac{\Tilde{\mathbf{x}} - \mathbf{x}}{\sigma^2} - s_{\theta} (\Tilde{\mathbf{x}})}^2_2 \right] , \\
    \end{split}
\end{equation}
where, $\Tilde{\mathbf{x}}$ is the perturbed data with Gaussian distribution with standard deviation $\sigma$. 
\cite{song2019generative} further proposed to use a variance scheduling for data perturbation. Formally, we can define a set of standard deviation of data perturbation, $\{\sigma_t\}^L_{t=1}$, that satisfies $\frac{\sigma_1}{\sigma_2}=\cdots=\frac{\sigma_{L-1}}{\sigma_L}>1$. Specifically, we start from a large enough data perturbation that can cover (or generate data in) the low data density regions in the data space, and then we gradually decrease the perturbation so that the learned score function can match with the actual score function. Intuitively, we can interpret the large perturbation as learning the general trend or landscape of the score function. In contrast, we can interpret the small perturbation as learning the details of the actual score function.

Then, the score network is trained given the current noise level ($\sigma_t$) and this is called the ``Noise Conditional Score Network.'' For a given $\sigma$, the updated learning objective is:
\begin{equation}
    \label{eqn:score_2}
    \begin{split}
\mathbbm{E}_{\mathbf{x} \sim p(\mathbf{x})} \left[ \mathbbm{E}_{\Tilde{\mathbf{x}} \sim \mathcal{N} \left(\mathbf{x} , \sigma^2 \mathbbm{I} \right)} \norm{ \frac{\Tilde{\mathbf{x}} - \mathbf{x}}{\sigma^2} - s_{\theta} (\Tilde{\mathbf{x}} , \sigma)}^2_2 \right]. \\
    \end{split}
\end{equation}

Once the score network is trained, data generation can be accomplished through an iterative method known as \textit{Langevin dynamics}, or Langevin Monte Carlo \citep{parisi1981correlation, grenander1994representations}. Langevin dynamics is originally formulated to describe the behavior of molecular systems, such as Brownian motion. A more detailed derivation of equations for Langevin dynamics in the generative model setting is presented in Appendix~\ref{appendix:score_langevin}.
Given a fixed step size $\epsilon>0$, and an initial random noise $\mathbf{x}$, the following iterative procedure can be used:
\begin{equation}\label{eq:score_wiener2}
\mathbf{x}_{i+1} \leftarrow \mathbf{x}_{i} + \nabla_{x} \log p(\mathbf{x}) \cdot \epsilon + \sqrt{2\epsilon} \cdot \mathbf{z}_i ,
\end{equation}
where $\mathbf{z}_i \sim \mathcal{N}(0,I)$. Since the values from the trained score network should match the true score, we can use the following iterative procedure to draw samples from the distribution of interest:
\begin{equation}
\mathbf{x}_{i+1} \leftarrow \mathbf{x}_{i} +  s_{\theta} (\mathbf{x}_i) \cdot \epsilon + \sqrt{2\epsilon} \cdot \mathbf{z}_i .
\end{equation}

\subsection{Diffusion Models}\label{sec:diff}

Diffusion models are a class of DGMs that have shown a remarkable ability to generate high-quality samples across various domains. The core concept behind diffusion models is inspired by the physical process of ``diffusion,'' where particles move from areas of higher concentration to lower concentration until they reach an equilibrium. Diffusion models work by gradually adding noise to data until it becomes indistinguishable random noise (known as a forward diffusion process, or diffusion process), and then learning the reverse of this process of denoising the data (known as a reverse diffusion process, or generation process).

\begin{figure}[b]
  \centering
  \includegraphics[width=0.9\textwidth]{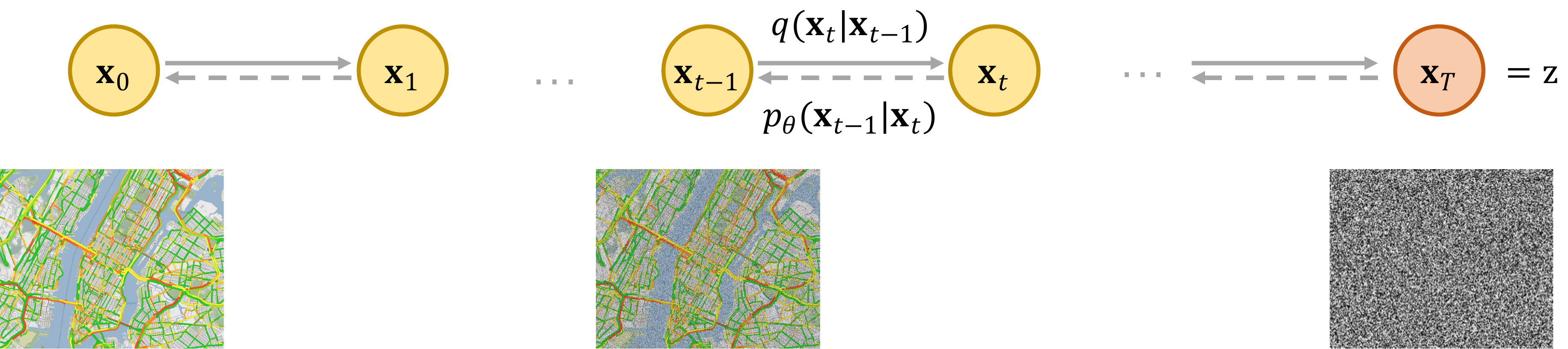}
 \caption{{Schematic overview of diffusion model. Here, $\mathbf{x}_0$ represents the original data, $\mathbf{x}_t$ represents the intermediate noisy states, and $\mathbf{x}_T$ represents the noise data (e.g., Gaussian distribution).}}

\label{fig: diffusion}

\end{figure}

Figure~\ref{fig: diffusion} shows the general framework of the diffusion model. We start from $\mathbf{x}_0$ which is the real data. At each diffusion step, we add a small noise. After a sufficiently large number of steps (here we denote it as $k$), the data is nearly complete noise as shown in $\mathbf{x}_k=\mathbf{z}$. The forward process of adding noise is given, and the main `learning' part of the diffusion model is to learn the reverse process which transforms the complete noise $\mathbf{z}$ to the data $\mathbf{x}_0$. This is similar to VAE, GAN, and Normalizing Flows that we start from a complete noise (or well-known distribution such as Gaussian distribution or uniform distribution) and we learn to transform the noise into the data distribution $p(\mathbf{x})$. The main difference is the idea inspired by the physical ``diffusion'' process and denoising process which recovers the data from a complete noise.

\cite{ho2020denoising} presents the \textit{Denoising Diffusion Probabilistic Model}, or DDPM, which extends the diffusion model framework by incorporating a denoising autoencoder-like architecture. This architecture allows for more efficient learning of the reverse diffusion process, enabling the model to generate high-fidelity samples from complex data distributions.

The forward diffusion process is designed to corrupt the original data \( \mathbf{x}_0 \) over a series of time steps \( t = 1, 2, \ldots, T \). At each step \( t \), Gaussian noise is added to the data, which is mathematically described by the conditional distribution \( q(\mathbf{x}_t | \mathbf{x}_{t-1}) \). This distribution specifies how the data at step \( t \) is generated from the data at the previous step \( t-1 \).

Formally, the forward process is defined as:
\begin{equation}\label{eq:diffusion_forward}
 q (\mathbf{x}_t | \mathbf{x}_{t-1}) = \mathcal{N}(\mathbf{x}_t; \sqrt{1-\beta_t} \mathbf{x}_{t-1}, \beta_t \mathbbm{I}) ,
\end{equation}
where $q(\mathbf{x}_t | \mathbf{x}_{t-1})$ follows the conditional Gaussian distribution with mean $\sqrt{1-\beta_t} \mathbf{x}_{t-1}$ and covariance $\beta_t \mathbbm{I}$, $t$ is the diffusion timestep, and $\beta_t$ is a time-dependent coefficient that controls the amount of data retained from the previous step where $0 < \beta_t \ll 1$. The choice of $\sqrt{1-\beta_t}$ and $\beta_t$ ensures that the variance of each diffusion step is maintained at 1. This is important because it stabilizes the diffusion process, preventing the variance from either exploding or vanishing over time. 

Based on the reparameterization trick, the transition from \( \mathbf{x}_{t-1} \) to \( \mathbf{x}_t \) can be also expressed as:
\begin{equation}\label{eq:diffusion_forward_repar}
\mathbf{x}_t = \sqrt{1-\beta_t} \mathbf{x}_{t-1} + \beta_t \boldsymbol{\epsilon}_t ,
\end{equation}
where \( \boldsymbol{\epsilon}_t \sim \mathcal{N}(0, I) \) is Gaussian noise with mean 0 and identity covariance matrix.

From Equation~\eqref{eq:diffusion_forward}, we can derive the equation for $q(\mathbf{x}_{0:T})$ representing the diffusion trajectory from the original data $\mathbf{x}_{0}$ to the complete noise $\mathbf{x}_{T}=z$ based on properties of Markov chain and chain rules as follows:
\begin{equation}\label{eq:diffusion_q}
\begin{split}
&q\left(\mathbf{x}_{0:T}) = q(\mathbf{x}_{0}\right) \prod_{t=1}^T q\left(\mathbf{x}_t | \mathbf{x}_{t-1} \right) ,\\
&\text{or,}\\
&q\left(\mathbf{x}_{1:T} | \mathbf{x}_0\right) = \prod_{t=1}^T q\left(\mathbf{x}_t | \mathbf{x}_{t-1}\right) .\\
\end{split}
\end{equation}

The reverse diffusion process (the denoising process) represents the generative distribution parameterized with $\theta$. The joint distribution $p_\theta \left( \mathbf{x}_{0:T} \right)$ is defined as a Markov chain with Gaussian transition distributions, $p_\theta \left( \mathbf{x}_{t-1}  | \mathbf{x}_t \right)$ starting at $p\left( \mathbf{x}_{T} \right)  = \mathcal{N} (\mathbf{x}_T; \mathbf{0}, \mathbbm{I})$ as follows:
\begin{equation}\label{eq:diffusion_p}
p_{\theta} \left(\mathbf{x}_{0:T}\right)  = p \left(\mathbf{x}_T \right) \prod_{t=1}^T p_\theta \left( \mathbf{x}_{t-1} | \mathbf{x}_t \right) ,
\end{equation}
where
\begin{equation}\label{eq:diffusion_pdef}
p_\theta \left( \mathbf{x}_{t-1} | \mathbf{x}_t \right) = \mathcal{N} \left( \mathbf{x}_{t-1} ; \boldsymbol{\mu}_\theta( \mathbf{x}_t, t) , \mathbf{\Sigma}_\theta( \mathbf{x}_t, t )  \right) .
\end{equation}

Similar to VAE and Normalizing Flows (models using explicit densities for training), the training objective is to maximize the likelihood or minimize the negative log-likelihood 
\begin{equation}\label{eq:diffusion_objective}
    \theta^* = \arg\min_\theta \mathbbm{E} \left[ - \log p_\theta \left( \mathbf{x}_0 \right) \right].
\end{equation}

Instead of directly training the model with negative log-likelihood, we can consider $q$ as the approximate posterior and use the variational bound on negative log-likelihood to train the model. A detailed derivation is presented in Appendix~\ref{appendix:diffusion_derivation} and \cite{ho2020denoising}. As a result, we can find an upper bound of the optimization
\begin{equation}
\begin{split}
    \mathbbm{E} \left[ -\log p_\theta \left( \mathbf{x}_0 \right)  \right] \leq \mathbbm{E} \left[ 
\underbrace{D_{KL} \left( q\left(\mathbf{x}_{T} | \mathbf{x}_0 \right) || p \left(\mathbf{x}_T \right) \right) }_{L_T}
+ \sum_{t=2}^T \underbrace{D_{KL} \left(  q\left(\mathbf{x}_{t-1} | \mathbf{x}_{t} , \mathbf{x}_0 \right)  || p_\theta \left( \mathbf{x}_{t-1} | \mathbf{x}_t \right) \right) }_{L_{t-1}}
\underbrace{- \log p_\theta \left(\mathbf{x}_{0} | \mathbf{x}_1 \right)}_{L_0}
\right] .
\end{split}
\end{equation}

Therefore, the overall loss function of minimizing the negative log-likelihood in Equation~\eqref{eq:diffusion_objective} is decomposed into several losses, $L_T$, $L_{t-1}$, and $L_0$. Here, $L_T$ is constant since both $q\left(\mathbf{x}_{T} | \mathbf{x}_0 \right) $ and $p \left(\mathbf{x}_T \right) $ are fixed, and therefore, we can ignore this term. Also, in \cite{ho2016generative}, $L_0$ is explicitly defined by using the characteristics of the image generation problem, and as a result, $L_0$ can be interpreted as a reconstruction loss of a problem-specific decoder. As a result, the actual learning process of the diffusion model is related to $L_{t-1}$.

$L_{t-1}$ measures the KL-divergence of $q\left(\mathbf{x}_{t-1} | \mathbf{x}_{t} , \mathbf{x}_0 \right)$ from $p_\theta \left( \mathbf{x}_{t-1} | \mathbf{x}_t \right)$. The diffusion process, $q$, represents the process of adding small noise to the data; i.e., given a less noisy data $\mathbf{x}_{t-1}$, the distribution of a more noisy data $\mathbf{x}_{t}$. The first term, $q\left(\mathbf{x}_{t-1} | \mathbf{x}_{t}, \mathbf{x}_0 \right)$, represents the true denoising process which is derived from the definition of $q$ given the true data without noise, $\mathbf{x}_0$. What the diffusion models try to learn is the denoising process $p_\theta \left( \mathbf{x}_{t-1} | \mathbf{x}_t \right)$; i.e., given a more noisy data $\mathbf{x}_{t}$, the distribution of a less noisy data $\mathbf{x}_{t-1}$. As a result, $L_{t-1}$ captures the distributional difference between the true denoising process (given the true data) and the approximated denoising process (without the true data).

Since the diffusion process follows Gaussian distribution, the true reverse process,  $q\left(\mathbf{x}_{t-1} | \mathbf{x}_{t}, \mathbf{x}_0 \right)$, can be assumed to follow a Gaussian distribution if $T$ is sufficiently large, or $T\rightarrow \infty$. Let $q\left(\mathbf{x}_{t-1} | \mathbf{x}_{t}, \mathbf{x}_0 \right) = \mathcal{N} (\mathbf{x}_{t-1} ; \Tilde{\boldsymbol{\mu}}_t (\mathbf{x}_t, \mathbf{x}_0) , \Tilde{\beta}_t \mathbbm{I})$. 
To derive explicit form of $\Tilde{\boldsymbol{\mu}}_t (\mathbf{x}_t, \mathbf{x}_0)$ and $\Tilde{\beta}_t$, first, we should derive a closed-form equation for sampling $\mathbf{x}_t$ at an arbitrary timestep $t$ from Equation~\eqref{eq:diffusion_forward} as shown in Appendix \ref{appendix:diffusion_derivation}.

Then, \cite{ho2020denoising} found that using $L_{simple}$ results in better training in empirical testing.
\begin{equation}
\label{eqn:diff_1}
L_{simple}= \mathbbm{E}_{t,\mathbf{x}_0, \boldsymbol{\epsilon}} \left[ \norm{\boldsymbol{\epsilon} - \boldsymbol{\epsilon}_{\theta} \left(\sqrt{\Bar{\alpha}}_t \mathbf{x}_0 + \sqrt{1-\Bar{\alpha}_t }\boldsymbol{\epsilon} , t \right)}_2^2  \right] .
\end{equation}

\subsection{Discussion}

Different classes of Deep Generative Models (DGMs) use various approaches to model data distributions. Since all models have their imperfections, it's crucial to understand the commonly known pros and cons of each class. Typically, we expect three key things from DGMs: \textbf{1) High-Quality Sample Generation} --- The ability to produce samples that are indistinguishable from real data; \textbf{2) Mode Coverage and Sample Diversity} --- The capacity to capture all the variations in the data, including minority modes, ensuring that the generated samples are diverse and representative of the entire data distribution; and \textbf{3) Fast and Computationally Efficient Sampling}--- The ability to generate samples quickly without requiring extensive computational resources. 

However, most DGMs cannot yet simultaneously satisfy these three key requirements, a challenge often referred to as the \textit{Generative Learning Trilemma} \citep{xiao2021tackling}. This trilemma highlights the inherent trade-offs between high-quality sample generation, mode coverage and diversity, and fast, computationally efficient sampling. For example, Generative Adversarial Networks (GANs) have the advantage of generating high-quality samples rapidly. As discussed in Section~\ref{sec:gan}, GANs are renowned for producing realistic samples through a single forward pass of the generator network, making them computationally efficient at inference time. However, common shortcomings of GANs include mode collapse, where the model fails to capture the full diversity of the data distribution, generating samples from only some modes but not all. On the other hand, Variational Autoencoders (VAEs) and Normalizing Flows usually perform well in terms of mode coverage and fast computation time. VAEs promote diverse and representative sampling by modeling the entire data distribution, and they allow for quick sample generation by direct decoding from latent variables. Normalizing Flows also enables efficient sampling and exact density estimation. However, the quality of the samples generated by VAEs and Normalizing Flows may not be as high as those produced by other models. Diffusion Models present another approach, capable of covering diverse modes and generating high-quality samples. Recent advances have enabled diffusion models to produce samples that rival or even surpass GANs in terms of quality and diversity. However, diffusion models typically consist of a large number of denoising steps (often ranging from 500 to 1000), which require substantial computation to generate a single sample. This makes the sampling process slow and computationally expensive compared to other models.

As a result, understanding these trade-offs of using each class of DGM is essential when applying DGMs in practice. Since no current model perfectly satisfies all three key requirements, researchers should choose the model class that best aligns with the specific needs of their application, whether that be sample quality, diversity, or computational efficiency. Additionally, some previous research has explored combining different modeling approaches from several DGMs, such as \cite{xiao2021tackling} which combined Diffusion models with GAN, \cite{grover2018flow} which combined Normalizing Flows with GAN, and \cite{zhang2021diffusion} which combined Normalizing Flows with Diffusion model. These examples demonstrate that by integrating different DGMs, researchers can also address the limitations imposed by using a single DGM.

\section{Deep Generative Models in Different Areas of Transportation Research}\label{sec:dgmreview}

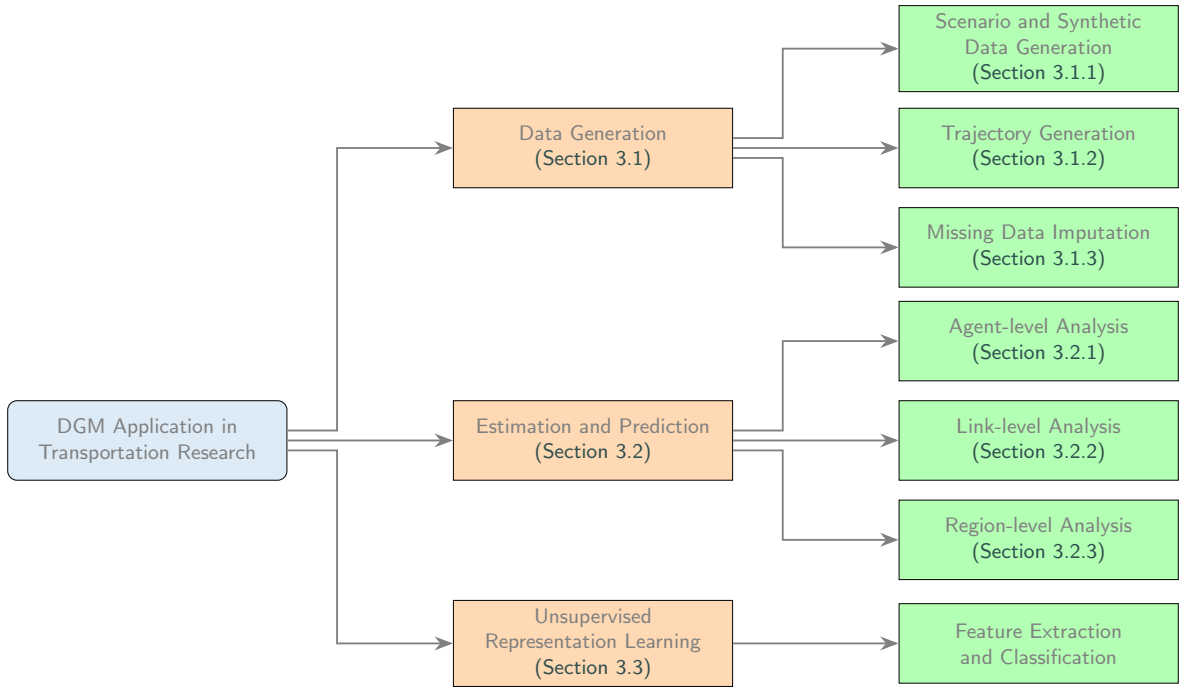
\begin{figure}[t]
    \centering
\resizebox{0.95\textwidth}{!}{
\begin{tikzpicture}[node distance=2.5cm and 2.5cm]
\node (start) [startstop, align=center, text width = 4cm] {DGM Application in \\ Transportation Research};
\node (process1) [process, above right=of start, yshift=0.7cm, align=center, text width = 4cm] {Data Generation \\ \hyperref[sec:3.1]{(Section 3.1)}};
\node (process2) [process, right=of start, align=center, text width = 4cm] {Estimation and Prediction \\ \hyperref[sec:3.2]{(Section 3.2)}};
\node (process3) [process, below right=of start, yshift=0.7cm, align=center, text width = 4cm] {Unsupervised \\Representation Learning \\ \hyperref[sec:3.3]{(Section 3.3)}};

\node (sub1) [subprocess, right=of process1, yshift=1.5cm, align=center, text width = 4cm] {Scenario and Synthetic \\ Data Generation \\ \hyperref[sec:3.1.1]{(Section 3.1.1)}};
\node (sub2) [subprocess, right=of process1, align=center, text width = 4cm] {Trajectory Generation\\ \hyperref[sec:3.1.2]{(Section 3.1.2)}};
\node (sub3) [subprocess, right=of process1, yshift=-1.5cm, align=center, text width = 4cm] {Missing Data Imputation\\ \hyperref[sec:3.1.3]{(Section 3.1.3)}};

\node (sub4) [subprocess, right=of process2, yshift=1.5cm, align=center, text width = 4cm] {Agent-level Analysis\\ \hyperref[sec:3.2.1]{(Section 3.2.1)}};
\node (sub5) [subprocess, right=of process2, align=center, text width = 4cm] {Link-level Analysis\\ \hyperref[sec:3.2.2]{(Section 3.2.2)}};
\node (sub6) [subprocess, right=of process2, yshift=-1.5cm, align=center, text width = 4cm] {Region-level Analysis\\ \hyperref[sec:3.2.3]{(Section 3.2.3)}};

\node (sub7) [subprocess, right=of process3, align=center, text width = 4cm] {Feature Extraction and Classification};

\draw [arrow] (start.east) ++(0,+0.15cm) -- ++(0.75cm,0) |- (process1.west);
\draw [arrow] (start.east) -- ++(0.75cm,0) |- (process2.west);
\draw [arrow] (start.east) ++(0,-0.15cm) -- ++(0.75cm,0) |- (process3.west);

\draw [arrow] (process1.east) ++(0,+0.15cm) -- ++(0.75cm,0) |- (sub1.west);
\draw [arrow] (process1.east) -- ++(0.75cm,0) |- (sub2.west);
\draw [arrow] (process1.east) ++(0,-0.15cm) -- ++(0.75cm,0) |- (sub3.west);

\draw [arrow] (process2.east) ++(0,+0.15cm) -- ++(0.75cm,0) |- (sub4.west);
\draw [arrow] (process2.east) -- ++(0.75cm,0) |- (sub5.west);
\draw [arrow] (process2.east) ++(0,-0.15cm) -- ++(0.75cm,0) |- (sub6.west);

\draw [arrow] (process3.east) -- (sub7.west);

\end{tikzpicture}
}
\caption{An overview of DGMs in transportation research.}
\label{fig:sec3overview}
\end{figure}

DGMs have become important in current research for their ability to model complex data distributions and generate data samples that closely mimic real-world observations. Particularly, in transportation research, DGMs offer tools to simulate, predict, and optimize transportation systems effectively
\citep{huynh2021optimal,lin2023generative,yan2023survey}.
Therefore, in this section, we provide a systematic review of the applications of DGMs in transportation research. 

This review focuses on three main areas where DGMs have significantly impacted transportation research: data generation, estimation and prediction, and unsupervised representation learning. 
%
%
By examining their applications in these areas, we gain a clear understanding of how DGMs impacted and will impact transportation research. An overview of this section is shown in Figure~\ref{fig:sec3overview}.

\begin{enumerate}
\item  \textbf{DGM for data generation}: The core functionality of DGMs lies in their ability to generate realistic synthetic data that closely replicate real-world conditions. This capability is particularly valuable in transportation research, where collecting large, high-quality datasets can be challenging and costly. Through the generation of diverse traffic scenarios, DGMs enable researchers to investigate situations that are difficult to capture through traditional data collection methods.

\item \textbf{DGM for estimation and prediction}: 
Beyond generating data, DGMs are also highly effective for estimation and prediction tasks. DGMs learn the underlying distribution of the given (training) data, and as a result, they can model the inherent uncertainties and variations in traffic data.
This capability allows DGMs to provide probabilistic estimations and forecasts of traffic states. The probabilistic approach enhances the accuracy, robustness, and reliability of predictions, which support better decision-making in transportation planning and operation.

\item \textbf{DGM for unsupervised representation learning}: 
Most DGMs are designed to learn the transformation function from a sample from a simple distribution (also known as a base distribution) to a target distribution. The `noisy' data sample from the base distribution can be considered as a \textit{latent representation} of the synthetic data sample.
These latent representations can be used for various analyses, such as classifying transportation modes, detecting traffic anomalies, or understanding driving behaviors. This ability to derive insights without requiring labeled data makes DGMs powerful tools in unsupervised learning scenarios.
\end{enumerate}

\subsection{DGM for Data Generation}\label{sec:3.1}

The rapid urbanization of society requires effective traffic management systems to reduce congestion, improve urban planning, and enhance transportation network efficiency \citep{xu2015estimation}. These systems are fundamentally built on large volumes of high-quality traffic data. Sufficient data volume is essential for accurate analyses and modeling of complex traffic behaviors and patterns, while data quality ensures precise and reliable insights and applications. Despite advancements in technology that have improved data collection capabilities in transportation, traffic sensors often face issues of sparsity and unreliability, making it challenging to obtain adequate, high-quality data in real-world scenarios \citep{chen2003detecting,ni2005markov}.
DGMs address these challenges by both increasing the volume and improving the quality of traffic datasets. They can generate synthetic data to supplement existing datasets, effectively expanding their size. Additionally, DGMs replicate real-world conditions with high fidelity, which enhances the dataset's qualitative aspects. These capabilities are particularly valuable for transportation research, where traditional data collection methods may be costly, time-consuming, and impractical, especially for capturing rare or non-reproducible scenarios. For example, synthetic data generated by DGMs can simulate a range of traffic conditions, from typical daily commutes to rare events like accidents or extreme weather. Consequently, the synthetic data from DGMs is crucial for developing robust traffic models and systems that can handle a wide range of scenarios.
In the following sections, we will explore data generation through DGMs in three key areas: scenario and synthetic data generation, trajectory generation, and missing data imputation.

\subsubsection{Scenario and Synthetic Data Generation}\label{sec:3.1.1}

Scenario and synthetic data generation involve using DGMs to produce a wide range of synthetic data. This process includes creating diverse driving scenarios to test and evaluate autonomous vehicles, synthesizing demographic and travel behavior data to represent different populations in studies, and generating synthetic traffic data for training other machine learning models for traffic operation and management. Additionally, DGMs can generate rare or unexpected traffic situations to improve anomaly detection and response strategies. These capabilities allow researchers to simulate various conditions and events, which can help enhance the robustness and accuracy of analysis. This section highlights how the DGMs enable comprehensive testing, analysis, and improvement of transportation networks through effective scenario and synthetic data generation. {We also categorize recent studies by generation type, task, model used, dataset, and evaluation metrics, as summarized in Table~\ref{tab:summary_scenario_generation}.}


\newcolumntype{C}[1]{>{\centering\arraybackslash}p{#1}}
\begin{table}[h]
\caption{A summary of recent studies in scenario and synthetic data generation using DGM}
\label{tab:summary_scenario_generation}
\centering
\resizebox{\textwidth}{!}{
\begin{tabular}{c|c|c|c|c|c}
\toprule
\toprule
\makecell{\textbf{Generation Type}} & 
\makecell{\textbf{Author}} & 
\makecell{\textbf{Task}} & 
\makecell{\textbf{Model}} & 
\makecell{\textbf{Dataset}} & 
\makecell{\textbf{Evaluation Metrics}}\\
\midrule
\midrule
\multirow{8}{*}{\makecell[l]{\textbf{Driving}\\\textbf{Scenario}\\\textbf{Generation}}}
  & \makecell{\cite{zhong2023guided}}
  & \makecell{Driving Scenarios}    
  & \makecell{Diffusion}     
  & \makecell{nuScenes} 
  & \makecell{Rule-specific Metrics, WD}
  \\\cmidrule{2-6}
  & \makecell{\cite{sun2023drivescenegen}}   
  & \makecell{Driving Scenarios}     
  & \makecell{DriveSceneGen\\(Diffusion-based)}     
  & \makecell{Waymo Motion Data}  
  & \makecell{FID, Average Feature Distance,\\Precision, Recall, F1-Score}
  \\\cmidrule{2-6}
  & \makecell{\cite{xu2023diffscene}}        
  & \makecell{Driving Scenarios}     
  & \makecell{DiffScene\\(Diffusion-based)}   
  & \makecell{CARLA Simulation}   
  & \makecell{Rule-specific Metrics, SSPD,\\FID, DTW, WD, KLD}
  \\\cmidrule{2-6}
  & \makecell{\cite{niedoba2024diffusion}}        
  & \makecell{Driving Scenarios}     
  & \makecell{DJINN\\(Diffusion-based)}     
  & \makecell{Argoverse, INTERACTION}   
  & \makecell{minADE\(_k\), minFDE\(_k\), MFD\(_k\),\\minSceneADE\(_k\), minSceneFDE\(_k\)}
  \\
\midrule\midrule
\multirow{18}{*}{\makecell[l]{\textbf{Population}\\\textbf{and}\\\textbf{Activity}\\\textbf{Synthesis}}}
  & \makecell{\cite{borysov2019generate}}        
  & \makecell{Population Synthesis}      
  & \makecell{VAE}             
  & \makecell{Danish National Travel Survey}    
  & \makecell{SRMSE, $R^2$, Diversity Measures,\\Distribution Comparison,\\Pearson's Correlation Coefficient}
  \\\cmidrule{2-6}
  & \makecell{\cite{garrido2020prediction}}      
  & \makecell{Population Synthesis}      
  & \makecell{VAE, WGAN}       
  & \makecell{Danish National Travel Survey}      
  & \makecell{SRMSE, $R^2$,\\Sampling Zeros, Structural Zeros,\\Pearson's Correlation Coefficient}
  \\\cmidrule{2-6}
  & \makecell{\cite{kim2023deep}}                
  & \makecell{Population Synthesis}      
  & \makecell{VAE, WGAN}       
  & \makecell{Korea Household Travel Survey}    
  & \makecell{SRMSE, Recall, Precision, F1-Score,\\Number of Unique Combinations}
  \\\cmidrule{2-6}
  & \makecell{\cite{jutras2024copula}}                
  & \makecell{Population Synthesis}      
  & \makecell{CTGAN, TVAE}      
  & \makecell{American Community Survey}
  & \makecell{SRMSE, Recall, Precision, F1-Score,\\Sampling Zeros, Structural Zeros}
  \\\cmidrule{2-6}
  & \makecell{\cite{jeong2021variational}}       
  & \makecell{Activity-based Modeling}      
  & \makecell{VAE}          
  & \makecell{Seoul Household Travel Survey}               
  & \makecell{Matching Rate, RMSE,\\Precision, Recall, F1-Score, Accscore}
  \\\cmidrule{2-6}
  & \makecell{\cite{badu2022composite}}          
  & \makecell{Activity-based Modeling}      
  & \makecell{CTGAN}        
  & \makecell{2013 Montreal OD Travel Survey}           
  & \makecell{SRMSE, $R^2$,\\Pearson's Correlation Coefficient,\\Distribution Comparison}
  \\\cmidrule{2-6}
  & \makecell{\cite{lee2025collaborative}}       
  & \makecell{Activity-based Modeling}      
  & \makecell{CollaGAN}    
  & \makecell{Seoul Household Travel Survey,\\Smart Card Data}  
  & \makecell{SRMSE, Feasibility, Heterogeneity,\\F1-Score, Distribution Comparison}
  \\
\midrule\midrule
\multirow{17}{*}{\makecell[l]{\textbf{Synthetic}\\\textbf{Data}\\\textbf{and}\\\textbf{Model}\\\textbf{Training}}}
  & \makecell{\cite{zhou2023towards}}                
  & \makecell{Urban Flow Generation}
  & \makecell{KSTDiff}
  & \makecell{Urban Flow Data from 4 Cities \\(Washington D.C., NYC, \\ Beijing, Baltimore)}  
  & \makecell{MAE, RMSE, SMAPE,\\MMD, NRMSE, JSD,\\Pearson's Correlation Coefficient}
  \\\cmidrule{2-6}
  & \makecell{\cite{rong2023complexity}}                
  & \makecell{OD Generation}
  & \makecell{DiffODGen}
  & \makecell{OD Data from 3 US City\\
  (Cook County, New York City, Seattle)}
  & \makecell{RMSE, NRMSE, CPC, JSD,\\Rate of Nonzero Flows,\\Accuracy, 1-Recall, Precision}
  \\\cmidrule{2-6}
  & \makecell{\cite{rong2023origin}}      
  & \makecell{OD Generation}
  & \makecell{ODGN (cGAN)}
  & \makecell{OD Data from 8 US City \\(NYC, LA, Chicago, Houston, SF, \\ Seattle, Washington D.C., Memphis)}
  & \makecell{JSD, RMSE, CPC}
  \\\cmidrule{2-6}
  & \makecell{\cite{ozturk2020development}}       
  & \makecell{RL Agent Training}
  & \makecell{GAN}
  & \makecell{NGSIM}
  & \makecell{Crash Counts, Reward Comparison}
  \\\cmidrule{2-6}
  & \makecell{\cite{dewi2022synthetic}}       
  & \makecell{Recognition Models Training}
  & \makecell{DCGAN}
  & \makecell{Taiwan Prohibitory Signs Dataset}
  & \makecell{MSE, Structural Similarity Index,\\Intersection over Union, mAP,\\Detection time, Classification accuracy}
  \\\cmidrule{2-6}
  & \makecell{\cite{zhu2022spatiotemporal}}          
  & \makecell{Prediction Model Training}
  & \makecell{STDGAN}
  & \makecell{PEMSD4, PEMSD7}
  & \makecell{MAE, RMSE, MAPE}
  \\
\midrule\midrule
\multirow{10}{*}{\makecell[l]{\textbf{Anomaly}\\\textbf{Data}\\\textbf{Generation}}}
  & \makecell{\cite{cai2020real}}        
  & \makecell{Crash Data Augmentation}      
  & \makecell{DCGAN}             
  & \makecell{Orlando Expressway SR 408 Data}
  & \makecell{AUC, Sensitivity, Specificity}
  \\\cmidrule{2-6}
  & \makecell{\cite{chen2022efficient}}      
  & \makecell{Crash Data Augmentation}     
  & \makecell{GAM, cGAN,\\GDAGAN}     
  & \makecell{Taiwan Provincial\\Highway Accident Records}     
  & \makecell{Specificity, FPR, Precision,\\Recall, F1-Score, BA, G-mean}
  \\\cmidrule{2-6}
  & \makecell{\cite{chen2024novel}}                
  & \makecell{Crash Data Augmentation}      
  & \makecell{CTGAN-RU}      
  & \makecell{Washington State Crash Data}
  & \makecell{Specificity, Recall, G-mean,\\Distribution Consistency,\\Parameter Recovery}
  \\\cmidrule{2-6}
  & \makecell{\cite{huo2021text}}       
  & \makecell{Extreme Situations Generation}      
  & \makecell{T\textsuperscript{2}GAN}          
  & \makecell{Sina Weibo Traffic-related\\Text Data, Beijing Fourth\\Ring Road Passenger Flow Data} 
  & \makecell{MAE, RMSE}
  \\
\bottomrule
\bottomrule
\end{tabular}
}
\end{table}

\paragraph{Driving Scenario Generation}

Many studies in this section have focused on driving scenario generation, which involves creating diverse and realistic driving situations to test and evaluate autonomous vehicles and traffic management systems \citep{ghosh2016sad,yun2019data,jin2023surrealdriver,xu2023generative,huang2024versatile,10579044, jiang2024scenediffuser}. {
The use of DGMs is beneficial in this context, as they generate scenarios that overcome the limited variety of real-world data and eliminate the need for costly, time-intensive data collection, thereby enabling scalable testing frameworks.} Most of these studies utilize additional information as conditions for their generative models, such as car-following theory, safety critics, or physical restrictions, to generate more realistic and high-performing models.
\cite{zhong2023guided} presented a framework using a conditional diffusion model to generate realistic and controllable traffic simulations. The key benefit of the model is that it allows users to specify trajectory properties, such as reaching a goal or following speed limits while ensuring that the generated trajectories are physically feasible and realistic through the guidance of signal temporal logic rules. 
Similarly, \cite{sun2023drivescenegen} introduced a data-driven method DriverSceneGen for generating diverse and realistic driving scenarios to address data limitations. DriverSceneGen employed a diffusion model to create initial driving scenes in a rasterized Bird-Eye-View (BEV) format and used a simulation network to predict multiple future scenarios based on the initial setup. 
Furthermore, \cite{xu2023diffscene} developed DiffScene, a diffusion-based framework designed to generate safety-critical driving scenarios for evaluating and enhancing autonomous vehicle safety. DiffScene used a diffusion model to approximate the distribution of low-density areas in traffic data, creating realistic safety-critical situations. These scenarios were further optimized using guided adversarial objectives to maintain both realism and safety-criticality. 
\cite{niedoba2024diffusion} introduced DJINN, a diffusion-based generative model that simulates joint traffic scenarios by diffusing the trajectories of all agents in a scene simultaneously. Rather than predicting trajectories for each agent independently, DJINN models the full joint distribution conditioned on flexible observation masks—capturing past, present, or even future state information—and maps context. This approach enables the generation of diverse, realistic traffic scenes, including rare and safety-critical events, while also allowing for test-time guidance and scenario editing.
These studies have demonstrated the capability of using DGMs in driving scenario generation by providing more diverse and realistic scenarios for traffic management and Autonomous Vehicle development. 
{Despite these promising developments, evaluating the quality of generated scenarios remains an open problem, as current metrics often fail to fully capture both realism and safety-critical aspects. In particular, there is no appropriate metric to quantitatively compare the distribution of generated driving scenarios to the ground truth. Moreover, existing metrics typically assess only specific features—for instance, verifying whether scenarios obey traffic rules—without capturing the full spectrum of driving realism and safety. Furthermore, current  DGM-based frameworks often fail to support long-duration simulations, unlike traditional simulators, which poses an additional challenge. In addition, current studies sometimes fall short in replicating extreme scenarios—such as sudden braking or rapid lane changes—that are critical in highly dynamic, localized environments. This gap underscores the need for future research to develop comprehensive, unified evaluation metrics and more robust simulation platforms that can support extended runs, thereby facilitating rigorous and reproducible testing of various driving systems.}

%

\paragraph{ {Population and Activity Synthesis}}
Population synthesis is a critical component of transportation research that involves the generation of synthetic demographic and travel behavior data to accurately represent populations. 
{ The primary challenge is to capture the full joint distribution of the population while generating new, diverse samples that include rare or unobserved events. Accurately modeling this joint distribution becomes increasingly complex as the number of attributes grows. Simultaneously, during the sampling stage, the exponential increase in possible attribute combinations leads to the sampling zero problem, making it difficult to generate diverse samples that include rare or unobserved events. Recent advances in DGMs, such as VAEs and GANs, offer promising solutions by approximating high-dimensional joint distributions and ensuring sample diversity.  Several studies have demonstrated that these approaches can outperform traditional methods such as Gibbs sampling and Bayesian Networks\citep{borysov2018scalable,badu2020differentially,mensah2022robustness,johnsen2022population}. 
For instance, \cite{borysov2019generate} employed a VAE to learn latent representations of agents' attributes, thereby generating synthetic micro-agents that maintain the statistical properties of the original population.
Similarly, \cite{garrido2020prediction} compared VAE and WGAN methods for recovering sampling zeros in high-dimensional travel survey data and found that while WGANs can achieve higher predictive accuracy, VAEs tend to generate more diverse populations. 
Correspondingly, \cite{kim2023deep} introduced custom loss functions for both VAE and WGAN to better handle structural-zeros issues, demonstrating improvements in both feasibility and diversity on a large-scale South Korean Household Travel Survey (HTS).
Recently, \cite{jutras2024copula} introduced a copula-based generative framework that generates synthetic data for a target population using only known marginal totals with a sample from a similar population. The proposed method combines copula theory with DGMs (Conditional Tabular GAN (CTGAN) and Tabular VAE (TVAE)) to decouple dependency structures from marginal distributions, ensuring that the synthesized data accurately reflects both the structural relationships and the target population's marginals.
Overall, these advances highlight the versatility of DGMs in addressing the complexities of population synthesis, balancing between achieving predictive accuracy and generating diverse, representative samples.

While synthetic populations form a strong base for transportation analysis, capturing dynamic human mobility requires Activity-Based Modeling (ABM) to simulate sequences of individual activities and predict their interconnections \citep{liao2024deep}. The challenge in ABM lies in collecting sufficient data to represent the joint distribution of individual attributes—such as income, job type, gender, and age—which becomes even more complex when incorporating spatial dimensions. Adding spatial attributes increases the likelihood of rare combinations, thereby exacerbating the sampling zero problem. Recent advances in DGMs have been applied to address these challenges \citep{kim2022imputing,jeong2021variational,badu2022composite,lee2025collaborative}.
For example, \cite{jeong2021variational} integrated a VAE with a Hidden Markov Model (HMM) to infer and synthesize human activity sequences from mobile data. By embedding high-dimensional, mixed discrete, and continuous features into a lower-dimensional latent space, the proposed VAE-HMM mitigates the challenges of inconsistent clustering and overfitting commonly encountered in traditional HMMs. 
In another study, \cite{badu2022composite} proposed a dual-GAN structure, Composite Travel GAN (CTGAN), which simultaneously generates socioeconomic attributes and sequential mobility data learned from the joint distribution of both tabular attributes and sequential trip chain locations data.
More recently, \cite{lee2025collaborative} proposed Collaborative GAN (CollaGAN), a GAN-based data fusion method that combines household travel surveys with smart card data. CollaGAN utilizes a semi-supervised variational embedding to harmonize the two datasets within a shared, low-dimensional latent space, employing multiple loss functions—including boundary loss and expert-designed constraints—and dual discriminators to enhance both the feasibility and heterogeneity of the generated activity schedules. 
These advances demonstrate the potential of DGMs to overcome the challenges of data sparsity and high-dimensional complexity in ABM, ultimately improving the accuracy and diversity of synthesized activity patterns.

However, current DGMs for population and activity synthesis face several challenges. First, these models often suffer from a lack of interpretability, making it difficult to understand the underlying decision processes. Additionally, the characteristics of HTS including a relatively small data size compared to the number of attributes, infrequent updates, and lengthy collection durations, limit the effective utilization of these models. Moreover, the sequential nature of the generated outputs can lead to logically inconsistent travel patterns—such as implausible transitions between transportation modes or mismatches between socioeconomic status and residential characteristics—that necessitate extensive post-processing corrections. Addressing these gaps is essential for advancing DGMs toward more realistic, transparent and policy-relevant transportation models.}

\paragraph{Synthetic Data and Model Training}
The DGM can also be used to generate various types of synthetic traffic data for traffic modeling and training purposes. These synthetic datasets enable researchers to simulate traffic scenarios and train models without relying solely on real-world data, which can be expensive and difficult to collect. In related research, instead of using basic models, most researchers combine various deep learning architectures or incorporate additional information to further improve performance \citep{chen2019traffic,zhang2020cgail,ozturk2020development,wu2020spatiotemporal,chen2021traffic,zhou2023towards,rong2023origin,kumar2023generative,devadhas2023generative}.
For example, \cite{rong2023complexity} introduced a large-scale OD generation method using a graph-denoising diffusion model. This approach simplifies the generation process by decomposing it into two stages: first, generating the network topology, and second, generating the edge weights. By tackling these stages separately, the model accurately generates realistic large-scale OD data.
In a similar way, \cite{rong2023origin} introduced an Origin-Destination Generation Networks
(ODGN), a physics-informed machine learning framework for generating OD by combining Multi-view Graph Attention Networks (MGAT) to extract urban features and a gravity-guided predictor to estimate mobility flows between regions. The model employs a cGAN training strategy with a random walk sampling discriminator to capture overall network topology, producing OD networks that closely match real-world properties.
In addition, \cite{zhou2023towards} introduced KSTDiff, a Knowledge-enhanced Spatio-temporal Diffusion model that generates dynamic urban flow data for regions lacking historical records by leveraging an Urban Knowledge Graph (UKG) and a region-customized diffusion process guided by a learnable volume estimator. The framework can also be adapted for urban flow prediction, achieving competitive performance compared to state-of-the-art methods.

For model training, \cite{ozturk2020development} introduced a GAN-based traffic simulator that generates realistic and stochastic vehicle trajectories from real data, which are then used to train reinforcement learning agents. These generated data help RL agents train in a more realistic environment and significantly improve their generalization capabilities compared to agents trained on simple rule-based simulators.
\cite{dewi2022synthetic} applied a Deep Convolutional GAN (DCGAN) to generate synthetic traffic sign images. This approach captures the distribution of real traffic sign images and produces new samples to augment the original dataset. The researcher further combined these synthetic images with real ones to improve the performance of deep recognition models, such as CNN and ResNet 50.
Additionally, \cite{zhu2022spatiotemporal} employed the Spatio-Temporal Dependencies GAN (STDGAN) to generate high-quality synthetic traffic volume data for prediction tasks. Specifically, STDGAN captures implicit variation patterns in traffic volume based on information from traffic speed and occupancy data. Experiment results indicate that enriching the training dataset with the synthetic volume data led to more robust and accurate performance in prediction models.

The provided examples demonstrate the ability of DGMs to generate various traffic data that support traffic management and enhance model training, offering valuable resources for transportation research and applications. {One key limitation of using DGMs is ensuring that the synthetic data accurately reflects the complex, dynamic, real-world environment. In general, the generated data fails to fully capture the intricate patterns, dynamic shifts, and non-stationary behaviors seen in real scenarios, which can introduce biases into the model training process. This limitation not only affects the quality of the synthetic data but also complicates its integration with real data. Determining the optimal balance is another critical issue in model training. Specifically, an excess of synthetic data may overwhelm the true data distribution, while too little might not adequately address data scarcity. In summary, these challenges can hinder the model's ability to generalize and perform effectively under dynamic conditions.}

\paragraph{Anomaly Data Generation}
{
Anomaly data generation is crucial for creating rare or unexpected traffic situations, which are essential for testing and evaluating systems under extreme conditions. Generating such data helps develop robust models that can effectively handle imbalanced datasets. In this research field, DGMs have been highly adapted to augment imbalanced traffic data. This augmentation not only addresses the scarcity of anomalous events in real-world datasets but also plays a pivotal role in enhancing the accuracy and resilience of prediction and classification models used in traffic safety and management applications.}\citep{islam2021crash,zarei2021crash,li2024difftad,chen2024novel}. 
For instance, \cite{cai2020real} utilized a DCGAN to generate synthetic crash data from learning the distribution of crash-related traffic data. The synthetic data was then combined with real data to create a balanced dataset, which was used to train various crash prediction algorithms, including Logistic Regression, SVM, ANN, and CNN.
Similarly, \cite{chen2022efficient} compared several data augmentation methods including Synthetic Minority Oversampling Technique (SMOTE), GAN, conditional GAN (cGAN), and Gaussian Discriminant Analysis GAN (GDAGAN) to address the imbalance in traffic collision datasets. These models generated additional samples for underrepresented classes, thereby balancing the traffic collision dataset, and enhancing the performance of classifiers trained on them.  
Furthermore, \cite{chen2024novel} developed a method that integrates Conditional Tabular GAN  with Random Under-sampling (CTGAN-RU) to generate synthetic crash data. This approach accounts for both the continuous and discrete characteristics of imbalanced crash datasets by incorporating various conditions, thereby creating a more balanced training set and improving the accuracy and reliability of crash severity prediction models.
\cite{huo2021text} introduced the Text-to-Traffic Generative Adversarial Network (T\textsuperscript{2}GAN), a novel framework that generates realistic traffic situations by fusing traditional traffic data with semantic information extracted from social media. In T\textsuperscript{2}GAN, a pre-trained GloVe model encodes the traffic-related text into semantic features that are integrated into the GAN training process, with a global-local loss employed to align the modalities and produce varied, contextually accurate traffic scenarios.
{ The current studies underscore the importance of using DGMs for generating anomaly data to address imbalanced datasets and enhance the robustness of traffic management and safety systems. However, current models face significant challenges. In particular, they struggle to effectively manage imbalanced crash data that often includes missing values, heteroscedasticity, noise, and small sample sizes—characteristics common in real-world datasets. This highlights the need for further research to refine these methods and ensure that synthetic data generation more accurately reflects the complexities of actual traffic anomalies.}

\subsubsection{Trajectory Generation}\label{sec:3.1.2}
Trajectory generation creates realistic movement patterns for vehicles and pedestrians, which is essential for developing accurate traffic simulations and urban planning tools. By leveraging DGMs, researchers can produce synthetic trajectories that mimic real-world behaviors of traffic agents, enabling more robust analysis and testing of transportation systems. These generated trajectories help understand traffic dynamics, evaluate transportation policies, and enhance infrastructure design. These trajectories can be further classified based on their scale and scope, such as micro and macro scales. Micro-scale trajectories focus on detailed, short-term movement patterns of individual agents in localized areas, like intersections, while macro-scale trajectories represent broader, long-term traffic flow across extensive road networks or regions. The classification into micro and macro scales significantly impacts their use cases, such as collision avoidance or urban planning, and determines the complexity of the problems they address. {Table~\ref{tab:summary_trajectory_generation} presents an overview of recent studies in trajectory generation, organized by generation scale, task, model employed, dataset, and evaluation metrics.}

\begin{table}[t]
\caption{A summary of recent studies in trajectory generation using DGM}
\label{tab:summary_trajectory_generation}
\centering
\resizebox{\textwidth}{!}{
\begin{tabular}{c|c|c|c|c|c}
\toprule
\toprule
\makecell{\textbf{Generation} \\ \textbf{Scale}} & 
\makecell{\textbf{Author}} & 
\makecell{\textbf{Task}} & 
\makecell{\textbf{Model}} & 
\makecell{\textbf{Dataset}} & 
\makecell{\textbf{Evaluation Metrics}} \\
\midrule
\midrule
\multirow{17}{*}{\makecell[l]{\textbf{Micro}}}
  & \makecell{\cite{krajewski2018data}}        
  & \makecell{GPS-level Generation\\(High Sampling)} 
  & \makecell{TrajGAN, TrajVAE} 
  & \makecell{HighD} 
  & \makecell{MSE} 
  \\\cmidrule{2-6}
  & \makecell{\cite{krajewski2019beziervae}} 
  & \makecell{GPS-level Generation\\(High Sampling)} 
  & \makecell{BézierVAE}
  & \makecell{HighD} 
  & \makecell{RMSE,\\ Normalized Variance-of-Differences}
  \\\cmidrule{2-6}
  & \makecell{\cite{ding2019multi}}      
  & \makecell{GPS-level Generation\\(High Sampling)}
  & \makecell{MTG\\($\beta$-VAE-based)}
  & \makecell{UMTRI Driving\\Encounter Dataset}
  & \makecell{Rule-specific Metrics,\\ Disentanglement Metric}
  \\\cmidrule{2-6}
  & \makecell{\cite{bhattacharyya2022modeling}}      
  & \makecell{GPS-level Generation\\(High Sampling)}
  & \makecell{GAIL}
  & \makecell{NGSIM}
  & \makecell{RMSE,\\ Rule-specific Metrics}
  \\\cmidrule{2-6}
  & \makecell{\cite{chen2022combining}}      
  & \makecell{GPS-level Generation\\(High Sampling)}
  & \makecell{GAIL}          
  & \makecell{HighD}     
  & \makecell{RMSE}  
  \\\cmidrule{2-6}
  & \makecell{\cite{ma2023physics}}      
  & \makecell{GPS-level Generation\\(High Sampling)}
  & \makecell{PICGAN}
  & \makecell{NGSIM}
  & \makecell{MSE,\\ Rule-specific Metrics}
  \\\cmidrule{2-6}
  & \makecell{\cite{yu2024theory}}      
  & \makecell{GPS-level Generation\\(High Sampling)}
  & \makecell{TDS-GAN}
  & \makecell{Closed Test Site from\\ Highway Research\\ Institute of the Ministry of\\ Transport of China}
  & \makecell{RMSE, MAPE, MAE,\\ Rule-specific Metrics}
  \\
\midrule
\midrule
\multirow{13}{*}{\makecell[l]{\textbf{Macro}}}
  & \makecell{\cite{choi2021trajgail}}        
  & \makecell{Link-level Generation}     
  & \makecell{TrajGAIL}
  & \makecell{Aimsun Simulation,\\ Seoul Taxi DTG}
  & \makecell{BLEU, METEOR, JSD}
  \\\cmidrule{2-6}
  & \makecell{\cite{sun2023toward}}      
  & \makecell{Link-level Generation} 
  & \makecell{MAGAIL-VL}
  & \makecell{pNEUMA}
  & \makecell{BLEU, JSD, MAPE, NMAE}
  \\\cmidrule{2-6}
  & \makecell{\cite{xiong2023trajsgan}}                
  & \makecell{Grid-level Generation}  
  & \makecell{TrajSGAN}
  & \makecell{MTL Trajet Dataset}
  & \makecell{JSD-based Metrics,\\ Recall, $R^2$}
  \\\cmidrule{2-6}
  & \makecell{\cite{zhu2024difftraj}}       
  & \makecell{GPS-level Generation\\(Low Sampling)}
  & \makecell{DiffTraj}
  & \makecell{Chengdu \& Xi'an\\GPS Trajectory Dataset}
  & \makecell{Density Error (JSD-based),\\ Trip Error (JSD-based),\\ Length Error,\\ Pattern Score (F1-Score-based),\\ MAE, MSE, RMSE} 
  \\\cmidrule{2-6}
  & \makecell{\cite{wei2024diff}}          
  & \makecell{Link-level Generation}
  & \makecell{Diff-RNTraj}
  & \makecell{Porto \& Chengdu\\Taxi Datasets}
  & \makecell{JSD,\\ Road Segment Connectivity}
  \\
\bottomrule
\bottomrule
\end{tabular}
}
\end{table}

\paragraph{Micro-scale Trajectory Generation}
{ Micro-scale trajectory generation involves producing detailed agent movements, closely focusing on behaviors within localized areas like intersections or specific road segments. It focuses on generating detailed vehicle dynamics, such as acceleration, deceleration, and lane-changing. The DGMs are particularly effective for this task as they efficiently learn complex and varied behaviors directly from real-world data, producing realistic trajectories. These accurate and detailed trajectories significantly enhance applications such as collision avoidance, precise maneuver planning, and road safety \citep{kuefler2017imitating, krajewski2018data, ding2019multi,krajewski2019beziervae, demetriou2020generation, zhou2020modeling, zhang2022generative, gong2023interactive,dong2023transfusor, singh2023bi, ma2023application,shi2024generative}.}
For example, \cite{krajewski2018data} used two DGMS, Trajectory GAN(TraGAN) and VAE(TraVAE), which generate realistic lane change trajectories by learning intuitive, disentangled latent parameters from real-world driving data. These models can accurately reproduce observed maneuvers while also synthesizing new trajectories to fill gaps in simulation datasets.
To further solve the smoothness problem, \cite{krajewski2019beziervae} integrated a Bézier-curve output layer with additional loss terms into TraVAE, resulting in BézierVAE. This model generates smooth trajectories in both the position and speed domains, enhancing trajectory modeling for safety validation of highly automated vehicles.
\cite{ding2019multi} introduced a Multi-Vehicle Trajectory Generator (MTG) that integrates a VAE framework with bi-directional Gate Recurrent Units (GRUs) in the encoder and a multi-branch GRU decoder to simulate realistic vehicle-to-vehicle encounters. The authors also propose a novel disentanglement metric to assess the model's stability and interpretability.
In another study, \cite{bhattacharyya2022modeling} employed GAIL with several extensions (PS-GAIL, RAIL, and Burn-InfoGAIL) to model and replicate human driving behavior in simulation. By addressing the challenges of stochasticity, multimodality, and latent variability in human driving behavior, the proposed methods generate realistic car-following and lane-changing behaviors, as validated on the NGSIM dataset.
\cite{chen2022combining} developed a two-step hybrid driving model that combines model-based controllers with GAIL for traffic simulation. The model generates high-level driver traits used as parameters for low-level model-based controllers to simulate human-like driving in multi-agent traffic scenarios. 
\cite{ma2023physics} developed the Physics-Informed Conditional GAN (PICGAN) to control Connected Autonomous Vehicles (CAVs) in mixed traffic environments. The proposed model integrates theoretical physics with a dual conditional GAN framework, which consists of an encoder-decoder generator and two discriminators. By training on both observed and simulated data, PICGAN demonstrates improved robustness and adaptability.
Similarly, \cite{yu2024theory} proposed a Theory-data Dual Driven Stochastically GAN (TDS-GAN) by integrating a physics-informed GAN with a two-dimensional Intelligent Driver Model (2D-IDM) to capture the stochastic and heterogeneous car-following behavior of human-driven vehicles in mixed traffic. In this framework, the GAN captures the inherent randomness and uncertainty of car-following, generating realistic, varied trajectories that enhance the predictive accuracy of the theory-driven 2D-IDM for short-term forecasts and macroscopic traffic flow simulations.
{DGMs have demonstrated an ability to produce highly detailed and realistic micro-scale trajectories, which are crucial for applications that demand precise and immediate responses in traffic systems. Despite these promising results, several limitations remain to be solved. First, fine-tuning the guidance parameters of DGMs is often sensitive and requires extensive calibration to balance between trajectory diversity and fidelity. Specifically, even minor deviations can significantly affect the model's ability to replicate rare or extreme driving events, such as sudden braking or abrupt lane changes, which are critical for safety. Moreover, the performance of DGMs is highly dependent on the quality and representativeness of the training data and any deficiencies in data coverage can compromise the model's accuracy and its generalizability to unseen scenarios. }

\paragraph{Macro-scale Trajectory Generation}
Macro-scale trajectory generation covers broad spatial and temporal scopes, including extensive areas such as entire road networks or large regions over longer periods. Such trajectories are essential for strategic planning, traffic management, and long-term behavioral analysis. By examining OD pairs, travel time distributions, and overall traffic flow patterns, macro-scale trajectories provide valuable insights into larger trends and systemic issues. The information is crucial for urban planning, optimizing public transportation systems, and enhancing the efficiency of traffic networks. 
Numerous studies have applied DGMs to generate such macro-scale trajectories, offering valuable insights for strategic decision-making in transportation engineering.\citep{chen2021traffic,chen2021trajvae,zhang2022factorized,zhu2024difftraj,wei2024diff,wang2024dptraj,zhu2024controltraj}.
For example, \cite{choi2021trajgail} developed TrajGAIL, a GAIL framework for generating link sequences of urban vehicle trajectories. The model combines the capabilities of GAIL with RNNs to learn the underlying distributions of urban vehicle trajectory data. TrajGAIL allows for the generation of synthetic vehicle trajectories that closely resemble real-world patterns, even from limited observations. 
Similarly, \cite{sun2023toward} introduced MAGAIL-VL, a data-driven, network-wide traffic simulation framework considering both Vehicle and Link agents. Specifically, MAGAIL-VL learns the distribution of vehicle movements and link states from observed data, enabling the generation of realistic traffic scenarios across an entire network. 
In another study, \cite{xiong2023trajsgan} introduced TrajSGAN, a semantic-guiding adversarial network for urban trajectory generation that synthesizes human mobility trajectories at a grid scale. It integrates an attention-based generator for trajectory location prediction with a rollout module and a CNN-based discriminator to assess the spatial structure of the generated trajectories. This integrated framework effectively reduces divergence in spatial metrics and has been successfully applied to epidemic diffusion studies with high accuracy.
Recently, diffusion-based models have gained popularity for generating fine-level urban trajectories. 
Recently, diffusion-based models have become popular for generating fine-level urban trajectories\citep{zhu2024difftraj,wei2024diff,wang2024dptraj,zhu2024controltraj}. For example, \cite{zhu2024difftraj} introduced DiffTraj, a novel approach for urban-scale trajectory generation that produces high-quality synthetic GPS trajectories using a spatial-temporal diffusion probabilistic model. DiffTraj leverages a U-Net architecture enhanced with residual blocks and multi-scale feature fusion to accurately estimate noise levels during the reverse denoising process. This design effectively captures complex spatial-temporal dependencies, enabling the generation of realistic synthetic trajectories that are well-suited for urban mobility analysis and preserving the statistical properties of real-world data.
Similarly, \cite{wei2024diff} presented Diff-RNTraj, a diffusion-based model designed to generate trajectories that are geographically accurate and adhere to road network constraints. The model uses a continuous diffusion framework combined with a pre-training strategy to handle the hybrid nature of road network–constrained trajectory data, resulting in realistic and usable synthetic trajectories.
{ Even though the current studies have demonstrated significant advances in macro-scale trajectory generation, several challenges remain to be addressed. First, integrating external conditions—such as dynamic traffic states, diverse regional behaviors, and factors like departure times and local events—into the generative process is still limited, which restricts the models' ability to accurately reflect real-time traffic dynamics. Second, ensuring that generated trajectories are physically plausible and strictly adhere to road network connectivity is challenging, often resulting in unrealistic or disconnected routes. Third, capturing the inherent stochasticity and heterogeneity of human mobility remains a major hurdle. Finally, the absence of unified evaluation metrics further complicates objective assessment and comparison among models, making it difficult to benchmark progress in urban-level trajectory generation using DGMs.}

\subsubsection{Missing Data Imputation}\label{sec:3.1.3}

The missing data imputation in transportation aims to reconstruct original data $X$ from missing or corrupted data $\hat{X}$ with conditional information $\phi$ such as known historical data, geometry configuration, and external factors ($\hat{X} = f(X,\phi)$). This process inherently involves data generation, as it creates new values to fill gaps in existing datasets, thereby enhancing the volume and quality of data available for analysis. In traditional imputation methods, missing data is often challenging to reconstruct due to the complex spatio-temporal dependencies in traffic data, which limit the accuracy and robustness of simple statistical methods.
In recent years, DGMs have become powerful tools for handling the complexities of traffic data imputation. Their ability to model the underlying distribution of traffic data enables them to generate realistic and accurate reconstructions of missing entries across various scenarios. DGMs can also integrate a wide range of conditional information, including the spatio-temporal dependencies of traffic data, to enhance imputation performance \citep{zhang2024spatial,cai2023spatial,yang2021st,zhang2021missing,peng2023generative,shin2023missing,duan2024traffic,chen2022mtsvae}.
Traditionally, missing data has been categorized into three types: Missing Completely at Random (MCAR), Missing at Random (MAR), and Missing Not at Random (MNAR) \citep{lin2020missing,hasan2021missing}. However, the unique spatio-temporal characteristics of traffic data require a reevaluation of these categories. To address this, \cite{chan2023missing} proposed a new classification specifically for transportation research, dividing missing data into Fiber Missing Data, Block Missing Data, and Random Missing Data. In this paper, we adopt this updated categorization to better address the specific challenges of traffic data imputation. {Table~\ref{tab:summary_data_imputation} summarizes recent studies and classifies them according to missing type, imputation task, model, dataset, missing rate, and evaluation metrics.}

\begin{table}[h]
\centering
\caption{A summary of recent studies in missing data imputation using DGM}
\label{tab:summary_data_imputation}
\resizebox{\textwidth}{!}{
\begin{tabular}{c|c|c|c|c|c|c}
\toprule
\toprule
\makecell{\textbf{Missing Type}} & 
\makecell{\textbf{Author}} & 
\makecell{\textbf{Imputation Task}} & 
\makecell{\textbf{Model}} & 
\makecell{\textbf{Dataset}} & 
\makecell{\textbf{Missing Rate}} & 
\makecell{\textbf{Evaluation Metrics}} \\
\midrule
\midrule
\multirow[c]{9}{*}{\makecell{\textbf{Fiber}}}
& \makecell{\cite{han2020content}}   & \makecell{Speed}               & \makecell{CA-GAN}   & \makecell{PEMS}                               & \makecell{10--100\%} & \makecell{MRE,\\ Time Cost Comparison} \\\cmidrule{2-7}
& \makecell{\cite{xu2021traffic}}    & \makecell{Speed}               & \makecell{GA-GAN}   & \makecell{PEMS-BAY,\\ Seattle Dataset}       & \makecell{10--70\%}  & \makecell{MAE, RMSE,\\ MAPE, Residual Analysis} \\\cmidrule{2-7}
& \makecell{\cite{zhang2021gated}}   & \makecell{Lane-level Speed}    & \makecell{GaGAN}    & \makecell{Hangzhou Signalized\\ Road Data}     & \makecell{20--100\%} & \makecell{MAE, RMSE, MAPE,\\ Pearson's Correlation\\ Coefficient} \\\cmidrule{2-7}
& \makecell{\cite{liu2023pristi}}    & \makecell{Speed}               & \makecell{PriSTI\\ (Diffusion-based)} & \makecell{AQI-36, METR-LA,\\ PEMS-BAY} & \makecell{10--90\%}  & \makecell{MAE, MSE, CRPS} \\
\midrule
\midrule
\multirow[c]{7}{*}{\makecell{\textbf{Block}}}
& \makecell{\cite{li20183d}}         & \makecell{Flow}    & \makecell{3DConvGAN}   & \makecell{Beijing Taxi Dataset}        & \makecell{20--80\%}  & \makecell{RSE} \\\cmidrule{2-7}
& \makecell{\cite{zhang2021missing}}  & \makecell{Volume}  & \makecell{SA-GAIN (GAN-based)} & \makecell{Seattle I-5 Highway}   & \makecell{10--80\%}  & \makecell{MAE, MMD, RMSE} \\\cmidrule{2-7}
& \makecell{\cite{yuan2022stgan}}    & \makecell{Speed\\ Passenger Flow} & \makecell{STGAN}    & \makecell{Beijing Road Dataset\\ Beijing Subway Dataset} & \makecell{20--80\%}  & \makecell{RMSE, NMAE, MAE} \\\cmidrule{2-7}
& \makecell{\cite{hou2023missii}}    & \makecell{Flow}    & \makecell{MissII (GAN-based)}  & \makecell{Beijing Taxi Dataset}      & \makecell{20--60\%}  & \makecell{MAE, RMSE,\\ CosineSim} \\
\midrule
\midrule
\multirow[c]{11}{*}{\makecell{\textbf{Random}}}
& \makecell{\cite{chen2019traffic}}  & \makecell{Flow}      & \makecell{GAN}        & \makecell{PEMS}                       & \makecell{30--80\%}  & \makecell{MAE, RMSE, MRE} \\\cmidrule{2-7}
& \makecell{\cite{boquet2019missing}}& \makecell{Speed}     & \makecell{VAE}        & \makecell{PEMS}                       & \makecell{10--40\%}  & \makecell{RMSE, MAPE} \\\cmidrule{2-7}
& \makecell{\cite{shin2023missing}}  & \makecell{Speed}     & \makecell{AAE}        & \makecell{Korean ITS Data,\\ PEMSD7}     & \makecell{10--50\%}  & \makecell{RMSE, MAPE} \\\cmidrule{2-7}
& \makecell{\cite{yang2021st}}       & \makecell{In\&Outflow} & \makecell{ST-LBAGAN}  & \makecell{Beijing Taxi,\\ NYC Bike Datasets} & \makecell{10--60\%}  & \makecell{RMSE, MAE} \\\cmidrule{2-7}
& \makecell{\cite{li2023multistate}} & \makecell{Volume\\ Speed} & \makecell{TGAIN (GAN-based)}  & \makecell{I90 Dataset,\\ Changchun Speed Dataset}  & \makecell{10--90\%}  & \makecell{RMSE, MAPE} \\\cmidrule{2-7}
& \makecell{\cite{zheng2024recovering}} & \makecell{Speed}     & \makecell{DPRDDM\\ (Diffusion-based)}  & \makecell{Zen \& Beijing Traffic Data} & \makecell{10--30\%}  & \makecell{RMSE, MAPE} \\
\bottomrule
\bottomrule
\end{tabular}
}
\end{table}

\paragraph{Fiber Missing Data} 

Fiber Missing Data occurs when there is a sudden, temporary failure of data-acquisition devices, leading to gaps in data collection that can be short-term or long-term. Many advanced DGMs have been developed to address this issue effectively \citep{li20183d,zhang2021missing,zhang2021gated,huang2022data,yuan2022stgan,shen2022traffic,cai2023spatial,zhang2023sasdim,hou2023missii,cai2023spatial, li2024self,duan2024traffic,zhang2024spatial}.
For instance, \cite{han2020content} developed the Content-Aware GAN (CA-GAN) to handle missing traffic data in time series. In this method, traffic data were modeled as tensors to capture the inherent spatial-temporal correlations. The CA-GAN learns the underlying traffic distribution and generates realistic traffic patterns, effectively completing gaps in speed data, particularly in cases with consecutive data losses. 
Similarly, \cite{xu2021traffic} introduced the Graph Aggregate GAN (GA-GAN), which combines the strengths of Graph Sample and Aggregate (GraphSAGE) with a GAN to impute missing traffic speed data. In this approach, GraphSAGE aggregates information from neighboring nodes in the road network to capture spatial-temporal correlations, which are then used by a Wasserstein GAN (WGAN) to generate the missing data. The GA-GAN has been tested in both fiber and random missing data environments, demonstrating its versatility. 
\cite{zhang2021gated} proposed the Gated GAN (GaGAN) to address missing speed data on signalized roads. The GaGAN integrates attention mechanisms within graph convolutional operations to capture spatial correlations and enhances GRUs with Self-Attention (SA-GRU) to learn temporal dependencies across signal cycles. This approach is effective for both fiber and random missing data scenarios, ensuring that the imputed values retain the inherent spatio-temporal patterns observed in lane-level traffic measurements.
In another study, \cite{liu2023pristi} introduced PriSTI, a conditional diffusion framework for spatio-temporal data imputation. Specifically, the PriSTI uses a diffusion-based noise estimation module and a spatio-temporal feature extraction process to enhance its performance under high missing rates and diverse spatial configurations. It has been evaluated in environments with both fiber and random missing data, showing significant robustness. 
%
%
These models demonstrate the capabilities of DGMs in addressing the challenges posed by fiber missing data, ensuring the continuity and quality of traffic datasets even in the face of temporary data collection failures.

\paragraph{Block Missing Data}
Block Missing Data occurs when there are no data-acquisition detectors in the area of interest, leading to complete loss in the dataset for the region. This type of missing data is particularly challenging to address because of the absence of any observations over an extended spatial and temporal range. Advanced DGMs have been developed to tackle this challenge, though most studies do not focus exclusively on block missing data. Instead, they often evaluate model performance under various conditions, including fiber and random missing data scenarios \citep{li20183d,zhang2021missing,yuan2022stgan,hou2023missii,li2023dynamic,zhang2024spatial}.
For example, \cite{li20183d} combined a 3D Convolutional Neural Network with GAN (3DConvGAN) to address missing traffic data. Unlike traditional methods that may not fully exploit the spatial-temporal features of historical data, 3DConvGAN uses a fractional strided 3D CNN in both the generator and discriminator to enhance imputation performance. The model's effectiveness was also tested in environments with Fiber, Block, and Random Missing Data, highlighting its versatility.
In another study, \cite{zhang2021missing} introduced SA-GAIN, a Self-Attention Generative Adversarial Imputation Network, designed to impute missing traffic flow data.  By incorporating a self-attention mechanism in GAN, SA-GAIN effectively captures important features across spatial and temporal dimensions. The model was tested in both fiber and block missing data scenarios, demonstrating robust performance.
In \cite{yuan2022stgan}, the authors enhanced the performance of GAN in traffic flow imputation by designing generative and center losses. These losses enable the model to more accurately reconstruct missing data while preserving local spatio-temporal correlations. This approach was evaluated across all three types of missing data environments: Fiber, Block, and Random Missing Data. The proposed method demonstrated significant improvements in imputing missing traffic data under these conditions.
\cite{hou2023missii} developed MissII, a novel theory-guided deep learning framework for traffic data imputation. The approach first estimates traffic flow between points of interest using mobility models that leverage environmental and social factors to capture complex traffic dynamics. The estimated traffic flow data are then used as real samples to guide the GAN's training process to improve the imputation accuracy.
These studies illustrate the effectiveness of DGMs in addressing block missing data, which presents unique challenges due to the complete absence of data in specific areas. By leveraging spatio-temporal modeling and integrating advanced neural network architectures, these models provide comprehensive solutions to ensure continuity and quality in traffic data even when entire sections of data are missing.

\paragraph{Random Missing Data}
Random Missing Data occurs when the data-acquisition detectors fail unpredictably in the spatio-temporal domain, which results in gaps in data with little to no correlation among the missing entries. Due to its unpredictable nature, this scenario presents significant challenges and is commonly used to evaluate the performance of imputation models \citep{li20183d,chen2021learning,chen2022mtsvae,wu2022traffic, qin2021network,kazemi2021igani,tu2021incomplete,zhang2021gated, xu2021traffic,shen2022traffic,yuan2022stgan,yang2022st,zhang2022tsr,wang2023network,liu2023pristi,li2023dynamic,huang2023deep,zhang2023sasdim,cai2023spatial,hou2023missii,cai2023spatial,li2024self,zheng2024recovering,duan2024traffic,zhang2024effective,zhang2024spatial}. 
For example, \cite{chen2019traffic} developed a GAN-based method to impute missing traffic flow data by combining real and synthetic data. This approach utilizes a parallel data paradigm where the GAN generates synthetic data that are used alongside real data to train the imputation model, enhancing its ability to fill in gaps accurately.
\cite{boquet2019missing,boquet2020variational} used a VAE-based method for imputing missing traffic data, aimed at improving the accuracy of traffic forecasting systems impacted by sensor or system failures. The unsupervised approach learns the underlying data distribution from a latent space, resulting in improved post-imputation performance and effective data augmentation.
To fully exploit the benefits of both VAE and GAN in data imputation, \cite{shin2023missing} presented an Adversarial Autoencoder (AAE)-based model that leverages spatio-temporal feature extraction to address missing traffic data. AAE combines the principles of VAE and GAN to capture the complex dependencies within traffic data, providing robust imputation capabilities even in scenarios with high rates of missing data. 
\cite{yang2021st} introduced the Spatio-Temporal Learnable Bidirectional Attention GAN (ST-LBAGAN) that combines a U-Net generator with bidirectional attention to capture spatio-temporal dependencies for missing traffic data imputation. The model uses multi-channel inputs with a mask to focus on missing regions and is trained with a composite loss to ensure realistic outputs, achieving high accuracy even at high missing rates.
Additionally, \cite{li2023multistate} proposed the Time Series Generative Adversarial Imputation Network (TGAIN) to tackle the challenge of imputing time series data. This model is designed to learn the multi-state distribution of missing time series data under conditional vector constraints. By employing a multiple imputation strategy, it effectively handles the uncertainty inherent in the imputation process.
Recently, \cite{zheng2024recovering} used traffic domain knowledge and a denoising diffusion model to develop the Doubly Physics-Regularized Denoising Diffusion Model (DPRDDM) for recovering corrupted traffic speed data. This model demonstrates robustness in handling various types of noise, including Gaussian white noise, random corrupted noise, spatially correlated noise, temporally correlated noise, and a mixture of these noise types.
The current studies in random missing scenario illustrate the diverse applications and robust capabilities of DGMs in addressing the challenges posed by random missing data. By effectively capturing and utilizing spatio-temporal dependencies, these models ensure accurate and reliable data imputation even in the most unpredictable scenarios.

{DGMs show promising solutions for handling missing data in traffic systems; however, some key challenges still persist at this stage. A primary concern is limited transferability: models tend to perform well on their training datasets but struggle to generalize across diverse traffic conditions and sensor networks, limiting their practical applicability. Additionally, high missing rates and irregular patterns substantially affect imputation quality, compounded by the difficulty of capturing complex spatiotemporal dependencies in multidimensional traffic data. Furthermore, the lack of standardized evaluation protocols—such as consistent missing rate ranges, evaluation metrics, and unified datasets—makes it difficult to compare and validate different approaches effectively. These limitations highlight the need for further research to improve the robustness and adaptability of DGMs in addressing missing data in traffic systems.}

\subsection{DGM for Estimation and Prediction} \label{sec:3.2}

One of the significant aspects of DGMs in transportation studies is their capability to learn and accurately model the distribution of input data. This foundational property enables DGMs to handle complex datasets and is critical for advanced analytical tasks such as traffic data generation, prediction, and classification. Another critical yet less explored advantage of DGMs is their ability to perform density estimation—modeling the probability distribution of potential outcomes based on input data. This capability is a key function in predictive analytics and estimation tasks within transportation systems. Specifically, ``prediction'' refers to forecasting future traffic conditions using historical and real-time data, while ``estimation'' focuses on determining the current status of the system, which often requires filling in data gaps or refining data accuracy. The ability to generate and evaluate multiple potential scenarios from DGMs is invaluable for effective traffic management. It allows transportation planners and engineers to anticipate and respond to a range of possible current and future conditions, thereby improving the efficiency and safety of the traffic network. By providing a probabilistic view of potential outcomes, DGMs empower decision-makers to accommodate the inherent variability and dynamic nature of traffic flows, significantly enhancing both the precision and reliability of transportation systems planning.

Building on the foundational understanding of DGMs and their application in traffic estimation and prediction, the following sections will discuss current research from three distinct scale levels: agent, link, and region. At the agent level, we will explore how these models are applied in individual vehicle and pedestrian prediction and estimation, critical for autonomous driving and safety systems. At the link level, the focus will shift to how DGMs are used for modeling traffic flows and speed on specific roadway segments, which is crucial for daily traffic management and dynamic routing. Finally, at the regional level, we will discuss the role of DGMs in shaping strategic planning and operational decisions by analyzing and predicting wide-area traffic patterns. Through this comprehensive examination, we aim to highlight the versatility and impact of DGMs in addressing the complexities of modern transportation challenges, thereby showcasing their essential role in the evolution of smart mobility solutions.

\subsubsection{Agent-level Analysis} \label{sec:3.2.1}
{ Agent-level analysis focuses on understanding and predicting the behaviors of individuals, such as vehicles, cyclists, or pedestrians. This detailed analysis is crucial for various applications like autonomous vehicle navigation, pedestrian safety systems, and dynamic routing. DGMs are particularly critical in this context since they can learn complex agent behavior patterns directly from data while inherently accounting for uncertainty in the estimation or prediction process. In other words, DGMs can reflect the inherent variability of real-world actions and results in more robust and reliable analysis. Table~\ref{tab:summary_agent_level} summarizes recent studies, categorizing them by model, observation/prediction length, sampling time, prediction samples, dataset, and evaluation metrics.} 

\begin{table}[h]
\caption{A summary of recent studies in agent-level analysis using DGM}
\label{tab:summary_agent_level}
\centering
\resizebox{\textwidth}{!}{
\begin{tabular}{c|c|c|c|c|c|>{\centering\arraybackslash}p{2cm}}
\toprule
\toprule
\makecell{\textbf{Author}} & 
\makecell{\textbf{Model}} & 
\makecell{\textbf{Observation} / \\ \textbf{Prediction} \\ \textbf{Length}} & 
\makecell{\textbf{Sampling} \\ \textbf{Time}} & 
\makecell{\textbf{Prediction} \\ \textbf{Samples}} & 
\makecell{\textbf{Dataset}} & 
\makecell{\textbf{Evaluation} \\ \textbf{Metrics}}\\
\midrule
\midrule
\makecell{\cite{li2019conditional}}       
& \makecell{GAN + cVAE}
& \makecell{2s/5s \\ 3.2s/4.8s}
& \makecell{0.5s \\ 0.4s}
& \makecell{1}
& \makecell{INTERACTION \\ ETH, UCY, Stanford Drone Dataset}
& \makecell{ADE, FDE}
\\\midrule
\makecell{\cite{kim2021driving}}       
& \makecell{cVAE}
& \makecell{1s/1s}
& \makecell{0.1s}
& \makecell{1}
& \makecell{CarMaker HILS}
& \makecell{RMSE, MAE}
\\\midrule
\makecell{\cite{wang2020multi}}       
& \makecell{TS-GAN}
& \makecell{3s/1-5s}
& \makecell{0.1s}
& \makecell{1}
& \makecell{NGSIM}
& \makecell{RMSE}
\\\midrule
\makecell{\cite{gomez2022exploring}}      
& \makecell{GAN}
& \makecell{2s/3s}
& \makecell{0.1s}
& \makecell{1}
& \makecell{Argoverse} 
& \makecell{ADE, FDE}
\\\midrule
\makecell{\cite{chen2023equidiff}}  
& \makecell{EquiDiff}
& \makecell{3s/1-5s}
& \makecell{0.2s}
& \makecell{1}
& \makecell{NGSIM}
& \makecell{RMSE}
\\\midrule
\makecell{\cite{liu2024multi}}      
& \makecell{Diffusion}
& \makecell{3s/1-5s}
& \makecell{0.2s}
& \makecell{1}
& \makecell{NGSIM}
& \makecell{RMSE}
\\\midrule
\makecell{\cite{gupta2018social}}    
& \makecell{SocialGAN}
& \makecell{3.2s/4.8s}
& \makecell{0.4s}
& \makecell{1-100}
& \makecell{ETH, UCY}
& \makecell{minADE\(_k\), \\ minFDE\(_k\)}
\\\midrule
\makecell{\cite{kosaraju2019social}}   
& \makecell{Social-BiGAT \\ (GAN-based)}
& \makecell{3.2s/4.8s}
& \makecell{0.4s}
& \makecell{1-20}
& \makecell{ETH, UCY}
& \makecell{minADE\(_k\), \\ minFDE\(_k\)}
\\\midrule
\makecell{\cite{sun2021unified}}  
& \makecell{Flow}
& \makecell{3.2s/0.4-4.8s}
& \makecell{0.4s}
& \makecell{1-20}
& \makecell{ETH, UCY, \\ Stanford Drone Dataset}
& \makecell{minADE\(_k\), \\ minFDE\(_k\), \\ NLL}
\\\midrule
\makecell{\cite{salzmann2020trajectron++}}       
& \makecell{Trajectron++ \\ (VAE-based)}
& \makecell{3.2s/4.8s \\ 2s/1-4s}
& \makecell{0.4s \\ 0.5s}
& \makecell{20}
& \makecell{ETH, UCY \\ nuScenes}
& \makecell{minADE\(_k\), \\ minFDE\(_k\), \\ KDENLL\(_k\)}
\\\midrule
\makecell{\cite{vishnu2023improving}}      
& \makecell{TS-GAN, TS-CVAE}
& \makecell{3.2s/4.8s}
& \makecell{0.4s}
& \makecell{1-20}
& \makecell{EyeonTraffic, \\ INTERACTION}
& \makecell{minADE\(_k\), \\ minFDE\(_k\), \\ KDENLL\(_k\)}
\\\midrule
\makecell{\cite{li2023multi}}      
& \makecell{cVAE + Diffusion}
& \makecell{2s/6s \\ 2s/3s}
& \makecell{0.5s \\ 0.1s}
& \makecell{1-15}
& \makecell{nuScenes \\ Argoverse}
& \makecell{minADE\(_k\), \\ minFDE\(_k\)}
\\ 
\bottomrule
\bottomrule
\end{tabular}
}
\end{table}

Notably, a significant portion of research in agent-level traffic analysis using DGMs primarily emphasizes learning the underlying distributions of traffic data and produces single outcome\citep{roy2019vehicle,zhao2020novel, hegde2020vehicle,zhou2021sa,wang2021prediction,jagadish2021autonomous,jagadish2022conditional,chen2022cae,hsu2023deep,westny2024diffusion}. This learning process is essential because it allows models to capture the intricate patterns and correlations within data that traditional models might miss. 
For example, \cite{li2019conditional} proposed the Conditional Generative Neural System (CGNS), which combines the cVAE with GAN to generate feasible, realistic, and diverse future trajectories for multiple agents by leveraging both static context and dynamic interactions. 
\cite{kim2021driving} improved the accuracy of ego vehicle trajectory prediction using a driving style-based cVAE. Their model integrates a DeepConvLSTM network to recognize driving styles from in-vehicle sensor data. This approach not only predicts a single trajectory but also estimates a probability distribution over future trajectories, effectively capturing the inherent uncertainty in driving behavior.
Additionally, \cite{wang2020multi} presented a GAN-based framework TS-GAN for vehicle trajectory prediction that uses multi-vehicle collaborative learning. Their method combines an auto-encoder social convolution module to capture spatial interactions and a recurrent social mechanism to model temporal relationships among surrounding vehicles. These fused features feed into a conditional GAN that generates a multi-modal probability distribution over future trajectories for a target vehicle.
Similarly, \cite{gomez2022exploring} explored the use of attention mechanisms with GANs for vehicle trajectory prediction. The model generates feasible and realistic trajectories by considering both social interactions and the physical constraints of the road network. 
Recently, \cite{chen2023equidiff} introduced EquiDiff, a conditional equivariant diffusion model that combines a denoising diffusion probabilistic model with an SO(2)-equivariant transformer to effectively manage uncertainties in vehicle trajectory predictions
\cite{liu2024multi} aimed to predict the distribution of endpoints of multi-agent trajectories using denoising diffusion and Transformer models, capturing both the spatio-temporal dynamics and the intrinsic intent of vehicles to enhance overall prediction quality.

%
%
%


Another critical aspect of using DGMs in agent-level analysis is the ability to perform density estimation of outputs. This capability allows for generating multiple potential outcomes, offering a comprehensive probabilistic view that supports robust decision-making under uncertainty. Research has increasingly focused on this feature, enabling traffic managers and autonomous vehicle systems to adapt dynamically to varied and unpredictable conditions \citep{feng2019vehicle,zhao2019multi,bhattacharyya2019conditional, cheng2020context,neumeier2021variational,wang2020improving,eiffert2020probabilistic,agarwal2020imitative, dendorfer2021mg, rossi2021human,choi2021dsa,li2021vehicle,rossi2021vehicle,oh2022cvae,wu2022long,zhong2022stgm,guo2023map,xing2022multi,jagadish2022conditional,de2022vehicles,gui2022visual,yao2023graph,rempe2023trace,tang2024utilizing,yang2024wcdt}. 
For example, \cite{gupta2018social} proposed the Social GAN to predict socially acceptable future trajectories for pedestrians in crowded scenes. The GAN model captures complex dynamics and social behaviors, generating a wide range of potential trajectories. 
Extending this idea, \cite{kosaraju2019social} introduced Social-BiGAT, a trajectory forecasting model that integrates Bicycle-GAN with Graph Attention Networks (GATs) and LSTMs. This model generates multimodal predictions of pedestrian trajectories by leveraging complex social interactions and physical context cues.
Other studies have adopted different strategies to model uncertainty. \cite{sun2021unified} introduced a normalizing flow-based prediction model designed to model the exact probability distribution of future human trajectories. 
%
In a multi-agent context, \cite{salzmann2020trajectron++} developed Trajectron++, an advanced VAE-based model that forecasts the trajectories of multiple agents in dynamic environments. Trajectron++ integrates a CVAE with Gaussian Mixture Models (GMMs) within a graph-based recurrent neural network framework. This structure allows Trajectron++ to generate multiple potential trajectories in a probabilistic manner, providing a comprehensive representation of possible future scenarios.
\cite{vishnu2023improving} incorporated traffic states into Transformer, GAN, and CVAE models to enhance multi-agent trajectory predictions. These models generate diverse plausible trajectory outcomes that are contextually relevant and highly predictive. 
Recent work has further explored hybrid approaches. \cite{li2023multi} developed a framework combining cVAE and conditional diffusion models for multi-modal vehicle trajectory prediction. This approach addresses the challenge of predicting highly uncertain future vehicle trajectories in urban environments by first generating trajectories with cVAE and refining them using a diffusion model.


%

%
{ These examples underscore the diverse and robust capabilities of DGMs in agent-level analysis, which are crucial for intelligent transportation systems. However, as models incorporate richer contextual data and accommodate a variable number of agents, their computational complexity increases, posing challenges for real-time scalability. Moreover, effectively integrating complex dynamics and heterogeneous contextual cues remains a significant difficulty. }

\subsubsection{Link-level Analysis} \label{sec:3.2.2}

Effective traffic management at the link level is essential for maintaining smooth traffic flow and ensuring safety on individual road links within transportation networks. This process includes accurate predictions and estimations of traffic conditions such as flow, speed, and travel time. DGMs play a pivotal role at this scale by providing advanced insights that support real-time traffic management, routing optimization, and congestion control. By leveraging the power of DGMs, traffic managers can respond more effectively to changing conditions on each link and anticipate future variations with greater precision. {Table~\ref{tab:summary_link_level} provides an overview of recent studies, classifying them based on model type, observation/prediction length, sampling time, prediction samples, dataset, and evaluation metrics.}

\begin{table}[h]
    \caption{A summary of recent studies in link-level analysis using DGM}
    \label{tab:summary_link_level}
    \centering
    \resizebox{\textwidth}{!}{
    \begin{tabular}{c|c|c|c|c|c|c|c}
    \toprule
    \toprule
    \makecell{\textbf{Author}} & 
    \makecell{\textbf{Model}} & 
    \makecell{\textbf{Task}} & 
    \makecell{\textbf{Input} / \\ \textbf{Prediction} \\ \textbf{Length}} & 
    \makecell{\textbf{Sampling} \\ \textbf{Time}} & 
    \makecell{\textbf{Prediction} \\ \textbf{Samples}} & 
    \makecell{\textbf{Dataset}} & 
    \makecell{\textbf{Evaluation} \\ \textbf{Metrics}}\\
    \midrule
    \midrule
    \makecell{\cite{yu2019real}}       
    & \makecell{GCGA \\ (GAN-based)}
    & \makecell{Speed Estimation}
    & \makecell{-} 
    & \makecell{-}
    & \makecell{1}
    & \makecell{Cologne Speed Data} 
    & \makecell{MAPE}
    \\\midrule

    \makecell{\cite{li2019learning}}       
    & \makecell{DeepGTT \\ (VAE-based)}
    & \makecell{Travel Time \\ Distribution \\ Estimation}
    & \makecell{-} 
    & \makecell{-}
    & \makecell{1}
    & \makecell{Chinese Provincial Capital \\ City Taxi Dataset}
    & \makecell{RMSE, MAE}
    \\\midrule

    \makecell{\cite{xu2020ge}}       
    & \makecell{GE-GAN}
    & \makecell{Volume Estimation \\ Speed Estimation}
    & \makecell{-} 
    & \makecell{-}
    & \makecell{1}
    & \makecell{PEMS \\ Seattle Dataset}
    & \makecell{RMSE, \\ MAE, MAPE}
    \\\midrule

    \makecell{\cite{yu2020extracting}}       
    & \makecell{LSTM-CGAN}
    & \makecell{Taxi Hotspots \\ Prediction}
    & \makecell{60min/10min} 
    & \makecell{10min}
    & \makecell{1}
    & \makecell{Beijing Taxi trajectory \\ Taximeter Data}
    & \makecell{1-Recall, FPR, ROC, \\ Section Consistency}
    \\\midrule

    \makecell{\cite{zhou2020variational}}       
    & \makecell{VGRAN \\ (VAE + Flow)}
    & \makecell{Speed Prediction}
    & \makecell{60min/5-60min}
    & \makecell{5min}
    & \makecell{1}
    & \makecell{METR, PEMS}
    & \makecell{RMSE, \\ MAE, MAPE}
    \\\midrule

    \makecell{\cite{rasul2020multivariate}}       
    & \makecell{Flow}
    & \makecell{Flow Prediction \\ Volume Prediction}
    & \makecell{Flexible/24hr \\ Flexible/12hr}
    & \makecell{1hr \\ 0.5hr}
    & \makecell{100}
    & \makecell{PEMS-SF \\ NYC Taxi Data}
    & \makecell{CRPS\textsubscript{sum}, MSE}
    \\\midrule

    \makecell{\cite{rasul2021autoregressive}}       
    & \makecell{TimeGrad \\ (Diffusion-based)}
    & \makecell{Flow Prediction \\ Volume Prediction}
    & \makecell{Flexible/24hr \\ Flexible/12hr}
    & \makecell{1hr \\ 0.5hr}
    & \makecell{100}
    & \makecell{PEMS-SF \\ NYC Taxi Data}
    & \makecell{CRPS\textsubscript{sum}}
    \\\midrule

    \makecell{\cite{wen2023diffstg}}       
    & \makecell{DiffSTG \\ (Diffusion-based)}
    & \makecell{Flow Prediction}
    & \makecell{1hr/1hr}
    & \makecell{5min}
    & \makecell{100}
    & \makecell{PEMS08}
    & \makecell{CRPS, \\ RMSE, MAE}
    \\\midrule

    \makecell{\cite{feng2024latent}}       
    & \makecell{LDT \\ (Diffusion-based)}
    & \makecell{Flow Prediction \\ Volume Prediction}
    & \makecell{Flexible/24hr \\ Flexible/12hr}
    & \makecell{1hr \\ 0.5hr}
    & \makecell{100}
    & \makecell{PEMS-SF \\ NYC Taxi Data}
    & \makecell{CRPS\textsubscript{sum}, MSE}
    \\
    \bottomrule
    \bottomrule
    \end{tabular}
    }
\end{table}

As discussed earlier, one of the benefits of using DGMs in estimation and prediction tasks is their ability to learn the characteristics and distributions from training data, enabling them to generate realistic and reliable data for effective traffic decision-making \citep{lin2018pattern,chen2018traffic,zhang2019gcgan,zhang2019trafficgan,impedovo2019trafficwave,zang2019traffic,xu2020road,li2020spatial,zhou2020variational,yu2020forecasting, aibin2021short,sun2021traffic,zhang2021satp,song2021learn,jin2022gan, wang2022traffic,zhao2022graphsage,mo2022quantifying, xu2022gats,khaled2022tfgan}.
For example, \cite{yu2019real} proposed a deep neural network architecture called the Graph Convolutional Generative Autoencoder (GCGA) to address real-time speed estimation problems. In GCGA, the Graph Convolutional Network (GCN) extracts spatial characteristics, and an autoencoder-based GAN uses these features to generate traffic speed maps from incomplete data. 
%
Similarly, \cite{li2019learning} presented DeepGTT, a VAE-based model designed to predict travel time distributions by integrating real-time traffic data and spatial features. The VAE enables the model to generate realistic travel time distributions even under conditions of data sparsity and variability.
In another study, \cite{xu2020ge} developed the Graph Embedding GAN (GE-GAN), which selects adjacent links to estimate road traffic speed and volume more accurately. To be more specific, it uses Graph Embedding (GE) techniques to represent the spatial characteristics of road networks, while a Wasserstein GAN (WGAN) is employed to generate traffic state data.
%
%
\cite{yu2020extracting} introduced a framework combining clustering and generative models to predict taxi hotspots. This model LSTM-CGAN, which integrates LSTM with CGAN, learns the spatiotemporal distribution of taxi hotspots to generate accurate predictions. 
Moreover, \cite{zhou2020variational} proposed a Bayesian framework that combines VAEs with GNNs for robust traffic prediction. The model benefits from the generative capabilities of VAEs and Normalizing Flow to capture multimodal traffic data distributions, addressing the uncertainty and complexity of road sensor networks.
%

%

{Many studies of DGMs focus on learning complex data distributions to generate accurate and realistic outcomes. However, an equally important but less explored aspect is the density estimation of outputs. Without probabilistic modeling, it is challenging to differentiate forecasts between low and high-noise scenarios, which is essential for offering a full probabilistic view and enhancing decision-making under uncertainty \citep{arnelid2019recurrent,rasul2020multivariate,rasul2021autoregressive,wen2023diffstg,lin2024specstg}.}
%
%
\cite{rasul2020multivariate} introduced an autoregressive deep learning framework for multivariate probabilistic time series forecasting that leverages conditioned normalizing flow. Their model learns the joint distribution of future observations by conditioning on historical data, enabling it to capture complex dependencies among multiple time series. This approach not only improves prediction accuracy but also provides robust uncertainty estimates. 
Building on this work, \cite{rasul2021autoregressive} proposed TimeGrad, which employs a denoising diffusion model for the same forecasting task. TimeGrad generates forecasts by progressively transforming white noise into meaningful data through a learned Markov chain. At each time step, it estimates the gradient of the data distribution and uses Langevin sampling to refine the predictions, further improving performance and uncertainty quantification.
Similarly, \cite{wen2023diffstg} developed DiffSTG, a framework for probabilistic Spatio-Temporal Graph (STG) forecasting that combines Spatio-Temporal GNNs (ST-GNNs) with DDPMs. DiffSTG captures both spatial and temporal dependencies in STG data while modeling inherent uncertainties. This model addresses the inefficiencies of TimeGrad in long-term forecasting by using a non-autoregressive approach to predict multiple future horizons simultaneously. 
In another study, \cite{feng2024latent} introduced the Latent Diffusion Transformer (LDT) for high-dimensional multivariate probabilistic time series forecasting. LDT employs a latent space approach combined with a diffusion-based conditional generator and a symmetric statistics-aware autoencoder to enhance the expressiveness and manageability of forecasting complex time series data.
Building on previous studies, DGMs have demonstrated robust capabilities in link-level analysis, providing essential tools for predicting and managing traffic conditions on road networks and thereby contributing to more efficient and responsive traffic management systems.

{One of the limitations is that, compared to agent-level studies, there is a notable gap in research on density estimation at the link level. Furthermore, while some probabilistic models excel at quantifying uncertainty, they often show less performance in point forecast accuracy relative to deterministic methods. Their reliance on variational inference can also lead to inaccurate posterior estimates when data is limited, underscoring a trade-off between capturing uncertainty and achieving precise forecasts.}

\subsubsection{Region-level Analysis}  \label{sec:3.2.3}
At the regional level, traffic management includes estimating and predicting traffic conditions across large geographic areas encompassing multiple road networks and transportation systems. This type of analysis is essential for understanding broader traffic patterns and making informed decisions that impact urban planning, resource allocation, and emergency response strategies. Key tasks at the region level include predicting and estimating average speeds across the network, estimating OD flows to understand travel patterns, forecasting traffic demand to anticipate future needs, analyzing overall traffic flow to identify congestion areas, and performing crash analysis to enhance network safety. The complexity of managing traffic at the regional level arises from the need to integrate data from diverse sources and understand the dynamic interactions between various components of the transportation network. From this perspective, DGMs are particularly well-suited due to their capability to learn from extensive datasets, capture complex spatial and temporal patterns, and model uncertainties of output results\citep{liang2018deep,saxena2019d,zhang2020off,li2020gacnet,kakkavas2021future,wu2021deep,feng2021short,kakkavas2021future,naji2021forecasting,wang2022data,mo2022trafficflowgan,li2022mgc,li2022spatial,yuan2023traffic, lin2023origin,zhang2023spatiotemporal,zarei2024application,li2024network,shao2024generative}. {In Table~\ref{tab:summary_region_level}, we summarize recent studies by considering the model used, task, observation/prediction length, sampling time, dataset, and evaluation metrics.} 

\begin{table}[hb]
\centering
\caption{{A summary of recent studies in region-level analysis using DGM}}
\label{tab:summary_region_level}
\resizebox{\textwidth}{!}{
\begin{tabular}{c|c|C{3cm}|c|c|C{3.5cm}|C{2cm}}
\toprule
\toprule
\textbf{Author} & 
\textbf{Model} & 
\textbf{Task} & 
\makecell{\textbf{Observation} / \\\textbf{Prediction} \\\textbf{Length}} & 
\makecell{\textbf{Sampling} \\\textbf{Time}} & 
\textbf{Dataset} & 
\makecell{\textbf{Evaluation} \\\textbf{Metrics}}\\
\midrule
\midrule

{ \cite{yu2019taxi}}       
& { cGAN}
& \makecell{{Taxi-Passenger} \\{Demand Prediction}}
& { 60min/10min} 
& { 10min}
& \makecell{{Beijing Taxi Dataset,} \\{New York City} \\{Taxi Dataset}}
& \makecell{{MSE,} \\{MAE, MAPE}}
\\\midrule

{ \cite{wang2020seqst}}       
& { SeqST-GAN}
& \makecell{{Crowd Flow} \\{Prediction}}
& { 24hr/1-20hr} 
& { 1hr}
& \makecell{ BikeNYC, TaxiNYC}
& \makecell{ RMSE, MAE}
\\\midrule  

{ \cite{huang2022gan}}       
& { DMGC-GAN}
& \makecell{{OD Ride-Hailing} \\{Demand Prediction}}
& { 200min/20min} 
& { 20min}
& \makecell{{New York City Taxi,} \\{Limousine Commission} \\{Ride-Hailing Dataset}}
& \makecell{{RMSE, MAE,} \\{MAPE, $R^2$}}
\\\midrule

{ \cite{li2022attentive}}        
& { ADST-GAN}
& \makecell{{Crowd Flow} \\{Prediction}}
& { 24h/1-4h} 
& { 1h}
& \makecell{ BikeNYC, TaxiNYC}
& \makecell{ RMSE}
\\\midrule

{ \cite{zhang2020curb}}       
& { Curb-GAN}
& \makecell{{Speed Prediction}\\{Inflow Prediction}}
& { 12hr/1-12hr} 
& { 1h}
& \makecell{{Shenzhen Speed Dataset} \\{Taxi Inflow Dataset}}
& \makecell{ RMSE, MAPE}
\\
\bottomrule
\bottomrule
\end{tabular}
  }
\end{table}

\cite{yu2019taxi} integrated a modified DBSCAN algorithm with a CGAN model built on LSTM to predict taxi-passenger demand. The model accounts for spatial, temporal, and external dependencies of input data, thereby providing accurate demand predictions.
In another study, \cite{wang2020seqst} introduced the SeqST-GAN, a sequence-to-sequence GAN model comprising LSTM and CNN for multi-step urban crowd flow prediction. This model enhances long-term prediction accuracy by treating crowd flow data as "image frames" and incorporating contextual features such as weather, holidays, and point of interests. 
\cite{huang2022gan} presented the Dynamic Multi-Graph Convolutional Network with GAN framework (DMGC-GAN) for OD-based ride-hailing demand prediction. The framework builds dynamic, directed OD graphs—capturing geographic adjacency, mutual attraction, and mobility associations—to tackle data sparsity and complex spatio-temporal dependencies. These graphs are processed by a Temporal Multi-Graph Convolutional Network (TMGCN) with a GRU and integrated into a GAN framework to refine predictions.
Moreover, \cite{li2022attentive} proposed the Attentive Dual-Head Spatial-Temporal GAN (ADST-GAN) to predict crowd flows. The model integrates attentive temporal and spatial mechanisms by combining ConvLSTM, self-attention, and a dual-head discriminator within the GAN framework. These components capture complex spatial-temporal dependencies in crowd flow data and mitigate overfitting, ensuring that the synthetic data generated closely mimics real-world traffic patterns. 
Furthermore, \cite{zhang2020curb} developed Curb-GAN, a conditional GAN-based model for estimating urban traffic conditions under various travel demand scenarios. Curb-GAN integrates dynamic convolutional layers to capture local spatial correlations along irregular road networks and self-attention mechanisms to model temporal dependencies, producing accurate and realistic traffic estimations. 
%
%
Previous research has shown the effectiveness of DGMs in managing and predicting traffic conditions at the regional level, providing valuable insights for strategic planning and operational decision-making across large transportation networks. {However, they also face several limitations. Many models depend heavily on high-quality, abundant training data, which may not always be available—especially when sudden or atypical demand scenarios occur. Additionally, many approaches rely on grid-based or clustering partitioning of urban areas, a method that can oversimplify and fail to capture the complex geometries of real-world road networks. Finally, despite the use of DGMs to improve output quality, there remains a risk of generating overly smooth or “blurry” predictions that obscure critical localized variations.}

\subsection{DGM for Unsupervised Representation Learning} \label{sec:3.3}

Unsupervised representation learning involves training a model to learn useful features or representations of the data without requiring labeled inputs. This method is valuable in environments where labeled data is sparse or expensive, which is often the case in transportation datasets. DGMs, such as VAEs,  are adept at discovering intricate structures in unlabeled data by learning to compress data and reconstruct it back into the original space. In other words, DGMs learn to encode data into a compact, latent space—a lower-dimensional representation of the original data—which preserves much of the information but in a more compressed form.  The latent vector is a compact representation of the input data that captures its most critical features. These vectors are the output of the model's encoder component and serve as a compressed knowledge base of the data, containing the essential information needed to reconstruct or generate new data points. In transportation research, the manipulation and analysis of these latent vectors allow researchers to infer traffic patterns, and anomalies that might not be apparent in the high-dimensional original data. 

One primary research direction is feature extraction and classification. It gains popularity in the field of transportation when employing DGMs, particularly because of their potent capability to discern and categorize complex, multi-dimensional data without direct supervision. Many studies have used the latent space in various classification tasks such as transportation mode identification, driving behavior analysis, and anomaly detection \citep{krajewski2018data, rakos2021learning,rakos2020compression,yao2022variational,rakos2021adversarial,de2022vehicles,xie2023novel,santhosh2021vehicular,islam2021crash,ding2019multi,yuan2023traffic,kim2021driving}. {For clarity, Table~\ref{tab:unsupervised_representation} summarizes these recent studies, detailing the models used, latent space functions, and datasets.} 
One notable study by \cite{zhang2022geosdva} introduced the Dirichlet VAE (DirVAE) to identify transportation modes from GPS trajectory data. The latent space of the DirVAE visualizes and classifies different transportation modes, demonstrating the utility of latent space in distinguishing between various modes of transportation.
In another study, \cite{boquet2020variational} used the VAE to represent complex, high-dimensional traffic data. The VAE learns a low-dimensional latent space representation that captures the underlying patterns and dependencies within the traffic data. This latent space can be used in various tasks such as missing-data imputation, anomaly detection, traffic forecasting,
Likewise, \cite{neumeier2021variational} proposed an unsupervised model based on the VAE architecture to predict vehicle trajectories with an interpretable latent space. The latent space can be analyzed to understand and predict lane-changing maneuvers, enhancing the model's utility in studying vehicle behaviors.
Moreover, \cite{chen2021traffic} developed a method for generating realistic traffic flow data using GANs with semantic latent code manipulation. By exploring and manipulating the semantic representations in the latent space, their model generates traffic flow data that accurately reflects realistic patterns and conditions. 
Additionally, \cite{santhosh2021vehicular} proposed a novel approach combining CNN and VAE to classify vehicle trajectories and detect anomalies. This hybrid model addresses the challenges of classifying time-series data of varying lengths and identifying anomalies such as lane violations, sudden speed changes, and vehicles moving in the wrong direction. The visualization of the latent space in their model provides a clear understanding of how the VAE encodes trajectory data and distinguishes between normal and anomalous patterns.

{
These studies demonstrate that DGMs are effective for feature extraction and classification in transportation. Leveraging the latent space allows researchers to reveal intricate patterns and dependencies, leading to more precise and insightful analysis of transportation data. Future work should aim to develop methods that efficiently extract and utilize the information in latent representations, thereby enhancing model robustness and interpretability and deepening our understanding of transportation phenomena.}

\begin{table}[h]
\caption{A summary of recent studies in unsupervised representation learning using DGM}
\label{tab:unsupervised_representation}
\centering
\resizebox{\textwidth}{!}{
\begin{tabular}{c|c|c|c}
\toprule
\toprule
\makecell{\textbf{Author}} & 
\makecell{\textbf{Model}} & 
\makecell{\textbf{Latent Space Function}} & 
\makecell{\textbf{Dataset}} \\
\midrule
\midrule
\makecell{\cite{zhang2022geosdva}}       
& \makecell{DirVAE}
& \makecell{Transportation Mode Classification}
& \makecell{{Geolife V1.3, OSMNX} \\{MTL Trajet 206,2017}}
\\\midrule
\makecell{\cite{boquet2020variational}}       
& \makecell{VAE}
& \makecell{Missing-data Imputation, \\ Anomaly Detection, Traffic Forecasting}
& \makecell{PEMS, UKM1, UKM4} 
\\\midrule
\makecell{\cite{neumeier2021variational}}       
& \makecell{DVAE}
& \makecell{Understanding Lane-Changing Maneuvers}
& \makecell{HighD}
\\\midrule
\makecell{\cite{chen2021traffic}}       
& \makecell{GAN}
& \makecell{{Linear Interpolation,} \\ {Manipulation Modifies Traffic-Flow Properties}}
& \makecell{PEMS}
\\\midrule
\makecell{\cite{santhosh2021vehicular}}       
& \makecell{CNN-VAE}
& \makecell{Distinguish Normal and \\ Anomalous Patterns}
& \makecell{T15, QMUL, 4WAY datasets}
\\
\bottomrule
\bottomrule
\end{tabular}
}
\end{table}

\subsection{Summary} \label{sec:3.4}
This section reviewed the significant role of DGMs in transportation research, focusing on their applications in data generation, estimation and prediction, and unsupervised representation learning. In data generation, DGMs are invaluable for producing synthetic datasets that replicate real-world traffic conditions. This capability is essential for training models, testing new scenarios, and supplementing real data, particularly when data collection is challenging or costly. DGMs can generate diverse traffic scenarios, realistic trajectories, and even complete missing data, thereby ensuring comprehensive and robust datasets for traffic analysis and management. For estimation and prediction, DGMs excel in modeling the uncertainties and variations inherent in traffic data. By learning from extensive datasets, they provide probabilistic forecasts of traffic conditions and enable effective decision-making for traffic management and planning. These models support the anticipation and adaptation to probable future traffic patterns, contributing to more efficient and responsive transportation systems. Lastly, in unsupervised representation learning, DGMs facilitate the analysis of high-dimensional traffic data by encoding it into compact, interpretable latent spaces. This enables the classification of transportation modes, analysis of driving behaviors, and detection of anomalies, providing deeper insights without the need for labeled data. The ability of DGMs to reveal hidden structures within data is particularly valuable for understanding and improving transportation systems. 

Despite these advances in transportation studies, future research must address several key challenges. Standardizing evaluation methods is crucial to ensure consistency and comparability across studies. Future studies should also consider the dynamic characteristics of traffic data, which vary significantly across different spatial and temporal contexts. Additionally, understanding and incorporating the causal relationships within traffic data, as well as addressing ethical and privacy concerns associated with synthetic data generation, are critical areas that require further exploration. These challenges, along with other emerging trends and opportunities in the use of DGMs for transportation research, will be discussed in more detail in Section~\ref{sec:5}.

\section{Tutorial}\label{sec:tutorial}

In this section, we present practical examples of how DGMs can be applied in transportation research. To reach a broader audience, we provide two hands-on tutorials: 1) Generating Household Travel Survey Data in Section~\ref{sec:tutorial_survey}, and 2) Generating Highway Traffic Speed Contour (Time-Space Diagram) in Section~\ref{sec:tutorial_speed}. {To keep the paper concise, detailed explanations of the model structure and its loss functions in the code are provided in Appendices \ref{appendix:tutorials_hts} and \ref{appendix:tutorials_sc}.} Importantly, all data and code used in these tutorials—including preprocessing scripts, model training, inference code, and pre-trained model parameters—are available in our GitHub repository: \url{https://github.com/UMN-Choi-Lab/DGMinTransportation}. The tutorial code is implemented in Python, with PyTorch serving as the primary library for the tutorials. Additional requirements are detailed in the associated GitHub repository. 

This tutorial aims to equip researchers with the necessary tools to explore the application of DGMs in transportation studies. {Accordingly, we structured the tutorial and developed the accompanying code using one representative discrete dataset and one representative continuous dataset. Given the scope of this paper, it is impractical to address every possible data type; however, by covering these key examples, we expect to offer a wide range of valuable insights. Moreover, our goal is not to identify the top-performing models within each research domain. To this end, although we performed numerical evaluations, the results can vary significantly depending on the numerous hyperparameters and optimizers used for each model. Consequently, we want to emphasize that readers should not judge any particular model solely based on the results presented in this paper.} Instead, the focus is on providing qualitative insights, which is a common practice in DGM research. For readers interested in rigorous performance evaluations, including numerical or analytical assessments, we suggest referring to the existing literature. Nevertheless, the code provided in our repository offers a robust foundation for initiating DGM-based research, enabling researchers to extend these models for specific transportation study requirements. 

The code implementations for the tutorial and experiments are conducted in a Python-based environment. Key libraries used include \texttt{torch} (tested both on v1.13 and v2.4) for deep learning model development, leveraging GPU acceleration to handle large-scale data and neural networks efficiently, and \texttt{numpy} and \texttt{pandas} for numerical computations and data manipulation. All development is done in a notebook environment to allow for interactive data exploration, experimentation, and real-time feedback. This setup also ensures compatibility with widely used tools, including Jupyter Notebook and Google Colab, offering flexibility for development and testing.

\subsection{Generating Household Travel Survey Data}\label{sec:tutorial_survey}

\begin{figure}
    \centering
\begin{subfigure}[t]{0.40\textwidth}
    \includegraphics[width=\linewidth]{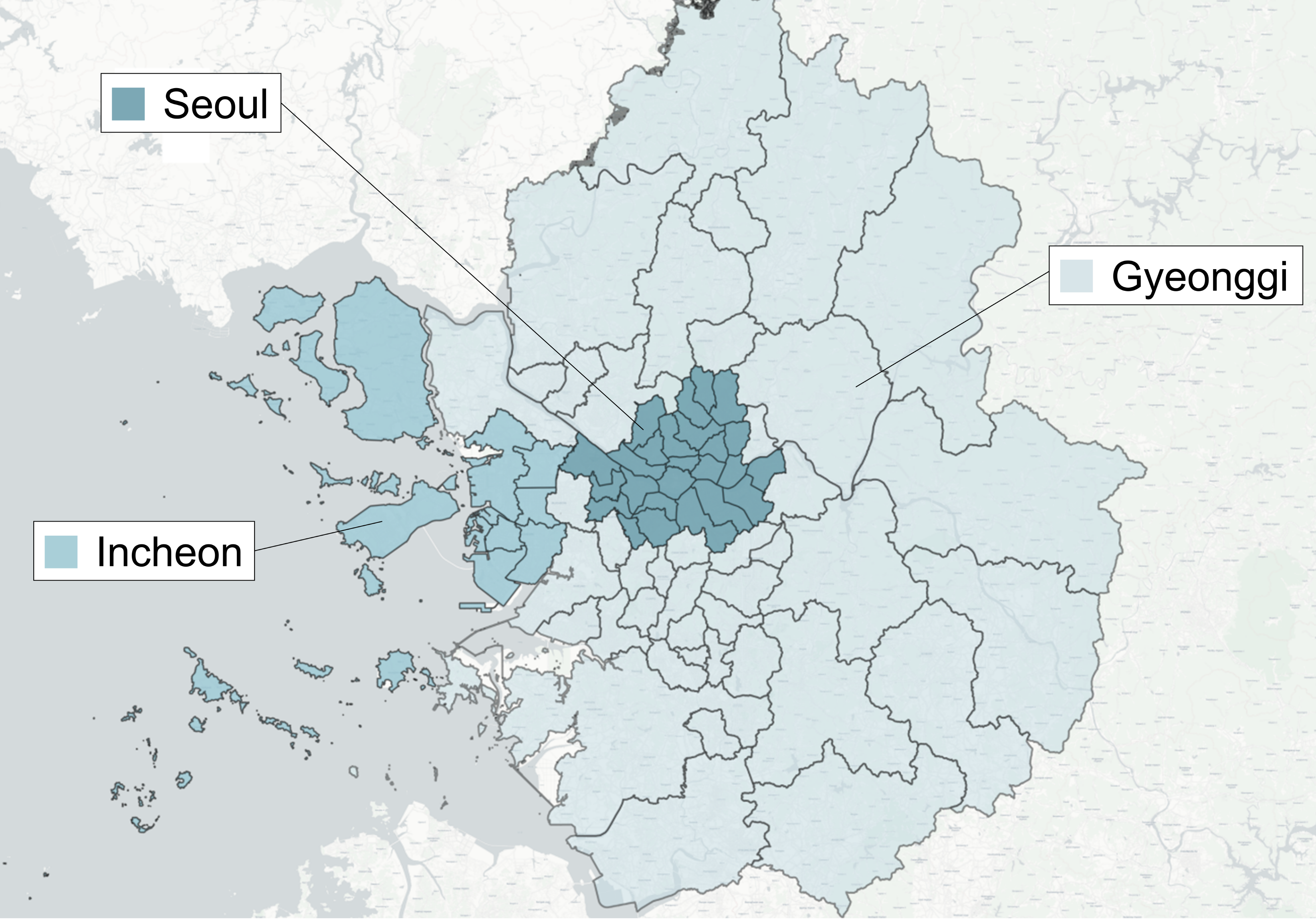}
    \caption{}    
    \label{fig:met_area}
\end{subfigure}
\begin{subfigure}[t]{0.40\textwidth}
    \includegraphics[width=\linewidth]{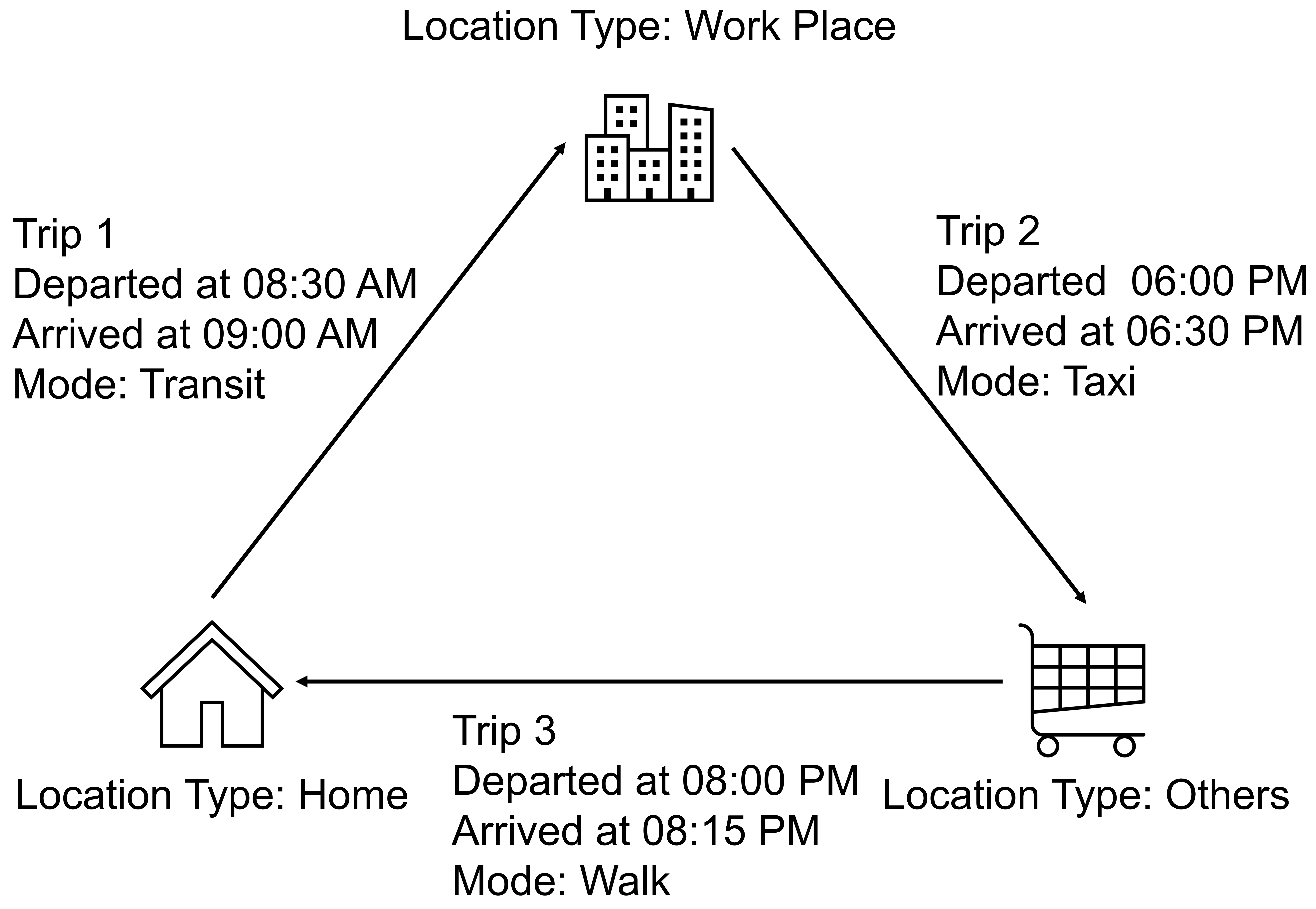}
    \caption{}
    \label{fig:trip_chain}
\end{subfigure}
\caption{Visual Description of Household Travel Survey dataset. (a) Study Area: Seoul Metropolitan Area (b) Example daily travel in the HTS data}
\end{figure}

Household Travel Survey (HTS) data provides a comprehensive overview of individuals' daily travel patterns, modes of transportation, and trip purposes within sampled households. These datasets often include sociodemographic and household-level characteristics, making them indispensable for transportation planners, policymakers, and researchers. HTS data is crucial for understanding mobility patterns, forecasting travel demands, and informing infrastructure developments, public transit planning, and sustainable transportation initiatives \citep{zhang2022generational}. However, traditional methods of collecting HTS data, such as face-to-face interviews or telephone surveys, typically yield sample sizes representing only 1–5\% of the population in a given region \citep{stopher2007household}, raising concerns about their representativeness and accuracy. Additionally, low sample rates, high survey costs, and safety concerns of interviewers further complicate the data collection process. While some countries, like Singapore, have adopted the online HTS collection system, questions about participant fidelity remain. These limitations underscore the need for innovative approaches, such as synthetic HTS data generation, to more accurately represent population-wide travel behaviors.

Advanced artificial intelligence techniques, particularly DGMs, have emerged as a promising solution to address the constraints of traditional HTS data collection. While prior studies have explored the use of GAN or VAE for synthetic data generation \citep{kim2023deep,garrido2020prediction,borysov2019generate}, other DGMs remain underexplored. Therefore, this section offers a tutorial on generating a simplified version of HTS data using DGMs, demonstrating their potential to overcome traditional data collection limitations and laying the groundwork for future research and applications in this field.

\subsubsection{Data and Preprocessing}

\begin{table}[t]
\caption{Trip chain data generated from example daily travel in Figure~\ref{fig:trip_chain}}
\label{tab:trip_chain}
\centering
\resizebox{\textwidth}{!}{
\begin{tabular}{l|l|l|l|l|l|l|l|l}
\toprule
\toprule
 & \makecell{Origin Type} & \makecell{Origin\\District Num.} & \makecell{Departure\\Time} & \makecell{Activity\\Type} & \makecell{Mode\\Type} & \makecell{Destination\\Type} & \makecell{Destination\\District Num.} & \makecell{Arrival\\Time} \\ \midrule\midrule
1 & Home & Gyeonggi-1 & 00:00 AM           & Start          & Stay      & Home                  & Gyeonggi-1              & 08:30 AM         \\ \midrule
2 & Home                      & Gyeonggi-1                  & 08:30 AM          & Travel        & Transit   & Work Place            & Seoul-5                 & 09:00 AM         \\ \midrule
3 & Work Place                & Seoul-5                     & 09:00 AM           & Work          & Stay      & Work Place            & Seoul-5                 & 06:00 PM      \\ \midrule
4 & Work Place                & Seoul-5                     & 06:00 PM        & Travel        & Taxi      & Others                & Gyeonggi-2              & 06:30 PM      \\ \midrule
5 & Others                    & Gyeonggi-2                  & 06:30 PM        & Shopping      & Stay      & Others                & Gyeonggi-2              & 08:15 PM      \\ \midrule
6 & Others                    & Gyeonggi-2                  & 08:15 PM        & Travel        & Walk      & Home                  & Gyeonggi-1              & 08:30 PM        \\ \midrule
7 & Home                      & Gyeonggi-1                  & 08:30 PM         & End          & Stay      & Home                  & Gyeonggi-1              & 12:00 PM     \\ 
\bottomrule
\bottomrule
\end{tabular}
}
\end{table}

In this tutorial, we applied DGMs to the 2010 HTS data from South Korea. This dataset provides comprehensive details on household travel patterns, including trip purposes, transportation modes, and travel times across various cities and districts in the region, making it a valuable resource for investigating urban mobility dynamics. While the original HTS dataset covers the entire country, this study only focuses specifically on the Seoul Metropolitan Area, as illustrated in Figure~\ref{fig:met_area}. The Seoul Metropolitan Area encompasses Seoul, Incheon, and Gyeonggi-do located in the northwest region of South Korea. With a population of approximately 26 million, this area represents more than half of the country's total population.

The original HTS includes both sociodemographic data and completed travel data for households. For simplicity, we exclude the sociodemographic data and focus solely on travel data. An example of the HTS travel data used in this study is shown in Figure~\ref{fig:trip_chain}. This travel data, which is composed of three different trips, can be transformed into the tabular data as the Table~\ref{tab:trip_chain}. To simplify the problem, we selected the types of origin, activity, mode choice, and destination as target features for data generation.

In our tutorial example, each DGM is designed to generate each row that contains the data in Table~\ref{tab:trip_chain} by learning the joint distribution of the selected features. The data includes origin and destination types divided into five categories: home, work, school, transfer points, and other locations. Transportation modes are categorized into nine discrete integers: 0 for the start of the day, 1 for the end of the day, 2 for staying at a location, and 3 through 8 representing various modes of transport such as walking, public transit, driving (driver and passenger), cycling, and taxis. Lastly,  activity types include eight categories: work, education, leisure, shopping, and escort activities. It is important to note that the goal of this tutorial is not generating the complete daily travel, but rather focusing on individual trips. This allows the model to capture the relationships between features. During model evaluation, we present qualitative assessments with marginal distributions of each feature. This qualitative approach prioritizes interpretability and understanding over direct numerical performance metrics. Nevertheless, the model is trained to capture the joint probability distribution, emphasizing its ability to generate realistic combinations of attributes.

Most DGMs are typically designed to operate in continuous spaces, learning to map the probability distribution of data. This design presents challenges when applied to discrete datasets, such as HTS data. When dealing with discrete variables, the probability distribution only has values at specific discrete points, while the probability is zero everywhere else. Additionally, to normalize the probabilities to sum to 1, the probability at these specific data points can become infinitely large. This causes DGMs to assign infinite likelihood to certain points, which makes the model unstable.
To address this issue, adding noise to the discrete variables, i.e., relaxed one hot encoding, was suggested from the literature. \citep{jang2016categorical} presented Categorical VAEs, which incorporate noise from Gumbel-Softmax distribution. \citep{maddison2016concrete} showed a similar approach using Concrete distribution to approximate the discrete value into continuous value.

In this tutorial, we use a simple dequantization technique to address the challenges of generating discrete data using DGMs. Rather than relying on complex methods such as Gumbel-Softmax to model discrete or categorical distribution, we add \textit{uniform random noise} to each discrete data sample before training the model, effectively transforming the discrete data into continuous space. After data generation using DGMs, we apply a \textit{floor} operation to convert it back to the discrete form. Mathematically, this process transforms the generation of a discrete value $\mathbf{x}_i$ into generating a continuous value within the range $[\mathbf{x}_i, \mathbf{x}_i+1)$. Consequently, computing the probability $p(\mathbf{x})$ is expressed as:
\begin{equation}
    \begin{split}
    p(\mathbf{x}) = \int p(\mathbf{x} + \mathbf{u}) d\mathbf{u} = \mathbb{E}_{\mathbf{u} \sim q(\mathbf{u}|\mathbf{x})} \left[ \frac{p(\mathbf{x} + \mathbf{u})}{q(\mathbf{u}|\mathbf{x})} \right],
    \end{split} 
\end{equation}
where $q(\mathbf{u}|\mathbf{x})$ is the noise distribution.
Since we assume it to be uniform, this can be rewritten as: 
\begin{equation}
    p(\mathbf{x}) =  \mathbb{E}_{\mathbf{u}\sim U\left[0,1\right)^D}\left[ p(\mathbf{x}+\mathbf{u}) \right],
\end{equation}
where $U\left[0,1\right)^D$ is a $D$-dimensional (data dimension) random uniform vector. This method effectively smooths the distribution and ensures stable training and generation of discrete variables.
This approach works well across various DGMs, as we demonstrate in this tutorial. The Python implementation of this dequantization process is included in the accompanying codebase. All five DGMs presented in this tutorial follow the same dequantization process, ensuring consistency and comparability across models.

    
        

\subsubsection{Results Comparison}
\begin{table}[!htbp]
\caption{Performance comparison across different DGMs in HTS Data}
\label{tab:hts_results}
\centering
{\fontsize{9pt}{11pt}\selectfont
\fontfamily{phv}\selectfont
\begin{tabular}{l c c c c c}
\toprule
\toprule
 & VAE & GAN & NF & NCSN & DDPM \\
\midrule
SRMSE & 0.53550 & 0.48710 & \underline{\textbf{0.32318}} & 1.39829 & 1.09457 \\
MAE   & 0.03031 & 0.02350 & \underline{\textbf{0.01432}} & 0.04501 & 0.03346 \\
KLD   & 0.04774 & 0.05254 & \underline{\textbf{0.01605}} & 0.07657 & 0.05494 \\
\bottomrule
\bottomrule
\end{tabular}
}
\end{table}

In this section, we share the empirical insights gained during model training. {Further details about each model can be found in Appendix \ref{appendix:tutorials_hts}.} Figure~\ref{fig:results_disc} presents the results of generating HTS data.

Similar to how VAEs tend to generate slightly blurry images in image generation, in this experiment with discrete values, the probability distribution also appeared relatively uniform rather than being overly skewed towards one side. However, it often struggled to accurately match the activity type and mode type, depending on the specified conditions. This may be attributed to the increased complexity resulting from the higher number of attributes. 
In contrast, the GAN frequently encountered mode collapse, leading to cases where all probabilities were assigned to a particular activity or mode. However, this issue was less pronounced in cases involving fewer data attributes, such as problems related to origin location and destination location type.
The Normalizing Flow model demonstrated superior performance, with robustness across a wide range of hyperparameters. The loss reduction pattern followed the typical behavior expected in learning models, with an initial sharp decrease followed by a gradual decline in later stages. Throughout the training process and in the results, the model consistently performed well.
The Diffusion model, which is commonly built using a U-Net architecture, was instead implemented with a simpler three-layer neural network in our approach. Despite initial concerns regarding its effectiveness, the model performed relatively well due to the simplicity of the dataset. However, occasional loss spikes occurred, which led the model to failure of training. Overall, the learning process remained stable, but future experiments could benefit from more sophisticated neural network architectures to further enhance performance.
The NCSN model exhibited high sensitivity to hyperparameters. For instance, when the max and min sigma values were low, the model displayed characteristics of high entropy or temperature, leading to unpredictable performance. Conversely, higher sigma values resulted in one-hot-like behavior. Additionally, the model's performance varied across different categories: when it performed well on generating the location type, performance on generating types of action and mode became poor, and vice versa. Balancing hyperparameters effectively remains a key challenge for ensuring consistent performance across all attributes.

{

Table~\ref{tab:hts_results} presents the numerical results obtained by comparing the distribution of the ground truth data with that of the generated data using the SRMSE, MAE, and KLD\footnote{See Appendix \ref{sec:prelim}} metrics. For SRMSE, we computed the distribution for each unique combination of two columns and then calculated the SRMSE from these distributions of ground truth and generated data; this procedure was repeated for all combinations of two columns, and then we averaged the sum of it. As we mentioned, results presented herein are provided solely as examples, and readers are encouraged to modify various hyperparameters and optimizers using the code available on GitHub to obtain alternative outcomes.
In the Table~\ref{tab:hts_results}, the VAE and GAN models yielded SRMSE, MAE, and KLD values of 0.53550/0.03031/0.04774 and 0.48710/0.02350/0.05254, respectively, indicating relatively higher error levels. Notably, the Normalizing Flow model demonstrated superior performance with values of 0.32318, 0.01432, and 0.01605 for SRMSE, MAE, and KLD, respectively, thereby exhibiting higher accuracy and robustness across varied hyperparameter configurations. In contrast, the NCSN and DDPM models showed comparatively larger error metrics, with the NCSN model displaying significant sensitivity to hyperparameter adjustments. These results emphasize that model performance is intricately linked to the selected architecture and training strategies.

}

\begin{figure}[t]
    \centering\includegraphics[width=0.97\textwidth]{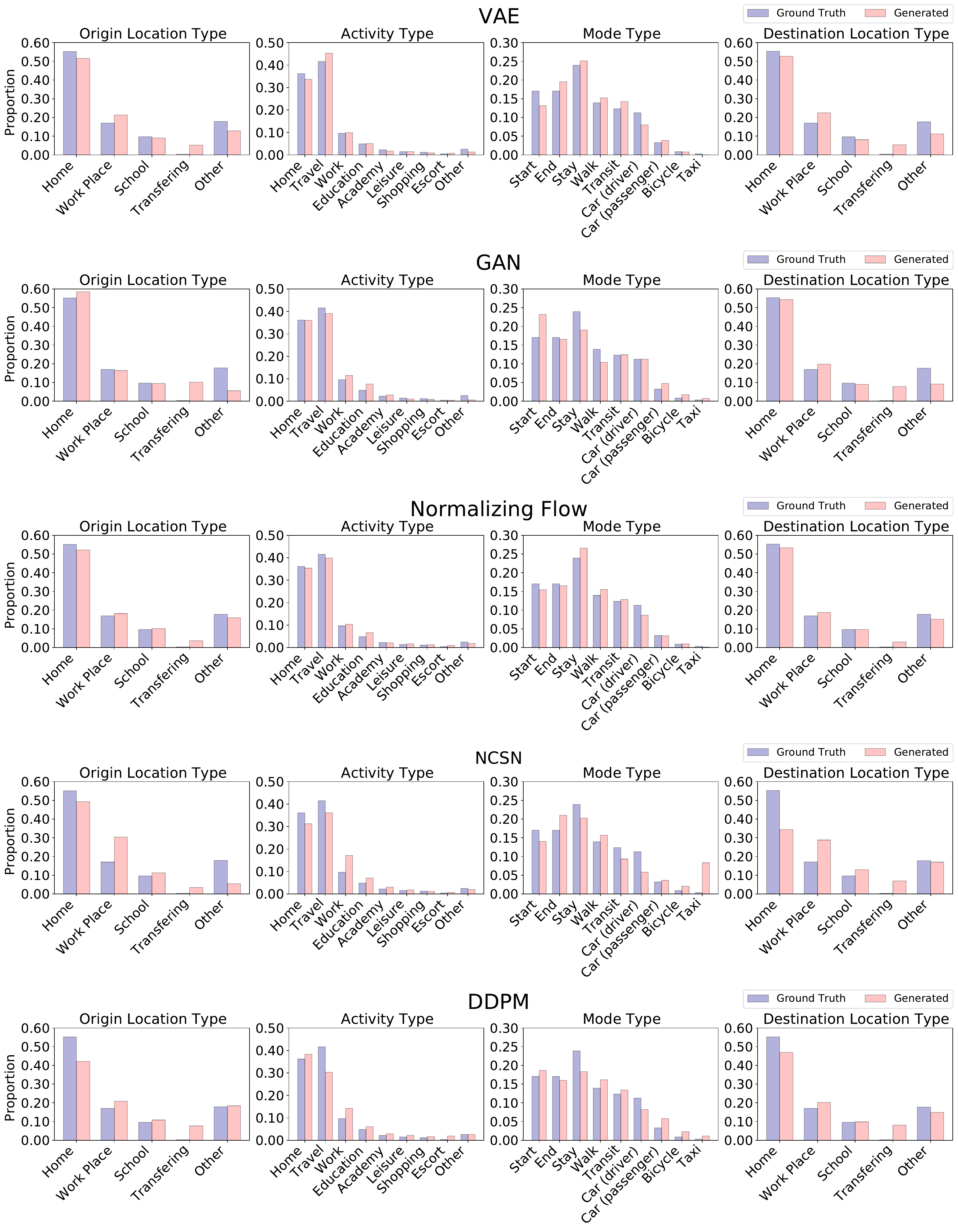}
    \caption{Results of each DGM in HTS data generation}
    \label{fig:results_disc}
\end{figure}

\clearpage
\subsection{Generating Highway Traffic Speed Contour}\label{sec:tutorial_speed}

The generation of highway traffic speed contours plays a crucial role in analyzing and visualizing traffic dynamics across both time and space. Traditionally, traffic flow has been represented using a time-space diagram of vehicles, which depicts the movement of vehicles along a road segment over time, providing insight into congestion patterns and traffic bottlenecks. However, converting this into a time-space speed contour offers several advantages. While traditional time-space diagrams focus on the position of vehicles at specific times, speed contours emphasize variations in traffic speed across both temporal and spatial dimensions, offering a clearer representation of traffic conditions such as slowdowns, jams, or free-flowing traffic across broader time periods. In this tutorial, we generated the traffic speed contour data based on the vehicle trajectory data.

Speed contours offer a continuous and intuitive visualization of traffic speeds across different locations and time intervals, making them particularly useful for traffic management and planning. They enable transportation professionals to quickly identify and quantify areas of concern, such as sections of highways prone to congestion or inconsistent traffic flow, thereby facilitating better decision-making. Moreover, these contours can serve as input for traffic forecasting models, aiding in the development of responsive traffic control systems, infrastructure improvements, and policy interventions aimed at reducing congestion and improving road safety. In this section, we aim to generate highway traffic speed contours from the input noise. Through the learning process, the model captures the spatial and temporal joint distribution of traffic dynamics, resulting in a detailed speed contour that reflects variations in traffic conditions.

\subsubsection{Data and Preprocessing}

\begin{figure}[t]
    \centering
    \includegraphics[width=0.97\linewidth]{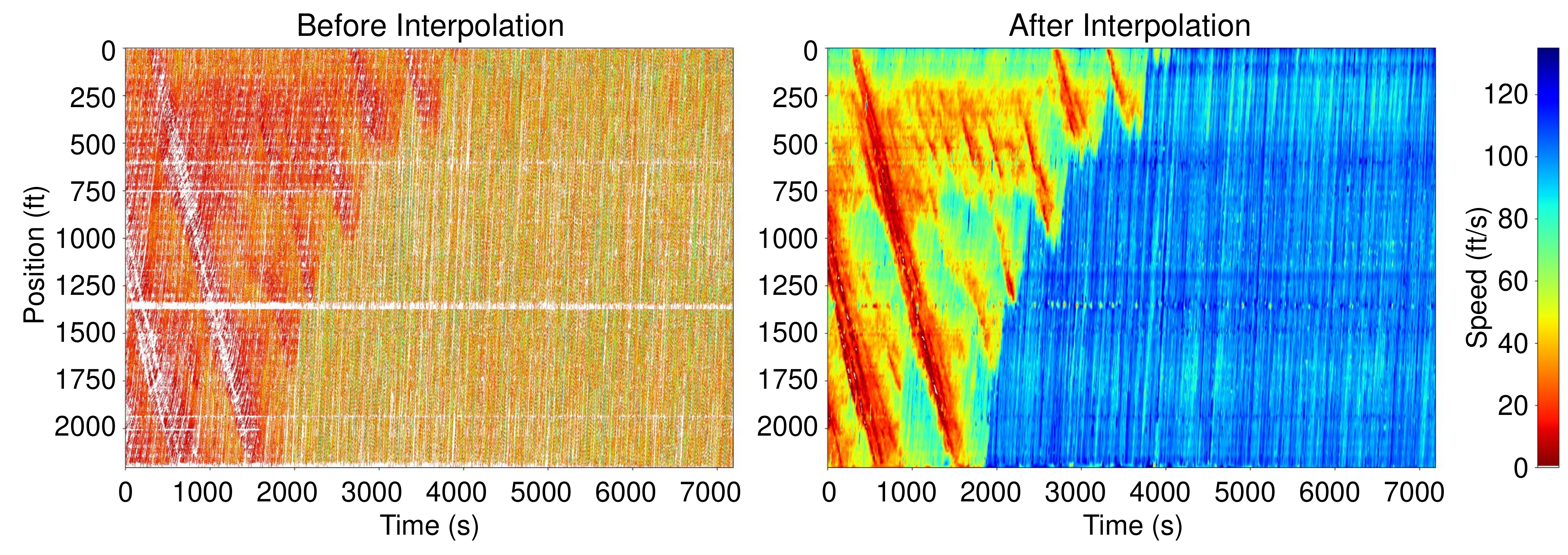}
    \caption{Comparison of traffic speed contours before and after the interpolation}
    \label{fig:smoothing_results}
\end{figure}

In our tutorial section, we apply the DGMs to the Tennessee Department of Transportation's I-24 Mobility Technology Interstate Observation Network (MOTION) data \citep{gloudemans202324}\footnote{\url{https://i24motion.org/data}}. I-24 MOTION data contains the vehicle trajectories in a four-mile section of I-24 in the Nashville-Davidson County Metropolitan area. It captures the trajectory in a high spatial and temporal resolution by using 294 ultra-high definition cameras and AI algorithms from Vanderbilt University. The AI algorithm transforms the raw videos into detailed 2-dimensional trajectories which is suitable to study vehicle behavior and traffic flow. The data collection occurred from Monday, November 21, 2022, to Friday, December 2, 2022, encompassing all eastbound and westbound lanes. The collection was done during peak morning hours, from 6 AM to 10 AM. This dataset is significant as it includes a variety of days as well as trajectories during an accident that occurred on November 21. However, it is important to note that when a single vehicle appears multiple times across different cameras, each instance is treated as a separate vehicle. Therefore, the whole trajectory will be sliced into multiple short sequences of trajectories. Consequently, this data needs a vehicle-matching algorithm to obtain the full trajectories of the single vehicle, but it can be effectively utilized for constructing time-space traffic contours without matching the vehicles.

The raw data from the MOTION project comprises segments of vehicle trajectory information. In this tutorial, our focus is on the reconstruction of traffic speed contours. To facilitate this, the trajectory data is transformed into velocity data corresponding to each specific time and position. Given that the calculation of velocity data based on trajectories inevitably results in the presence of zero or NaN values (except in the unlikely event of a vehicle being present at every possible position throughout the entire timespan), preprocessing was necessary. This preprocessing predominantly involved interpolation. The left and right of Figure~\ref{fig:smoothing_results} show the time-space traffic contours diagram before and after the interpolation.

The interpolation process is composed of two main steps. Firstly, we interpolate the zero and NaN values using linear and nearest-neighbor interpolation methods, respectively. Secondly, we apply a smoothing technique based on Edie's definition of traffic flow dynamics \citep{treiber2003adaptive}. The initial interpolation step is crucial; without it, if we attempt to interpolate every zero or NaN value directly using Edie's box method, the required size of Edie's box would become excessively large, leading to significant distortion of the dataset. Before interpolation, most of the regions are colored in red because the 0 and NaN values are depicted as red. The actual values of these regions are much higher than what is shown in the figure. After interpolation, we can observe a much clearer propagation of congestion and its subsequent dissipation. Free flow is also clearly illustrated after interpolation. The blue region forming a straight line after 3,600 seconds indicates an increase in traffic speed following the morning rush hour as the time passes 9 AM.

{The detailed pseudocode for the interpolation process, based on Edie's definition of traffic flow dynamics, is provided in Appendix~\ref{appendix:smoothing}}. The smoothing process, known as the ``adaptive smoothing method,'' is designed to filter out noisy fluctuations while considering the primary direction of flow propagation. This method employs spatiotemporal correlation analysis to identify the dominant characteristic line. The adaptive smoothing method uses a spatio-temporal low-pass filter that allows only low-frequency Fourier components to pass through, smoothing out high-frequency components. The filter eliminates high-frequency noises over a timescale shorter than $\tau$ and spatial noise over a length scale shorter than $\sigma$. The values for $\tau$ and $\sigma$ were determined through empirical trials to ensure the data is not overly blurred while effectively passing the main propagation. The filter captures two main propagation types: free flow and congested traffic flow. These two propagations have different coefficients of velocity and direction, reflecting the distinct properties of each traffic flow type. The property of each traffic flow is reflected based on the different values of each coefficient. 

\subsubsection{Results Comparison}
We discuss the key findings from our model training in this section. {As with the HTS data generation, a comprehensive description of each model is available in Appendix~\ref{appendix:tutorials_sc}.} Figure~\ref{fig:results} presents the 64 images of ground truth speed contour data and outputs of 64 samples generated by each DGM. It is worth noting that the results generated by the DGMs in Figure~\ref{fig:results} do not directly match the ground truth in the same figure. These images are instead "plausible" representations generated from noise. Therefore, readers should keep in mind that good images are those that closely resemble the ground truth, with clear and well-defined traffic speed patterns across the contours. In contrast, poor images may appear overly blurred, and noisy, or suffer from mode collapse, where the generated outputs fail to capture the underlying traffic dynamics.

\begin{figure}
    \centering
    \includegraphics[width=0.9\textwidth]{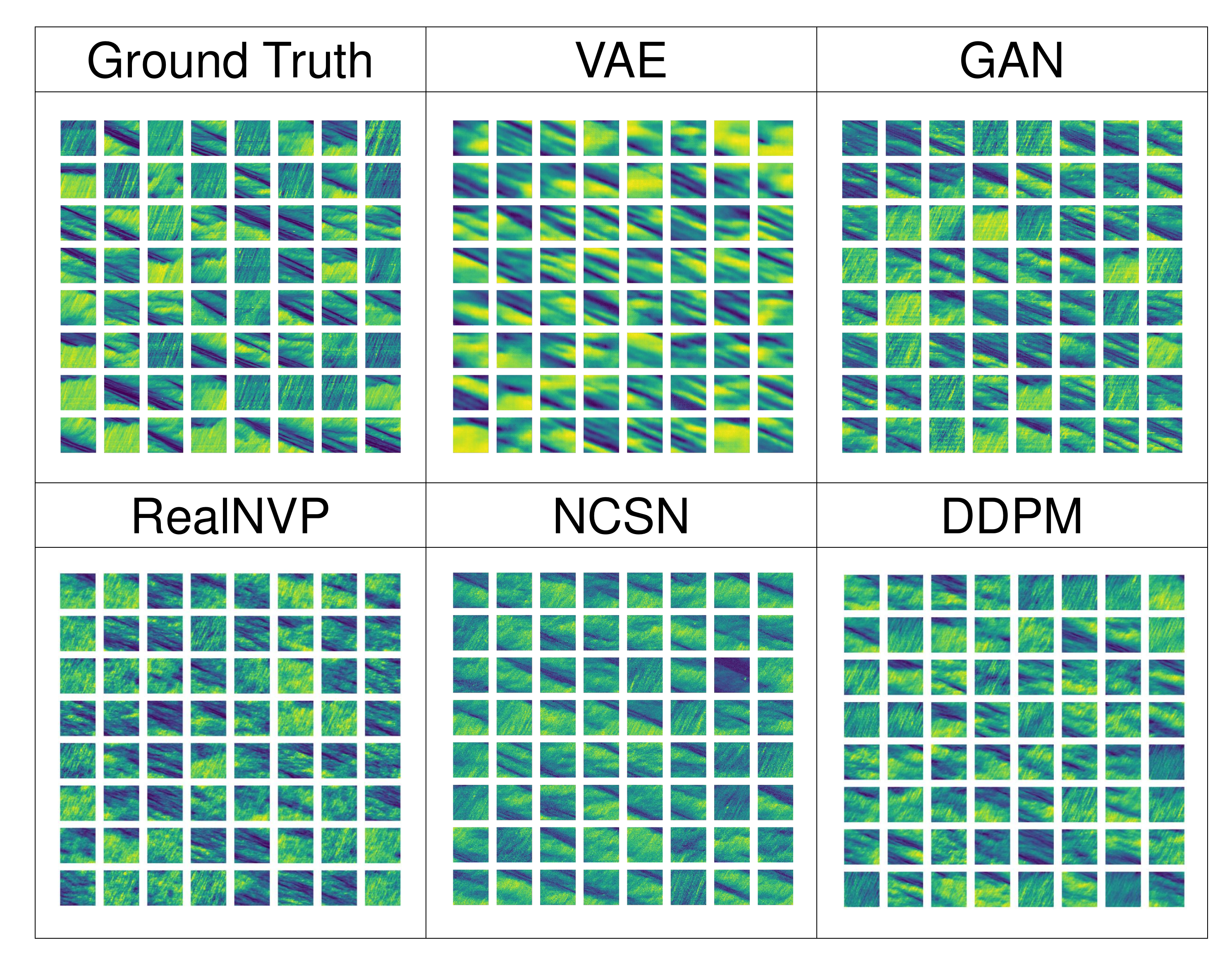}
    \caption{Ground truth data and generated samples from each DGM in highway traffic speed contour generation (Ground truth data and generated sample do not have a matching relationship. Readers should primarily focus on examining whether the visual patterns of the two datasets are similar.)}
    \label{fig:results}
\end{figure}

It is evident that the VAE outputs are noticeably more blurred compared to those produced by the other models. Throughout the training process, we experimented with a wide range of neural network architectures for the VAE. Despite attempting to train the VAE with neural networks that had four to eight times more parameters than the current configuration, we were unable to eliminate the inherent blurriness of the VAE outputs. Unlike other algorithms, which exhibited some variability in performance across training sessions, the VAE consistently produced stable results both numerically and visually. However, the VAE outputs remained uniformly blurred, with some generated images appearing severely noisy and closely resembling the mean of the dataset.

In contrast, GANs demonstrated superior performance, especially given the simplicity of their neural network architecture. Even when using a neural network architecture almost identical to that of the VAE, the GAN produced clearer and more diverse images. Although the training process occasionally showed abrupt fluctuations in the loss function, as long as the generator's training did not lag significantly behind that of the discriminator, the model eventually achieved a balance. Considering the simplicity of the code, training time, and generation speed, the strong performance of such a straightforward algorithm is very encouraging.

RealNVP displayed the most variability among all the algorithms. Occasionally, it produced excellent results, but this occurred in only about 1 in 20 training runs. Since the loss function consists of four terms, the model's training trajectory varied depending on the coefficients applied to each term. The coefficients we proposed in the code were the result of extensive trial and error. It was observed that both log-likelihood and regularization were crucial, and the performance was highly sensitive to the numerical value of the first term in each component. Additionally, once RealNVP fell into a local minimum during training, it was unable to escape, resulting in poor performance across all samples and mode collapse, where the images appeared more like homogeneous noise. The generated images resembled ground truth data with slightly mixed boundaries between data values.

Compared to RealNVP, DDPM demonstrated relatively stable performance. It consistently produced uniform models with good image quality across different training sessions. The resulting images resembled ground truth data with the addition of white noise, similar to photographs taken at a very high ISO setting. NCSN, which follows a mechanism fundamentally similar to that of DDPM, exhibited comparable performance. Despite the uniform resolution of 64x64, the images conveyed a sense of being highly fine-grained.

\begin{table}[!htbp]
\caption{Performance comparison across different DGMs in speed contour data}
\label{tab:lpips_sobel_results}
\centering
{\fontsize{9pt}{11pt}\selectfont
\fontfamily{phv}\selectfont
\begin{tabular}{l c c c c c}
\toprule
\toprule
      & VAE      & GAN                           & RealNVP     & NCSN     & DDPM     \\
\midrule
MSE   & 0.06646  & 0.08569 & \underline{\textbf{0.06421}} & 0.10583 & 0.23108 \\
LPIPS & 0.38957  & \underline{\textbf{0.14988}} & 0.20113  & 0.36131 & 0.23849 \\
Sobel-Edge-MSE & 0.00169 & \underline{\textbf{0.00147}} & 0.00237  & 0.00198 & 0.00201 \\
\bottomrule
\bottomrule
\end{tabular}
}
\end{table}

{
To complement our qualitative analysis, we introduced two additional metrics beyond the mean squared error (MSE) of the image distribution. The first is that we incorporated the Learned Perceptual Image Patch Similarity (LPIPS) metric to assess perceptual similarity between real and generated images\citep{zhang2018unreasonable}. LPIPS is a metric that leverages deep features extracted from pretrained convolutional neural networks, such as AlexNet, which we used in the current study, to capture both low-level details and high-level semantic information. Rather than relying solely on pixel-wise differences, LPIPS computes feature maps across multiple layers, which are first L2-normalized. The spatial and channel-wise differences between corresponding feature maps are then aggregated using spatial averaging and weighted by learned parameters that reflect each channel's importance. The final perceptual distance, obtained by summing these weighted differences across all layers, aligns closely with human perceptual judgments. In our evaluation framework, a pretrained LPIPS model—trained with human evaluation data (e.g., BAPPS) was utilized to ensure that the computed distances reliably mirror subjective image quality assessments.

In addition to LPIPS, we employed the Sobel-Edge-MSE metric to quantitatively evaluate the structural fidelity of the generated images. This metric harnesses the classical Sobel operator to extract edge information by computing the gradient magnitude in both horizontal and vertical directions, effectively capturing the structural content of an image. By applying the Sobel operator to both the real and generated images, we obtained corresponding edge maps, and the MSE between these maps was calculated. A lower Sobel-Edge-MSE value indicates that the generated image more accurately replicates the edge features of the real image, thereby preserving crucial structural details. Shock waves in time-space diagrams are typically manifested as distinct linear or planar features, characterized by clear and well-defined boundaries that delineate abrupt transitions in traffic flow. Consequently, when the generated images exhibit low Sobel-Edge-MSE, it indicates that the model effectively captures the essential dynamics of shockwave propagation by replicating these sharp transitions and geometrical features.

We generated 600 images and randomly sampled 600 real images. Subsequently, LPIPS and Sobel-Edge scores were computed for every one of the 600×600 pairs. We then solved a matching problem to minimize the mean of the total sum of these scores. This approach was adopted because selecting only the lowest LPIPS and Sobel-Edge scores among the real samples could yield falsely high-performance metrics when a model exclusively targets a single traffic state (image) among multiple states. In other words, a model might achieve high precision by accurately generating a specific traffic state but at the expense of generation diversity (i.e., low recall), which is undesirable in image generation tasks where diversity is critical. Consequently, we established the current metric framework, which effectively serves as a surrogate for the F1 score by balancing both precision and recall. This approach might be challenged on the basis of using only 600 images, and such concerns are valid. The more images generated and compared, the better the model's actual performance can be evaluated. However, since this comparison method scales quadratically $(O(n^2))$, 600 images were empirically the maximum number of images for this evaluation framework.

Table~\ref{tab:lpips_sobel_results} summarizes the quantitative evaluation of the generative models using the MSE, LPIPS, and Sobel-Edge-MSE metrics. Notably, the GAN model achieved the lowest LPIPS score (0.14988) and the lowest Sobel-Edge-MSE (0.00147), indicating superior perceptual similarity and structural fidelity compared to the other models. In contrast, the VAE exhibited the highest LPIPS (0.38957) and a relatively higher Sobel-Edge-MSE (0.00169), reflecting its tendency to produce blurred and less detailed outputs. The RealNVP, NCSN, and DDPM models demonstrated intermediate performance with respect to LPIPS and Sobel-Edge-MSE, suggesting a balanced capability in capturing both perceptual and structural characteristics of the ground truth data. It is important to note that MSE is not well-suited for reflecting visual similarity; in our results, the most prominent difference between the VAE and the other models is not captured by the MSE values (VAE: 0.06646, RealNVP: 0.06421, etc.). Overall, the quantitative results appear to be largely consistent with the visual outcomes: lower LPIPS scores indicate that the overall appearance of the generated images more closely resembles the real images, while lower Sobel-Edge-MSE scores suggest that the edge structures—particularly the shockwaves—exhibit greater similarity to those in the ground truth. These findings not only demonstrate the utility of these metrics in capturing essential qualitative aspects but also highlight the inherent challenges in their evaluation, given that each metric quantifies a subjective, qualitative feature.

\begin{figure}
    \centering
    \includegraphics[width=0.9\textwidth]{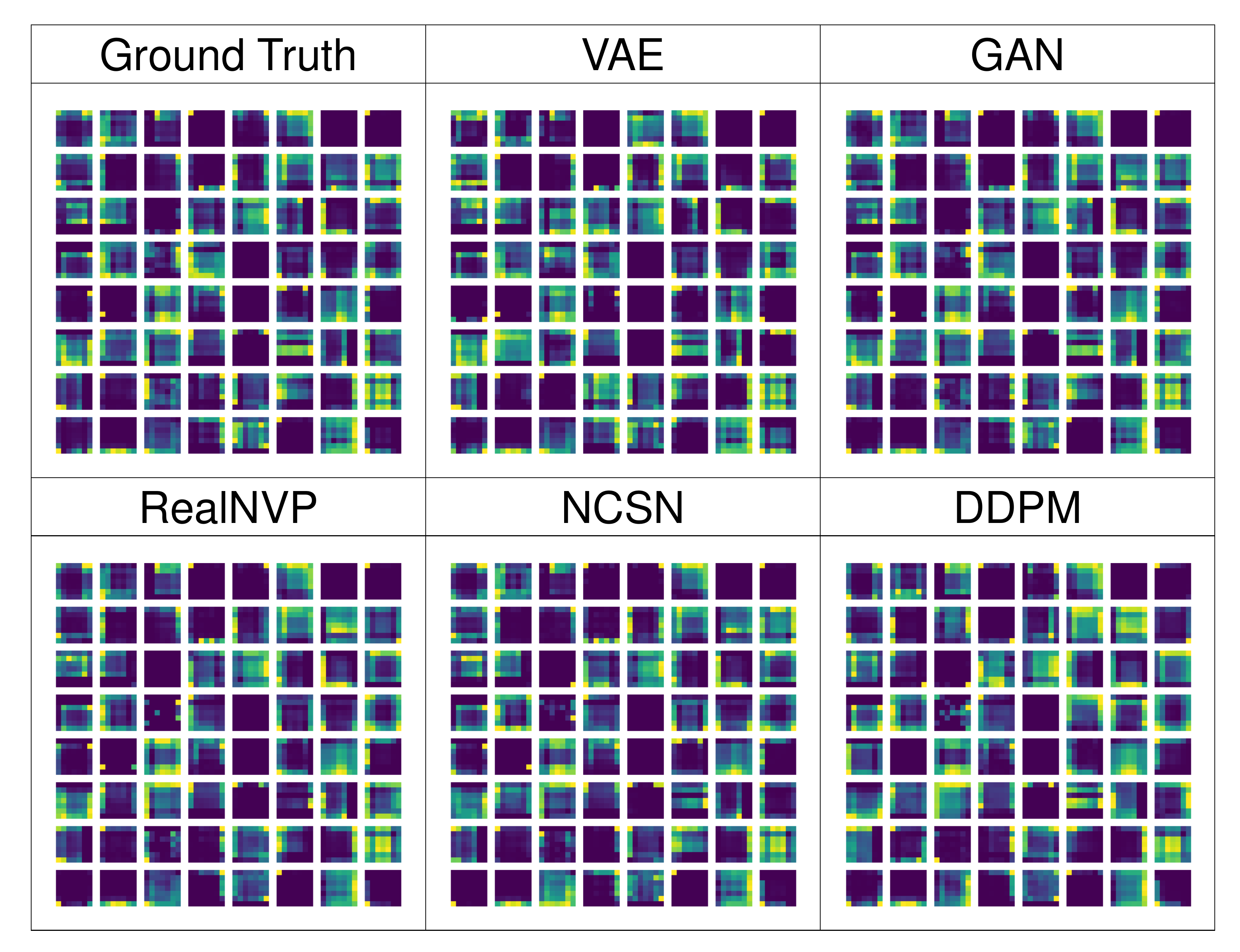}
    \caption{Feature map from ground truth data and generated samples from each DGM in highway traffic speed contour (This is a visualization of the 64 channels, from one random image from each set)}
    \label{fig:result_feautre_map}
\end{figure}

Figure~\ref{fig:result_feautre_map} illustrates averaged feature maps extracted from AlexNet for two distinct datasets: 600 ground truth images and 600 images generated by various models. Specifically, the feature map from the first convolutional layer was selected, and among its multiple channels, the first 64 channels were visualized. For each dataset, the feature maps across all 600 images were summed and then averaged to produce a representative visualization.

Overall, the general pattern of activations appears similar between the ground truth and the generated images, but qualitative differences are evident in specific regions. For instance, notable discrepancies between the ground truth and VAE outputs are observed in the feature map corresponding to the first row, fifth column, and in the seventh row, third and fourth column. In contrast, GAN-generated shows consistent results with their superior LPIPS performance. Their feature maps closely resemble those of the ground truth, albeit with a detectable difference in the second row, seventh column. Similarly, RealNVP demonstrates differences in the image at the sixth row and sixth column. NCSN shows significant differences in the sixth row, last column and at the last row, third column, while DDPM exhibits significant deviations in the third column of the second row, the last column of the second row, and the last column of the third row. It should be emphasized that these observations are qualitative in nature. Nonetheless, the qualitative analysis aligns with the LPIPS evaluation, particularly highlighting the high degree of feature similarity in GAN outputs. This indicates that while all generative models generally capture the underlying structure of the ground truth images, subtle, localized discrepancies in feature representation may affect perceptual similarity metrics.

}

In overall, it is crucial to approach model performance evaluations with caution. Qualitative {and quantitative} comparisons do not provide a comprehensive assessment of model efficacy. The performance of these models can vary significantly based on hyperparameter tuning and architectural choices, making it challenging to definitively claim the superiority of any single neural network configuration or hyperparameter set. Interestingly, our empirical observations during implementation revealed instances where simpler network structures yielded superior results, underscoring the complexity of model optimization. This phenomenon highlights the importance of thorough experimentation and the potential pitfalls of overgeneralization in neural network performance assessment. Future research should focus on developing more robust quantitative metrics for model comparison, considering the multifaceted nature of generative model performance across various tasks and datasets.

\clearpage

\begin{figure}[h]
    \centering
    \includegraphics[width=0.7\linewidth]{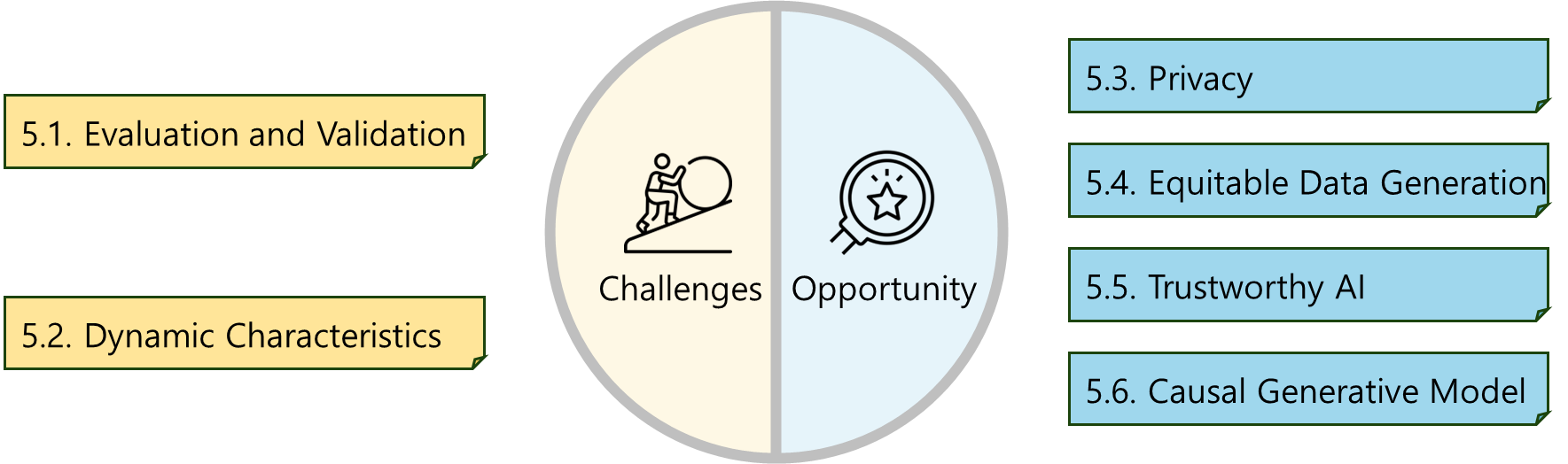}
    \caption{Challenges and opportunities of applying DGMs in transportation research}
    \label{fig:chal_opp}
\end{figure}

\section{Challenges and Opportunities}\label{sec:5}
{

Figure \ref{fig:chal_opp} provides a structured overview of the challenges and opportunities in applying Deep Generative Models (DGMs) to transportation research as discussed in this paper. The left side highlights key challenges, including evaluation and validation (Section \ref{sec:5.1}) and dynamic characteristics of transportation data (Section \ref{sec:5.2}). These challenges stem from the difficulties in assessing generative models in complex transportation systems and the evolving nature of mobility data, which can introduce distribution shifts over time.

On the right side, the figure presents opportunities for advancing DGMs in transportation research. These include leveraging DGMs for privacy preservation (Section \ref{sec:5.3}), ensuring equitable data generation to mitigate bias (Section \ref{sec:5.4}), fostering trustworthy AI through transparency and fairness (Section \ref{sec:5.5}), and integrating causal generative models to enhance interpretability and generalizability (Section \ref{sec:5.6}). By addressing the challenges and capitalizing on these opportunities, DGMs can become powerful tools for transportation research.
}

\subsection{Evaluation and Validation of Generative Models} \label{sec:5.1}

{
 
Deep generative models (DGMs) aim to capture the underlying probabilistic distributions of complex, high-dimensional data. Evaluating these models is inherently multifaceted, requiring a blend of rigorous quantitative metrics and nuanced qualitative assessments. For transportation applications, this strategy must address several key dimensions: ensuring fidelity through robust metrics to confirm that synthetic data closely replicate real-world statistical properties; maintaining realism by incorporating physical laws that govern traffic dynamics; and preserving diversity to capture the full spectrum of observed behaviors, thereby avoiding pitfalls like mode collapse. Additionally, it is essential to thoughtfully model the complex, structured nature of transportation data. Integrating these elements is crucial for generating synthetic data that are not only statistically robust but also practically valuable for simulation, decision support, and further analysis.

Evaluating DGMs involves assessing how well these models generate data that accurately reflect the underlying distribution of observed data using quantitative metrics. Achieving high fidelity in transportation data analysis is both critical and challenging, largely because available datasets are often limited, increasing the risk that deep learning models may overfit if not properly validated. One principled approach is to use scoring rules \citep{gneiting2007strictly} that quantify how well the predicted probabilities align with actual outcomes. For example, the Log Score measures the negative log-likelihood of the synthetic data under the probabilistic distribution produced by the model. Also, metrics like Kullback–Leibler Divergence (KLD), Jensen–Shannon Divergence (JSD), and standardized RMSE (SRMSE) could be used to compare distributions. In addition, the Continuously Ranked Probabilistic Score (CRPS) and the Energy Score (ES) provide measures of how well the cumulative distribution functions capture the nuances of continuous univariate and multivariate data, respectively. Another key challenge lies in ensuring realism. Traffic data are governed by physical laws, and DGMs must generate outputs that adhere to these constraints. When creating synthetic vehicle trajectories, it is essential that the models incorporate realistic driving dynamics---such as smooth acceleration, deceleration, and turning behaviors---to replicate actual traffic flow patterns. If models rely solely on spatial data without integrating temporal dynamics (like acceleration or turning angles), the resulting trajectories may appear statistically plausible but fail to capture the underlying physical realities, reducing their practical utility. This issue is similarly evident in the generation of traffic speed data as we introduced in Sec~\ref{sec:tutorial_speed}, which are fundamentally produced by the dynamics of microscopic traffic flow. Treating such data merely as images or numerical arrays can strip away their inherent physical context, leading to unrealistic outputs that violate known traffic flow patterns.

In addition to fidelity and realism, it is crucial for DGMs to capture the full range of variability inherent in transportation data. Real-world traffic exhibits a broad spectrum of behaviors---ranging from individual driving styles to complex vehicle interactions. A common pitfall is mode collapse, where the generated outputs lack diversity and fail to represent rare but significant events. Ensuring diversity is essential for reproducing realistic collective behaviors and is especially important in risk-sensitive applications such as training models for autonomous driving. A further challenge in the transportation domain is the definition of an appropriate probability distribution for complex, structured data. Consider travel survey data presented in Section~\ref{sec:tutorial_survey} as an example, where multiple constraints exist at different hierarchical levels \citep{sun2018hierarchical}: household composition follows specific structural patterns; individuals within the same household often share correlated socio-demographic features; and the sequence of trips must adhere to logical orderings, with interdependencies such as an adult's trip to drop off a child. Simplifying these relationships---often by assuming independence among household members---can ease computational challenges but may significantly compromise the quality and representativeness of the synthetic data.
}

\subsection{Dynamic Characteristics of Transportation data}\label{sec:5.2}

Mobility patterns and traffic data are continually evolving due to factors such as seasonal variations, changes in travel behavior, infrastructure updates, and technological advancements. These evolving conditions can lead to shifts in the underlying data distribution, posing challenges for DGMs trained on historical data. Such shifts can decrease the performance and reliability of the trained DGMs, especially in long-term deployments or applications where transportation patterns change over time. 
For instance, in tasks like driving scenario generation or population synthesis, the introduction of new vehicle types (e.g., electric or autonomous vehicles) or emerging transportation modes (e.g., shared micro-mobility) can alter driving behaviors and mode choice patterns, making the synthetic datasets generated by DGMs outdated. In anomaly data generation tasks, where anomalies are outliers in the data distribution (e.g., incidents, road blockages, or sudden traffic surges), dynamic changes in transportation can cause shifts in both normal and anomaly data distributions. This makes it harder to detect shifts in anomaly patterns and further complicates the distinction between normal and anomalous data. In estimation and prediction tasks, distribution shifts at different levels pose their own challenges. At the agent level, shifts in pedestrian trajectory data may occur due to behavioral changes driven by public health events, such as the COVID-19 pandemic. During such events, pedestrians tend to maintain larger physical distances, altering typical movement patterns and interactions in public spaces. Similarly, vehicle trajectory distributions can shift due to the introduction of autonomous vehicles or vehicle-to-vehicle communications that affect the way vehicles interact with traffic and can alter driving patterns. At the link level, new road construction or traffic control strategies can change the flow, speed, or density of links, making models trained on previous data less accurate. At the regional level, gradual shifts in population demographics, land use, or public transit services can alter travel demand, leading to outdated models for OD estimation or demand forecasting.

To enable DGMs to effectively manage and adapt to data distribution shifts in evolving transportation environments, strategies can be approached in two key steps: detecting changes in data distribution and updating the model to incorporate new datasets. 
The first step, detecting changes in data distribution, involves identifying when and how the characteristics of the data have evolved from what was seen during the model's training phase. This can be achieved using statistical tests, such as the Kolmogorov-Smirnov test for numerical features or the Chi-square test for categorical features, which assess whether significant differences exist between the distribution of new data and the original training dataset. However, these tests are typically used for single or pairs of features, and applying these tests to high-dimensional datasets can be computationally expensive and may not effectively capture interactions between features. In high-dimensional datasets, distance metrics like Wasserstein Distance, Jensen-Shannon Divergence, Kullback-Leibler Divergence, or Cosine Similarity can be used to quantitatively measure how far apart the distributions are. Additionally, visualization techniques such as histogram analysis and dimensionality reduction methods like t-distributed Stochastic Neighbor Embedding (t-SNE) or Principal Component Analysis (PCA) can provide visual insights into changes in feature distributions. DGMs themselves can also be leveraged to detect changes in data distributions by comparing the distributions of new data to those of the training data using the DGM's data generation and density estimation capabilities.
{

A key challenge in this process is the availability of labeled data to validate and monitor distribution shifts, as transportation datasets often suffer from label scarcity, especially in emerging behaviors and rare anomaly detection scenarios. Unsupervised methods and self-supervised learning may provide alternative solutions.
}

Once a shift in data distribution has been detected, the next step is to update the model to incorporate new trends. 
{

One crucial aspect is determining the appropriate time scale for model updates by considering the analysis time horizon. Some data shifts occur over short periods (e.g., traffic disruptions from special events), while others evolve gradually over years (e.g., shifts in travel demand due to changing land use or autonomous vehicle adoption). To address this, adaptive time-window analysis can be implemented, allowing DGMs to adjust their update frequency based on detected changes. For example, event-driven retraining enables rapid model updates following sudden changes (e.g., a major policy change), while periodic retraining (e.g., every six months) ensures models remain accurate in the face of gradual changes. 
}
Retraining the model from scratch with the new data can be computationally intensive and time-consuming, particularly for large datasets and complex models. An alternative approach is to employ incremental learning or continual learning methods, which update the model progressively as new data arrives, avoiding the need to retrain the entire dataset. Fine-tuning through transfer learning is another strategy, where a pre-trained model is adjusted to better align with the new distribution while retaining previously learned features. Fine-tuning involves updating a subset of the model parameters using a small amount of labeled data from the new distribution, providing a computationally efficient means to adapt to evolving data. 
{
However, model durability in evolving contexts remains a challenge, as continual updates can lead to catastrophic forgetting of prior knowledge. Techniques such as replay-based learning, memory-augmented models, or meta-learning approaches may offer potential solutions to improve long-term model stability.
}

\subsection{Deep Generative Models for Enhancing Transportation Data Privacy} \label{sec:5.3}
Synthetic data generation is one of the core applications of DGMs, as discussed in Section~\ref{sec:3.1.1}. The use of synthetic data generated by DGMs offers several significant advantages for privacy protection in transportation, highlighting the potential of synthetic data to balance the need for data utility with stringent privacy requirements.
One of the primary benefits of synthetic data generated by DGMs is its ability to preserve the statistical or distributional properties and patterns of real-world transportation and mobility data while excluding personally identifiable information. This means that the generated synthetic data can reflect the trends, behaviors, and characteristics found in actual data without risking individual privacy. For instance, in trajectory data containing personal origin-destination pairs and commuting patterns \citep{choi2018network}, synthetic data can accurately represent peak travel times, popular routes, and average journey durations without including any specific details about individual travelers.
This capability ensures that the generated synthetic data is useful for analysis, model training, and validation. Researchers and practitioners can derive meaningful insights for transportation policy and develop effective models for traffic management without compromising the quality and accuracy of the data. For example, urban planners can use synthetic data to simulate the impact of new infrastructure projects on traffic flow, while data scientists can train machine learning models to predict congestion or optimize public transport schedules.
Furthermore, synthetic data enables safe data sharing and collaboration between different organizations, stakeholders, and researchers. For instance, transportation agencies who collected privacy-sensitive trajectory data, activity sequence, and time use data, can share synthetic datasets generated by DGMs with third-party developers, policymakers, and academic institutions without exposing sensitive information. 
This can foster innovation, accelerate research, and support the development of new solutions and services, all while maintaining privacy.

\subsection{Addressing Data Bias for Equitable Data Generation} \label{sec:5.4}
The performance of DGMs heavily depends on the quality and representativeness of the training data. Ideally, Generative AIs can generate realistic and unbiased data if the training data itself is unbiased. However, in practice, training data often contains inherent biases, whether related to socioeconomic factors, geographic disparities, or historical inequalities. For example, if a generative model for urban planning is trained on data predominantly from affluent neighborhoods, the synthetic data it produces might overemphasize characteristics specific to those areas, neglecting the needs and patterns of lower-income regions. This could result in urban planning decisions that disproportionately benefit wealthier communities, thereby perpetuating or even amplifying existing disparities. As a result, ensuring that generative models do not amplify these biases is crucial for producing fair and representative synthetic data. This requires a multifaceted approach, beginning with the careful selection and preprocessing of training data. Data preprocessing steps might include normalizing data distributions, balancing underrepresented groups, and removing explicit bias indicators. For instance, transportation data, it might involve ensuring that data from various demographics and geographic regions are equally represented.

\subsection{Towards Trustworthy AI} \label{sec:5.5}


Beyond implementing technical measures, securing public and stakeholder acceptance and building trust in synthetic data presents significant challenges. As the use of synthetic data becomes more prevalent in AI systems, ensuring that these systems embody \textbf{trustworthy AI} principles is essential. Concerns may arise about the reliability and credibility of policies and decisions based on synthetic data. To promote trustworthy AI, it is crucial to maintain transparency in the documentation of data generation processes, open-sourcing code for reproducibility, and explaining the methodologies and safeguards used making them accessible to non-technical stakeholders. Additionally, demonstrating the validity of synthetic data through rigorous testing and comparison with real-world outcomes reinforces trustworthiness. For example, when using synthetic traffic data for new policy design, models should be able to replicate real traffic patterns under various conditions to demonstrate their adaptability and reliability. Fairness in synthetic data generation is equally important, as DGMs must be designed to avoid perpetuating biases inherent in training data. Employing fairness-aware algorithms and conducting regular audits of the synthetic data can help identify and mitigate biases, ensuring ethical and equitable outcomes. Moreover, involving stakeholders—such as community representatives and subject matter experts—in the development and implementation process further enhances credibility and acceptance. Their insights can help identify potential biases in data and modeling, as well as practical implications to ensure that synthetic data applications align with societal needs and ethical standards, thereby fostering trustworthy AI.


\subsection{Towards Causal Generative Models }\label{sec:5.6}
Causality refers to the relationship between cause and effect, where one event (the cause) directly influences another event (the effect). Assuming that event $\mathbf{x}$ is the direct cause of event $\mathbf{y}$, the causal relationship $\mathbf{x}\rightarrow \mathbf{y}$ signifies that altering $X$ can lead to a change in $\mathbf{y}$, whereas the reverse cannot hold true \citep{pearl1995causal,pearl2009causality}. The concept is fundamental in understanding how and why certain phenomena occur, especially in complex systems such as transportation networks. 
Different from traditional methods that rely on statistical inference, which usually calculates the conditional probability $p(\mathbf{y}|\mathbf{x})$ and cannot consider confounding bias, causal models focus on uncovering the true cause-and-effect relationships. In other words, these models can isolate the impact of specific variables and provide a deeper understanding of how changes in $\mathbf{x}$ directly lead to changes in $\mathbf{y}$ \citep{wagner1999causality,pearl2014external,bagi2023generative}.
In transportation research, causality has been actively considered in many areas such as traffic flow \citep{queen2009intervention,li2015robust,du2023spatial}, urban planning \citep{huang2022relationship,akbari2023spatial}, and traffic safety \citep{davis2000accident,elvik2011assessing,mannering2020big} to understand the impact of various interventions on outcomes. Studies in these areas evaluate the effects of variables that contribute to issues like traffic bottlenecks, vulnerabilities, and severe accidents. These applications demonstrate the value of causal inference in uncovering the true triggers of transportation phenomena, leading to more effective interventions and policy decisions. 
 


In transportation applications, integrating causality into DGMs can significantly enhance their effectiveness. Conventional generative models primarily rely on training data, which limits their effectiveness in new or changing environments. In contrast, causal generative models leverage underlying cause-effect relationships, enhancing their ability to generalize across different settings and maintain robustness in out-of-distribution (OOD) scenarios. By learning causality, these models can adapt to covariate shifts—where input data distributions vary between training and deployment—with fewer samples, reusing many of their components without the need for retraining. For example, a causal generative model trained to understand the relationship between traffic density and travel time in one city can adapt to another city with different traffic patterns. Additionally, causal generative models maintain reliable predictions under OOD conditions such as major social events, natural disasters, or unexpected road closures by understanding and leveraging causal structures. This dual capability of improved generalization and OOD robustness makes causal generative models particularly valuable for developing scalable and effective transportation solutions in diverse and dynamically changing environments, as discussed in Section~\ref{sec:5.2}.
Moreover, considering causality can improve the interpretability of DGMs. This capability not only helps stakeholders understand and trust the model's outputs but also allows them to control the generative process to produce targeted data. Traditional deep learning models are often perceived as black boxes, providing limited insights into the reasons behind certain outputs. Conversely, causal generative models focus on causal mechanisms, which offer clearer explanations for transportation phenomena. For example, in the context of traffic safety, a causal generative model can not only predict the likelihood of accidents at a particular intersection but also explain contributing factors to output such as poor road design, inadequate road signs, or high pedestrian activity. This interpretability is essential for developing trustworthy AI, as discussed in Section~\ref{sec:5.5} Since the model is interpretable based on these causal factors, it also allows for the control of the generative process by directly manipulating these variables. This kind of controllable generative model, grounded in causality, can be used to generate data representing specific scenarios, enabling transportation planners to simulate and evaluate the effects of interventions. This combination of interpretability and controllability makes causal generative models highly valuable in practical transportation applications.

\section{Conclusion}\label{sec:conclusion}
This review paper offers a comprehensive review of state-of-the-art DGMs, which are transforming how to understand the underlying patterns of increasingly large amount of data. This paper also offers a review of state-of-art research in various transportation-related topics using DGMs, examining the current landscape and practices in transportation research with DGMs. Furthermore, this paper includes an open-sourced tutorial, offering practical guidance and resources for those looking to explore and implement DGMs in transportation research and beyond. Finally, this paper offers discussions on potential challenges and opportunities related to utilizing, implementing, and developing DGMs in the transportation domain. The authors hope that this paper contributes to the advancement of the transportation academic field and promotes further research in DGMs.


\printcredits

\section*{Declaration of generative AI and AI-assisted technologies in the writing process}
During the preparation of this work, the authors used ChatGPT and Grammarly for language assistance. After utilizing these tools, the author(s) thoroughly reviewed and revised the content as necessary and take full responsibility for the final version of the published article.

\section*{Data and Code Availability Statement}
All data and code associated with this work are available at 
\begin{itemize}
    \item \url{https://github.com/UMN-Choi-Lab/DGMinTransportation}.
\end{itemize}

\bibliography{cas-sc-template.bib}
\bibliographystyle{cas-model2-names}

\clearpage
\appendix
\section{Preliminaries}\label{sec:prelim}

\begin{itemize}
\item \textbf{Bayes' Rule:} This is a fundamental theorem in probability theory that describes how to update the probability of a hypothesis based on new evidence. It combines prior knowledge (prior probability) with new evidence (likelihood) to form a posterior probability, which is a revised probability given the new information. Usually, in the context of DGMs, Bayes' Rule is frequently used to derive an approximated loss function. The equation for Bayes' Rule is: \begin{equation}
p(A|B) = \frac{p(B|A) p(A)}{p(B)},
\end{equation}
where \( p(A|B) \) is the posterior probability of the hypothesis \( A \) given the evidence \( B \), \( p(B|A) \) is the likelihood of observing the evidence \( B \) given that \( A \) is true, \( p(A) \) is the prior probability of the hypothesis \( A \) before observing the evidence, and \( p(B) \) is the probability of observing the evidence \( B \), also known as the marginal likelihood.
\item \textbf{Data Distribution:} This refers to the way data is spread or distributed in a particular space. Generative models aim to learn this distribution so they can generate new samples that resemble the original data.
\item \textbf{Likelihood:} Likelihood measures the plausibility of a model parameter value given specific observed data. In the context of generative models, it refers to how likely the observed data is under the model's assumed data distribution.
\item \textbf{KL divergence (Kullback–Leibler divergence; KLD):} This is a measure of how one probability distribution is different from a second, reference probability distribution. It provides a way to quantify the difference between two probability distributions in bits. Mathematically, given two probability distributions \( p(x) \) and \( q(x) \) over the same random variable \( x \), the KL divergence of \( q \) from \( p \) is defined as:

\begin{equation}
    D_{KL} \left(p(\mathbf{x}) || q(\mathbf{x}) \right) = \mathbbm{E}_{\mathbf{x}\sim p(\mathbf{x})} \left[\log\left(\frac{p(\mathbf{x})}{q(\mathbf{x})}\right)\right] = \int_{\mathbf{x} \in \mathcal{X}} p(\mathbf{x}) \log \left( \frac{p(\mathbf{x})}{q(\mathbf{x})} \right) dx , 
\end{equation}
where \( p(\mathbf{x}) \) and \( q(\mathbf{x}) \) are probability density functions (PDFs) over a continuous variable \( \mathbf{x} \). The integral is taken over the entire range of \( \mathbf{x} \) where the distributions are defined.
It is worth noting that KL divergence is non-negative and is not symmetric, meaning \( D_{KL}(p (\mathbf{x}) || q (\mathbf{x}) ) \neq D_{KL}(q (\mathbf{x}) || p (\mathbf{x}) ) \).
\item \textbf{Latent Space:} This is a conceptual space where the high-dimensional data is compressed into lower dimensions. The data in this latent space is usually the input to the generative part of the model that produces the output samples.
\end{itemize}

\subsection{Commonly used performance metrics}\label{app:metrics}

In what follows, let \( \{(y_i, \hat{y}_i)\}_{i=1}^n \) denote the set of true values \(y_i\) and corresponding predictions (estimation) \(\hat{y}_i\). We use \(\bar{y}\) to denote the sample mean of the true values and \(\bar{\hat{y}}\) to denote the sample mean of the predictions.

\subsubsection{Mean Absolute Error (MAE)}
\begin{equation}
\label{eq:mae}
\text{MAE} 
= \frac{1}{n} \sum_{i=1}^n \bigl|y_i - \hat{y}_i\bigr|.
\end{equation}

\subsubsection{Mean Relative Error (MRE) or Mean Absolute Percentage Error (MAPE)}
Sometimes referred to interchangeably (with slight differences in definition), a common version of MAPE is given by

\begin{equation}
\label{eq:mape}
\text{MAPE} 
= \frac{100\%}{n} \sum_{i=1}^n \left|\frac{y_i - \hat{y}_i}{y_i}\right|.
\end{equation}

For MRE, it is given by:
\begin{equation}
\label{eq:mre}
\text{MRE} 
= \frac{1}{n} \sum_{i=1}^n \left|\frac{y_i - \hat{y}_i}{y_i}\right|.
\end{equation}

\subsubsection{Symmetric Mean Absolute Percentage Error (SMAPE)}

\begin{equation}
\label{eq:smape}
\text{SMAPE} = \frac{100\%}{n} \sum_{i=1}^n \frac{|y_i - \hat{y}_i|}{\left(|y_i| + |\hat{y}_i|\right)/2},
\end{equation}

\subsubsection{Root Mean Squared Error (RMSE)}
\begin{equation}
\label{eq:rmse}
\text{RMSE} 
= \sqrt{\frac{1}{n} \sum_{i=1}^n (y_i - \hat{y}_i)^2}.
\end{equation}

\subsubsection{Normalized Root Mean Squared Error (NRMSE)}

The NRMSE is then given by normalizing the RMSE, commonly using the range of the observed data:
\begin{equation}
\label{eq:nrmse}
\text{NRMSE} = \frac{\text{RMSE}}{y_{\max} - y_{\min}},
\end{equation}
where \(y_{\max}\) and \(y_{\min}\) are the maximum and minimum true values, respectively.

\subsubsection{Standardized Root Mean Squared Error (SRMSE)}
A common normalization for the RMSE is to divide by a characteristic scale of the data (e.g., the mean of the true values):
\begin{equation}
\label{eq:srmse}
\text{SRMSE} 
= \frac{\text{RMSE}}{\bar{y}}
= \frac{\sqrt{\frac{1}{n}\sum_{i=1}^n (y_i - \hat{y}_i)^2}}{\frac{1}{n}\sum_{i=1}^n y_i}.
\end{equation}

\subsubsection{Pearson's Correlation Coefficient}
\begin{equation}
\label{eq:pearson_corr}
\rho 
= \frac{\sum_{i=1}^n \bigl(y_i - \bar{y}\bigr)\bigl(\hat{y}_i - \bar{\hat{y}}\bigr)}
{\sqrt{\sum_{i=1}^n (y_i - \bar{y})^2 \,\sum_{i=1}^n (\hat{y}_i - \bar{\hat{y}})^2}},
\end{equation}

\subsubsection{Continuous Ranked Probability Score (CRPS)}
For a probabilistic forecast \(F_i\) (cumulative distribution function) and an observation \(y_i\), the CRPS is defined as:
\begin{equation}
\label{eq:crps_single}
\text{CRPS}(F_i, y_i)
= \int_{-\infty}^{\infty} \Bigl[F_i(t) - \mathbf{1}\{t \ge y_i\}\Bigr]^2 \, dt,
\end{equation}
where \(\mathbf{1}\{\cdot\}\) denotes the indicator function. The average CRPS over \(n\) samples is then given by:
\begin{equation}
\label{eq:crps_mean}
\text{CRPS}_{sum} = \frac{1}{n} \sum_{i=1}^n \text{CRPS}(F_i, y_i).
\end{equation}

\subsubsection{Coefficient of Determination (R\textsuperscript{2})}
\begin{equation}
\label{eq:rsquare}
R^2
= 1 - \frac{\sum_{i=1}^n \bigl(y_i - \hat{y}_i\bigr)^2}{\sum_{i=1}^n \bigl(y_i - \bar{y}\bigr)^2}.
\end{equation}

\subsubsection{Relative Squared Error (RSE)}
In a regression context, it is defined as:
\begin{equation}
\label{eq:rse_regression}
\text{RSE} = \frac{\sum_{i=1}^n (y_i - \hat{y}_i)^2}{\sum_{i=1}^n (y_i - \bar{y})^2}.
\end{equation}
In a reconstruction context, for an original matrix or tensor \(X\) and its reconstruction \(\hat{X}\), the RSE is given by:
\begin{equation}
\label{eq:rse_reconstruction}
\text{RSE} = \frac{\|\hat{X} - X\|_F}{\|X\|_F},
\end{equation}
where the Frobenius norm is defined as \(\|A\|_F = \sqrt{\sum_{i,j} A_{ij}^2}\).

\subsubsection{Common Part of Commuting (CPC)}

The Common Part of Commuting (CPC) is a metric that quantifies the overlap between two flow distributions. Let $X_{i,j}$ represent the flow from location $i$ to location $j$ in the first distribution, and $Y_{i,j}$ the flow for the same OD pair in the second distribution. The CPC is defined as:
\begin{equation}
\label{eq:cpc_flow_general}
\text{CPC} 
= \frac{2 \sum_{i,j} \min\bigl(X_{i,j}, Y_{i,j}\bigr)}
       {\sum_{i,j} X_{i,j} + \sum_{i,j} Y_{i,j}}.
\end{equation}


\subsubsection{Performance Metrics (for classification)}



\subsubsection{Classification Metrics}

Let \(\text{TP}\) = True Positives, \(\text{FP}\) = False Positives, \(\text{FN}\) = False Negatives, and \(\text{TN}\) = True Negatives. The following metrics are defined as:

\begin{align}
\text{Precision} &= \frac{\text{TP}}{\text{TP} + \text{FP}}, \label{eq:precision}\\
\text{Recall (Sensitivity)} &= \frac{\text{TP}}{\text{TP} + \text{FN}}, \label{eq:recall}\\
\text{Specificity} &= \frac{\text{TN}}{\text{TN} + \text{FP}}, \label{eq:specificity}\\
\text{False Positive Rate (FPR)} &= \frac{\text{FP}}{\text{FP} + \text{TN}}, \label{eq:fpr}\\
\text{Accuracy} &= \frac{\text{TP} + \text{TN}}{\text{TP} + \text{TN} + \text{FP} + \text{FN}}, \label{eq:accuracy}\\
\text{Balanced Accuracy (BA)} &= \frac{1}{2}\Bigl(\frac{\text{TP}}{\text{TP} + \text{FN}} + \frac{\text{TN}}{\text{TN} + \text{FP}}\Bigr), \label{eq:ba}\\
\text{G-mean} &= \sqrt{\frac{\text{TP}}{\text{TP} + \text{FN}} \;\cdot\; \frac{\text{TN}}{\text{TN} + \text{FP}}}, \label{eq:gmean}\\
\text{F1-Score} &= 2 \cdot \frac{\text{Precision}\,\times\,\text{Recall}}{\text{Precision} + \text{Recall}}, \label{eq:f1}\\
\text{AUC} &= \text{Area Under the ROC Curve}. \label{eq:auc}
\end{align}

\subsubsection{mean Average Precision (mAP)}

 For each class, the Average Precision (AP) is calculated as the area under the Precision-Recall curve. The mAP is the average of these AP values across all classes:
\begin{equation}
\label{eq:map}
\text{mAP} = \frac{1}{C} \sum_{c=1}^{C} \text{AP}_c,
\end{equation}
where \(C\) is the total number of classes, and \(\text{AP}_c\) is the Average Precision for class \(c\).

\subsubsection{Cosine Similarity}
For two vectors \(\mathbf{u}\) and \(\mathbf{v}\) in \(\mathbb{R}^d\),
\begin{equation}
\label{eq:cosinesim}
\text{CosineSim}(\mathbf{u}, \mathbf{v})
= \frac{\mathbf{u} \cdot \mathbf{v}}{\|\mathbf{u}\|\|\mathbf{v}\|}.
\end{equation}

\subsubsection{BLEU Score}
The BLEU score is computed as
\begin{equation}
\text{BLEU Score} = BP \cdot \exp\left( \sum_{n=1}^{N} w_n \log p_n \right),
\end{equation}
where $p_n$ is the modified precision for n-grams, $w_n$ is the weight for each n-gram (typically $\frac{1}{N}$), and the brevity penalty $BP$ is defined as
\begin{equation}
BP = \begin{cases}
1, & \text{if } c > r, \\
\exp\left(1 - \frac{r}{c}\right), & \text{if } c \leq r,
\end{cases}
\end{equation}
with $c$ representing the length of the candidate translation and $r$ the effective reference length.

\subsubsection{METEOR Score}
The METEOR score incorporates precision, recall, and a penalty for fragmented matches. It is defined as:
\begin{equation}
\text{METEOR} = \left(1 - Penalty\right) \cdot F_{mean},
\end{equation}
where the harmonic mean $F_{mean}$ is computed as:
\begin{equation}
F_{mean} = \frac{10PR}{9P + R},
\end{equation}
with $P$ (precision) and $R$ (recall) based on unigram matches, and the penalty is given by:
\begin{equation}
Penalty = \gamma \left(\frac{ch}{m}\right)^\beta.
\end{equation}
Here, $ch$ denotes the number of chunks, $m$ is the number of mapped unigrams, and $\gamma$ and $\beta$ are parameters typically set by empirical tuning.

\subsubsection{Jensen-Shannon Divegence (JSD)}
The Jensen-Shannon distance measures the similarity between two probability distributions $P$ and $Q$. It is defined as:
\begin{equation}
JSD(P \parallel Q) = \sqrt{ \frac{1}{2} D_{KL}\left(P \parallel M\right) + \frac{1}{2} D_{KL}\left(Q \parallel M\right)},
\end{equation}
where
\begin{equation}
M = \frac{1}{2}(P + Q),
\end{equation}
and $D_{KL}$ represents the Kullback-Leibler divergence.

\subsubsection{Wasserstein Distance (WD)}

The Wasserstein Distance (WD), also known as the Earth Mover's Distance (EMD), measures the distance between two probability distributions over a space. For distributions \(\mu\) and \(\nu\) defined on a space \((\mathcal{X}, d)\), the \(p\)-Wasserstein Distance is given by,
\begin{equation}
\label{eq:wasserstein}
W_p(\mu, \nu) = \left( \inf_{\gamma \in \Gamma(\mu, \nu)} \int_{\mathcal{X} \times \mathcal{X}} d(x,y)^p \, d\gamma(x,y) \right)^{\frac{1}{p}},
\end{equation}
\noindent
where \(d(x,y)\) is the distance between points \(x\) and \(y\), and \(\Gamma(\mu, \nu)\) is the set of all ways to match the distributions \(\mu\) and \(\nu\).

\subsubsection{Fréchet Inception Distance (FID)}

The Fréchet Inception Distance (FID) is a metric used to assess the performance of generative models by comparing the distributions of feature representations extracted from real and generated images. Typically, these features are obtained using a pretrained Inception network. Assuming that the features are well-approximated by a multivariate Gaussian distribution, let \(\mu_r\) and \(\Sigma_r\) denote the mean and covariance of the features for real images, and \(\mu_g\) and \(\Sigma_g\) for generated images. The Fréchet Feature Distance (FFD) is then defined as:

\begin{equation}
\label{eq:ffd}
\text{FID} = \|\mu_r - \mu_g\|^2_2 + \operatorname{Tr}\Bigl(\Sigma_r + \Sigma_g - 2\Bigl(\Sigma_r \Sigma_g\Bigr)^{\frac{1}{2}}\Bigr),
\end{equation}
\noindent
where \(\|\mu_r - \mu_g\|^2_2\) represents the squared Euclidean Distance between the mean feature vectors, and the trace term \(\operatorname{Tr}(\cdot)\) measures the difference between the covariance matrices \(\Sigma_r\) and \(\Sigma_g\). Although this metric is commonly called the Fréchet Inception Distance when applied to images, it can be generalized to other domains—as the Fréchet Feature Distance—by employing an appropriate feature extractor.

\subsubsection{Segment-Path Distance (SPD) and Symmetrized Segment-Path Distance (SSPD)}

Let trajectory $T^1$ consist of $n_1$ points $\{p_i^1\}_{i=1}^{n_1}$. For each point $p_i^1$ in $T^1$, define $D_{\mathrm{pt}}(p_i^1, T^2)$ as the minimal distance from $p_i^1$ to any points of trajectory $T^2$. The SPD from $T^1$ to $T^2$ is defined as:
\begin{equation}
\label{eq:spd}
D_{\mathrm{SPD}}(T^1, T^2) = \frac{1}{n_1} \sum_{i=1}^{n_1} D_{\mathrm{pt}}(p_i^1, T^2).
\end{equation}
Since $D_{\mathrm{SPD}}(T^1, T^2)$ may differ from $D_{\mathrm{SPD}}(T^2, T^1)$, the Symmetrized Segment-Path Distance (SSPD) is defined as:
\begin{equation}
\label{eq:sspd}
D_{\mathrm{SSPD}}(T^1, T^2) = \frac{D_{\mathrm{SPD}}(T^1, T^2) + D_{\mathrm{SPD}}(T^2, T^1)}{2}.
\end{equation}

\subsubsection{Dynamic Time Warping (DTW)}

Let \(X = (x_1, x_2, \ldots, x_N)\) and \(Y = (y_1, y_2, \ldots, y_M)\) be two sequences, and let \(d(x_i, y_j)\) denote a local distance measure between elements \(x_i\) and \(y_j\). The DTW distance is defined recursively by:

\begin{equation}
\label{eq:dtw}
\text{DTW}(i, j) = d(x_i, y_j) + \min \Bigl\{
\text{DTW}(i-1, j), \;
\text{DTW}(i, j-1), \;
\text{DTW}(i-1, j-1)
\Bigr\},
\end{equation}

with the boundary conditions:

\begin{equation}
\label{eq:dtw_boundary}
\begin{aligned}
\text{DTW}(0, 0) &= 0, \\
\text{DTW}(i, 0) &= \infty \quad \text{for } i > 0, \\
\text{DTW}(0, j) &= \infty \quad \text{for } j > 0.
\end{aligned}
\end{equation}

The DTW distance between the full sequences \(X\) and \(Y\) is given by \(\text{DTW}(N, M)\).

\subsubsection{Maximum Final Distance (MFD)}

Let \(N\) be the number of agents, and for each agent \(i\), let the model generate \(K\) predicted trajectories. Denote the final position of the \(k\)-th predicted trajectory for agent \(i\) by \((x_T^{i,k}, y_T^{i,k})\). The indices \(k\) and \(l\) range over the set of predicted trajectories \(\{1, 2, \ldots, K\}\). For each agent \(i\), the MFD is calculated by taking the maximum Euclidean distance between the final positions of any pair of trajectories indexed by \(k\) and \(l\). Formally, the MFD over all agents is given by:

\begin{equation}
\label{eq:mfd}
\text{MFD}_K
= \frac{1}{N} \sum_{i=1}^N
\max_{k,l}
\sqrt{\,\bigl(x_T^{i,k} - x_T^{i,l}\bigr)^2 + \bigl(y_T^{i,k} - y_T^{i,l}\bigr)^2}.
\end{equation}

\subsubsection{Average Displacement Error (ADE) and minADE\(_k\)}
Given the ground truth trajectory \(\{\mathbf{q}_1, \dots, \mathbf{q}_T\}\) and a predicted trajectory \(\{\hat{\mathbf{q}}_1, \dots, \hat{\mathbf{q}}_T\}\), the ADE is defined as:
\begin{equation}
\label{eq:ADE}    
\text{ADE} = \frac{1}{T} \sum_{t=1}^{T} \|\hat{\mathbf{q}}_t - \mathbf{q}_t\|_2.
\end{equation}
When \(k\) different predictions \(\{\hat{\mathbf{q}}^{(i)}_1, \dots, \hat{\mathbf{q}}^{(i)}_T\}\) are available, we define
\begin{equation}
\label{eq:minADE}  
\text{minADE}_k = \min_{i\in\{1,\dots,k\}} \frac{1}{T} \sum_{t=1}^{T} \|\hat{\mathbf{q}}^{(i)}_t - \mathbf{q}_t\|_2.
\end{equation}

\subsubsection{Final Displacement Error (FDE) and minFDE\(_k\)}
For the final time step, the FDE is given by:
\begin{equation}
\label{eq:FDE}  
\text{FDE} = \|\hat{\mathbf{q}}_T - \mathbf{q}_T\|_2.
\end{equation}
Similarly, for \(k\) predictions,
\begin{equation}
\label{eq:minFDE} 
\text{minFDE}_k = \min_{i\in\{1,\dots,k\}} \|\hat{\mathbf{q}}^{(i)}_T - \mathbf{q}_T\|_2.
\end{equation}

\subsubsection{Kernel Density Estimate-based Negative Log Likelihood (KDENLL)}

Let \(\{\hat{\mathbf{q}}^{(i)}_t\}_{i=1}^k\) denote the \(k\) predicted positions at time step \(t\), and let \(\mathbf{q}_t\) be the corresponding ground truth position. Here, \(T\) denotes the total number of time steps in the trajectory. The Kernel Density Estimate (KDE) at \(\mathbf{q}_t\) is computed as:
\begin{equation}
\label{eq:KDE}  
\hat{p}(\mathbf{q}_t) = \frac{1}{k h^d} \sum_{i=1}^k K\!\left(\frac{\mathbf{q}_t - \hat{\mathbf{q}}^{(i)}_t}{h}\right),
\end{equation}
where \(K(\cdot)\) is a kernel function (commonly a Gaussian), \(h\) is the bandwidth, and \(d\) is the dimensionality of \(\mathbf{q}_t\).

The KDENLL for the trajectory is then defined as:
\begin{equation}
\label{eq:KDENLL}  
\text{KDENLL} = -\frac{1}{T} \sum_{t=1}^T \log \hat{p}(\mathbf{q}_t).
\end{equation}

When \(k\) distinct sets of predictions are available, one can compute a KDENLL for each set. Let \(\hat{p}^{(i)}(\mathbf{q}_t)\) be the KDE computed using the \(i\)-th prediction set (for \(i=1,\dots,k\)). The best (lowest) KDENLL among these is then given by
\begin{equation}
\label{eq:KDENLL_k}  
\text{KDENLL}_k = \min_{i \in \{1,\dots,k\}} \left[-\frac{1}{T} \sum_{t=1}^T \log \hat{p}^{(i)}(\mathbf{q}_t)\right].
\end{equation}

\subsubsection{Maximum Mean Discrepancy (MMD)}

The Maximum Mean Discrepancy (MMD) is a kernel-based measure for comparing two distributions using sample data. Let $\{x_i\}_{i=1}^n$ and $\{y_j\}_{j=1}^m$ be samples from distributions $P$ and $Q$, respectively, and let $k(\cdot,\cdot)$ be a positive-definite kernel (e.g., the RBF kernel). The squared MMD is given by:
\begin{equation}
\label{eq:mmd}
\text{MMD}(P, Q) 
= \frac{1}{n^2}\sum_{i=1}^n \sum_{j=1}^n k(x_i, x_j)
+ \frac{1}{m^2}\sum_{i=1}^m \sum_{j=1}^m k(y_i, y_j)
- \frac{2}{nm}\sum_{i=1}^n \sum_{j=1}^m k(x_i, y_j).
\end{equation}





\section{Derivation for Evidence Lower Bound for Variational Autoencoder}\label{app:vae}

The probability $p_{\theta}(\mathbf{x})$ with the latent vector $\mathbf{z}$ can be formulated as:
\begin{equation}\label{eq:vae_likelihood}
    p (\mathbf{x}) = \int p_{\theta} (\mathbf{x}|\mathbf{z}) p{(\mathbf{z})}  d\mathbf{z}.
\end{equation}

However, Equation~\eqref{eq:vae_likelihood} is intractable due to the integral over the $\mathbf{z}$ space. Even though Monte Carlo methods can be used to approximate the integral and apply maximum likelihood estimation, they may result in suboptimal generation due to poor-quality samples being assigned higher likelihoods. To address this issue, VAEs incorporate the encoder $q_{\phi} (\mathbf{z} | \mathbf{x})$ into their objective function, reformulated as:
%
\begin{equation}
\begin{split}
\log p(\mathbf{x}) 
& = \mathbbm{E}_{\mathbf{z} \sim q_{\phi} (\mathbf{z} | \mathbf{x})} \left[ \log p(\mathbf{x})  \right]  \\
& = \mathbbm{E}_{\mathbf{z}} \left[ \log \frac{p_{\theta}(\mathbf{x}, \mathbf{z})}{p_{\theta}(\mathbf{z} | \mathbf{x})} \right]  && \Rightarrow  
    \textit{ \small Bayes' Rule} \\
& = \mathbbm{E}_{\mathbf{z}} \left[ \log \frac{p_{\theta}(\mathbf{x}| \mathbf{z}) p (\mathbf{z}) }{p_{\theta}(\mathbf{z} | \mathbf{x})}\right]  \\
& = \mathbbm{E}_{\mathbf{z}} \left[ \log \frac{p_{\theta}(\mathbf{x}| \mathbf{z}) p (\mathbf{z}) }{p_{\theta}(\mathbf{z} | \mathbf{x})} \frac{q_{\phi} (\mathbf{z}|\mathbf{x})}{q_{\phi} (\mathbf{z}|\mathbf{x})}\right]   && \Rightarrow  
    \textit{ \small Multiply by $\frac{q_{\phi} (\mathbf{z}|\mathbf{x})}{q_{\phi} (\mathbf{z}|\mathbf{x})}=1$} \\ 
& = \mathbbm{E}_{\mathbf{z}} \left[ \log p_{\theta} (\mathbf{x} | \mathbf{z}) \right]
- \mathbbm{E}_{\mathbf{z}} \left[ \log \frac{q_{\phi} (\mathbf{z}|\mathbf{x})}{p (\mathbf{z})}\right]
+ \mathbbm{E}_{\mathbf{z}} \left[ \log \frac{q_{\phi} (\mathbf{z}|\mathbf{x})}{p_{\theta} (\mathbf{z} | \mathbf{x})} \right] 
&& \Rightarrow \textit{ \small Logarithms} \\ 
& = \mathbbm{E}_{\mathbf{z}}  \left[ \log p_{\theta} (\mathbf{x} | \mathbf{z}) \right]
- D_{KL} \left( q_{\phi} \left(\mathbf{z}|\mathbf{x} \right) || p (\mathbf{z})  \right)
+ D_{KL} \left( q_{\phi} \left(\mathbf{z}|\mathbf{x} \right) || p_{\theta} (\mathbf{z} | \mathbf{x} )  \right)
.
\end{split}\label{eq:vae_derivation}
\end{equation}

The first term in Equation~\eqref{eq:vae_derivation}, $\mathbbm{E}_{z} \left[ \log p_{\theta} (\mathbf{x} | \mathbf{z}) \right] $, is the likelihood of $\mathbf{x}$ generated from the decoder based on the sampled $\mathbf{z}$ from the posterior distribution. This term can be estimated through sampling using the reparameterization trick, which will be discussed later.
The second term is the KL divergence between the approximate posterior distribution $q_{\phi} (\mathbf{z} | \mathbf{x})$ and the prior distribution $p (\mathbf{z})$. Assuming both distributions follow a tractable distribution (typically Gaussian), this term can be computed in closed form.
The third term is intractable because the true posterior distribution $p_{\theta} (\mathbf{z} | \mathbf{x})$ involves evaluation of intractable Equation~\eqref{eq:vae_derivation} $\left( p_{\theta} (\mathbf{z} | \mathbf{x}) = \frac{p_{\theta} (\mathbf{x} | \mathbf{z}) p (\mathbf{z}) }{ { p_{\theta} (\mathbf{x})} } \right)$. However, the third term is always non-negative due to the properties of KL divergence.

\section{Discussion on implicit likelihood maximization of GANs}\label{sec:discuss_gan}
Consider $p_{\text{model}}$ as the learned model distribution imposed by the $\mathbf{z}\sim p(\mathbf{z})$ and the $\theta$-parameterized mapping function $G_{\theta}$. If $G_{\theta}$ is fixed, then $D$ would converge to $D_{\phi}^* (\mathbf{x})= \frac{p_{\text{model}}(\mathbf{x})}{p_{\text{data}}(\mathbf{x}) + p_{\text{model}}(\mathbf{x})}$, since
\begin{equation}
\begin{split}
& \max_{\phi} \mathbbm{E}_{\mathbf{x} \sim p_{\text{data}}(\mathbf{x})}[\log D_{\phi}(\mathbf{x})] + \mathbbm{E}_{\mathbf{x} \sim p_{\text{model}}(\mathbf{x})}[\log(1 - D_{\phi}(\mathbf{x}))]    \\
= & \max_{\phi} \int_{\mathbf{x}} \left( p_{\text{data}}  (\mathbf{x}) \log D_{\phi}(\mathbf{x}) + p_{\text{model}} (\mathbf{x}) \log \left(1 - D_{\phi}(\mathbf{x}) \right) \right) d\mathbf{x},
\end{split}
\end{equation}
which is optimal when $D_{\phi}^* (\mathbf{x})= \frac{p_{\text{model}}(\mathbf{x})}{p_{\text{data}}(\mathbf{x}) + p_{\text{model}}(\mathbf{x})}$.

For the optimal discriminator, the objective of the generator can be expressed as
\begin{equation}
\begin{split} 
&\min_{G}V(D_{\phi}^{*}, G_{\theta})  \\
= & \min_{\theta} \mathbbm{E}_{\mathbf{x} \sim p_{\text{data}}(\mathbf{x})}[\log D^*_{\phi}(\mathbf{x})] + \mathbbm{E}_{\mathbf{x} \sim p_{\text{model}}(\mathbf{x})}[\log(1 - D^*_{\phi}(\mathbf{x}))]    \\
= & \min_{\theta} \int_{\mathbf{x}} \left( p_{\text{data}}  (\mathbf{x}) \log D^*_{\phi}(\mathbf{x}) + p_{\text{model}} (\mathbf{x}) \log \left(1 - D^*_{\phi}(\mathbf{x}) \right) \right) d\mathbf{x}  \\
= & \min_{\theta} \int_{\mathbf{x}} \left( p_{\text{data}}  (\mathbf{x}) \log \frac{p_{\text{model}}(\mathbf{x})}{p_{\text{data}}(\mathbf{x}) + p_{\text{model}}(\mathbf{x})} + p_{\text{model}} (\mathbf{x}) \log \left(1 - \frac{p_{\text{model}}(\mathbf{x})}{p_{\text{data}}(\mathbf{x}) + p_{\text{model}}(\mathbf{x})} \right) \right) d\mathbf{x} \\
= & \min_{\theta} \int_{\mathbf{x}} \left( p_{\text{data}}  (\mathbf{x}) \log \frac{p_{\text{model}}(\mathbf{x})}{p_{\text{data}}(\mathbf{x}) + p_{\text{model}}(\mathbf{x})} + p_{\text{model}} (\mathbf{x}) \log \frac{p_{\text{data}}(\mathbf{x})}{p_{\text{data}}(\mathbf{x}) + p_{\text{model}}(\mathbf{x})} \right) d\mathbf{x} \\
= & \min_{\theta} \int_{\mathbf{x}} \left( p_{\text{data}}  (\mathbf{x}) \log \frac{p_{\text{model}}(\mathbf{x})}{(p_{\text{data}}(\mathbf{x})  + p_{\text{model}}(\mathbf{x}))/2} - p_{\text{data}} (\mathbf{x}) \log 2 + p_{\text{model}} (\mathbf{x}) \log \frac{p_{\text{data}}(\mathbf{x})}{p_{\text{data}}(\mathbf{x}) + p_{\text{model}}(\mathbf{x})/2} - p_{\text{model}} (\mathbf{x}) \log 2\right) d\mathbf{x}  \\
= & \min_{\theta} D_{KL} \left(p_{\text{model}} (\mathbf{x}) || \frac{(p_{\text{data}}(\mathbf{x})  + p_{\text{model}}(\mathbf{x}))}{2} \right) + D_{KL} \left(p_{\text{data}} (\mathbf{x}) || \frac{(p_{\text{data}}(\mathbf{x})  + p_{\text{model}}(\mathbf{x}))}{2} \right) -2 \log 2\\
= & \min_{\theta} 2 D_{JS} \left(p_{\text{model}} (\mathbf{x}) || p_{\text{data}}(\mathbf{x}) \right) -2 \log 2 ,
\end{split}
\end{equation} 
where $D_{JS}$ is the Jensen-Shannon divergence. The minimization is achieved when $D_{JS} \left(p_{\text{model}} (\mathbf{x}) || p_{\text{data}}(\mathbf{x}) \right)$ is minimized, i.e., the model distribution $p_{\text{model}}$ is close to the data distribution $p_{\text{data}}$.

The training process continues until a balance between the Generator and the Discriminator is achieved, where the Generator can produce realistic data and the Discriminator cannot differentiate between real and generated data, i.e., $D^*(\mathbf{x}) =\frac{p_{\text{model}}(\mathbf{x})}{p_{\text{data}}(\mathbf{x}) + p_{\text{model}}(\mathbf{x})}=\frac{p_{\text{data}}(\mathbf{x})}{p_{\text{data}}(\mathbf{x}) + p_{\text{data}}(\mathbf{x})} = 0.5$, for all $\mathbf{x}$.

However, in practice, achieving equilibrium is challenging due to issues like mode collapse, vanishing gradients, and non-convergence. Various modifications and extensions of the original GAN framework have been proposed to mitigate these issues, such as Wasserstein GANs \citep{arjovsky2017wasserstein}, Least-Squares GANs \citep{mao2017least}, and Conditional GANs \citep{shahbazi2022collapse}, each offering unique perspectives and solutions. Despite these challenges, the ability to generate highly realistic and diverse data has made GAN-based models essential tools in many domains of machine learning and data science.

\section{Derivation for Diffusion Model}\label{appendix:diffusion_derivation}

Similar to VAE and Normalizing Flows (models using explicit densities for training), the training objective is to maximize the likelihood, or minimize the negative log-likelihood as follows:
\begin{equation}\label{eq:diffusion_objective2}
    \theta^* = \arg\min_\theta \mathbbm{E} \left[ - \log p_\theta \left( \mathbf{x}_0 \right) \right]
\end{equation}

Similar to VAE, instead of directly training the model with negative log-likelihood, we can consider $q$ as the approximate posterior and use the variational bound on negative log-likelihood to train the model as follows:
\begin{equation}
    \begin{split}
& \mathbbm{E} \left[ - \log p_\theta \left( \mathbf{x}_0 \right) \right]   , \\
& = \mathbbm{E} \left[ - \log \frac{p_\theta \left( \mathbf{x}_0 , \mathbf{x}_1 , \cdots ,  \mathbf{x}_T \right)}{p_\theta \left( \mathbf{x}_1 , \cdots ,  \mathbf{x}_T  | \mathbf{x}_0 \right)} \right], && \Rightarrow \text{Baye's Rule} \\
& = \mathbbm{E} \left[ - \log \frac{p_\theta \left( \mathbf{x}_{0:T} \right)}{p_\theta \left(  \mathbf{x}_{1:T}   | \mathbf{x}_0 \right)} \cdot \frac{ q (\mathbf{x}_{1:T} | \mathbf{x}_0 ) }{q (\mathbf{x}_{1:T} | \mathbf{x}_0 )} \right] , \\
& = \mathbbm{E} \left[ - \log \frac{p_\theta \left( \mathbf{x}_{0:T} \right)}{q (\mathbf{x}_{1:T} | \mathbf{x}_0 )} \cdot \frac{ q (\mathbf{x}_{1:T} | \mathbf{x}_0 ) }{p_\theta \left(  \mathbf{x}_{1:T}   | \mathbf{x}_0 \right)} \right] = \mathbbm{E} \left[ - \log \frac{p_\theta \left( \mathbf{x}_{0:T} \right)}{q (\mathbf{x}_{1:T} | \mathbf{x}_0 )} \right] - \mathbbm{E} \left[  \log \frac{ q (\mathbf{x}_{1:T} | \mathbf{x}_0 ) }{p_\theta \left(  \mathbf{x}_{1:T}   | \mathbf{x}_0 \right)} \right]  ,\\
& \leq \mathbbm{E} \left[ - \log \frac{p_\theta \left( \mathbf{x}_{0:T} \right)}{q (\mathbf{x}_{1:T} | \mathbf{x}_0 )}  \right], && \Rightarrow \text{KL divergence } \geq 0  \\
    \end{split}
\end{equation}

Using Equation~\eqref{eq:diffusion_q} and Equation~\eqref{eq:diffusion_p}, we can further derive:
\begin{equation}
\begin{split}
& \mathbbm{E} \left[ - \log \frac{p_\theta \left( \mathbf{x}_{0:T} \right)}{q (\mathbf{x}_{1:T} | \mathbf{x}_0 )}  \right] , \\
& = \mathbbm{E} \left[ - \log \frac{  p \left(\mathbf{x}_T \right) \prod_{t=1}^T p_\theta \left( \mathbf{x}_{t-1} | \mathbf{x}_t \right) }{ \prod_{t=1}^T q\left(\mathbf{x}_t | \mathbf{x}_{t-1}\right)}  \right] , \\
& = \mathbbm{E} \left[ - \log  p \left(\mathbf{x}_T \right) - \log \frac{  \prod_{t=1}^T p_\theta \left( \mathbf{x}_{t-1} | \mathbf{x}_t \right) }{ \prod_{t=1}^T q\left(\mathbf{x}_t | \mathbf{x}_{t-1}\right)}  \right] , \\
& = \mathbbm{E} \left[ - \log  p \left(\mathbf{x}_T \right) - \sum_{t=1}^T \log \frac{  p_\theta \left( \mathbf{x}_{t-1} | \mathbf{x}_t \right) }{ q\left(\mathbf{x}_t | \mathbf{x}_{t-1}\right)}  \right] , \\
\end{split}
\end{equation}

Appendix A in \cite{ho2020denoising} further provides a detailed derivation of the reduced variance variational bound for diffusion models as follows:
\begin{equation}
\begin{split}
& \mathbbm{E} \left[ - \log  p \left(\mathbf{x}_T \right) - \sum_{t=1}^T \log \frac{  p_\theta \left( \mathbf{x}_{t-1} | \mathbf{x}_t \right) }{ q\left(\mathbf{x}_t | \mathbf{x}_{t-1}\right)}  \right] , \\
& = \mathbbm{E} \left[ - \log  p \left(\mathbf{x}_T \right) - \sum_{t=2}^T \log \frac{  p_\theta \left( \mathbf{x}_{t-1} | \mathbf{x}_t \right) }{ q\left(\mathbf{x}_t | \mathbf{x}_{t-1}\right)}  -  \log \frac{p_\theta \left(\mathbf{x}_{0} | \mathbf{x}_1 \right)}{q\left(\mathbf{x}_1 | \mathbf{x}_0 \right)}  \right] , \\
& = \mathbbm{E} \left[ - \log  p \left(\mathbf{x}_T \right) - \sum_{t=2}^T \log \frac{  p_\theta \left( \mathbf{x}_{t-1} | \mathbf{x}_t \right) }{ q\left(\mathbf{x}_{t-1} | \mathbf{x}_{t} , \mathbf{x}_0 \right)} \cdot \frac{q\left(\mathbf{x}_{t-1} | \mathbf{x}_0 \right)}{q\left(\mathbf{x}_{t} | \mathbf{x}_0 \right)}  -  \log \frac{p_\theta \left(\mathbf{x}_{0} | \mathbf{x}_1 \right)}{q\left(\mathbf{x}_1 | \mathbf{x}_0 \right)}  \right] ,\\
\end{split}
\end{equation}
since
\begin{equation}
\begin{split}
q\left(\mathbf{x}_t | \mathbf{x}_{t-1}\right)  
& = q\left(\mathbf{x}_t | \mathbf{x}_{t-1} , \mathbf{x}_0 \right) , && \Rightarrow \text{Markov chain property} \\
& = \frac{q\left(\mathbf{x}_t , \mathbf{x}_{t-1} , \mathbf{x}_0 \right)}{ q\left( \mathbf{x}_{t-1} , \mathbf{x}_0 \right)}  = \frac{q\left(\mathbf{x}_t , \mathbf{x}_{t-1} , \mathbf{x}_0 \right)}{ q\left( \mathbf{x}_{t-1} , \mathbf{x}_0 \right)} \cdot \frac{q\left( \mathbf{x}_t , \mathbf{x}_0 \right)}{q\left( \mathbf{x}_t , \mathbf{x}_0 \right)} , \\ 
& = \frac{q\left(\mathbf{x}_t , \mathbf{x}_{t-1} , \mathbf{x}_0 \right)}{ q\left( \mathbf{x}_t , \mathbf{x}_0 \right)} \cdot \frac{q\left( \mathbf{x}_t , \mathbf{x}_0 \right)}{q\left( \mathbf{x}_{t-1} , \mathbf{x}_0 \right)} ,\\ 
& = q\left(\mathbf{x}_{t-1} | \mathbf{x}_{t} , \mathbf{x}_0 \right) \cdot \frac{q\left( \mathbf{x}_t , \mathbf{x}_0 \right)}{q\left( \mathbf{x}_{t-1} , \mathbf{x}_0 \right)} . \\ 
\end{split}
\end{equation}

As a result,

\begin{equation}
\begin{split}
& \mathbbm{E} \left[ - \log  p \left(\mathbf{x}_T \right) - \sum_{t=2}^T \log \frac{  p_\theta \left( \mathbf{x}_{t-1} | \mathbf{x}_t \right) }{ q\left(\mathbf{x}_{t-1} | \mathbf{x}_{t} , \mathbf{x}_0 \right)} \cdot \frac{q\left(\mathbf{x}_{t-1} | \mathbf{x}_0 \right)}{q\left(\mathbf{x}_{t} | \mathbf{x}_0 \right)}  -  \log \frac{p_\theta \left(\mathbf{x}_{0} | \mathbf{x}_1 \right)}{q\left(\mathbf{x}_1 | \mathbf{x}_0 \right)}  \right] , \\
& =  \mathbbm{E} \left[ - \log  p \left(\mathbf{x}_T \right) - \sum_{t=2}^T \log \frac{  p_\theta \left( \mathbf{x}_{t-1} | \mathbf{x}_t \right) }{ q\left(\mathbf{x}_{t-1} | \mathbf{x}_{t} , \mathbf{x}_0 \right)}  - \sum_{t=2}^T \log \frac{q\left(\mathbf{x}_{t-1} | \mathbf{x}_0 \right)}{q\left(\mathbf{x}_{t} | \mathbf{x}_0 \right)}  -  \log \frac{p_\theta \left(\mathbf{x}_{0} | \mathbf{x}_1 \right)}{q\left(\mathbf{x}_1 | \mathbf{x}_0 \right)}  \right] , \\
& =  \mathbbm{E} \left[ - \log  p \left(\mathbf{x}_T \right) - \sum_{t=2}^T \log \frac{  p_\theta \left( \mathbf{x}_{t-1} | \mathbf{x}_t \right) }{ q\left(\mathbf{x}_{t-1} | \mathbf{x}_{t} , \mathbf{x}_0 \right)}  + \log q\left(\mathbf{x}_{T} | \mathbf{x}_0 \right) - \log p_\theta \left(\mathbf{x}_{0} | \mathbf{x}_1 \right)  \right] , \\
\end{split}
\end{equation}
since
\begin{equation}
\begin{split}
& -\sum_{t=2}^T \log \frac{q\left(\mathbf{x}_{t-1} | \mathbf{x}_0 \right)}{q\left(\mathbf{x}_{t} | \mathbf{x}_0 \right)}  -  \log \frac{p_\theta \left(\mathbf{x}_{0} | \mathbf{x}_1 \right)}{q\left(\mathbf{x}_1 | \mathbf{x}_0 \right)} , \\
& = -\log \frac{q\left(\mathbf{x}_{1} | \mathbf{x}_0 \right)}{q\left(\mathbf{x}_{2} | \mathbf{x}_0 \right)} \cdot \frac{q\left(\mathbf{x}_{2} | \mathbf{x}_0 \right)}{q\left(\mathbf{x}_{3} | \mathbf{x}_0 \right)}\cdot \frac{q\left(\mathbf{x}_{3} | \mathbf{x}_0 \right)}{q\left(\mathbf{x}_{4} | \mathbf{x}_0 \right)} \cdots \frac{q\left(\mathbf{x}_{T-1} | \mathbf{x}_0 \right)}{q\left(\mathbf{x}_{T} | \mathbf{x}_0 \right)} \cdot \frac{p_\theta \left(\mathbf{x}_{0} | \mathbf{x}_1 \right)}{q\left(\mathbf{x}_1 | \mathbf{x}_0 \right)} , \\
& = \log q\left(\mathbf{x}_{T} | \mathbf{x}_0 \right) - \log p_\theta \left(\mathbf{x}_{0} | \mathbf{x}_1 \right) . \\
\end{split}
\end{equation}

Therefore, 
\begin{equation}\label{eq:diffusion_loss}
\begin{split}
& \mathbbm{E} \left[ - \log  p \left(\mathbf{x}_T \right) - \sum_{t=2}^T \log \frac{  p_\theta \left( \mathbf{x}_{t-1} | \mathbf{x}_t \right) }{ q\left(\mathbf{x}_{t-1} | \mathbf{x}_{t} , \mathbf{x}_0 \right)}  + \log q\left(\mathbf{x}_{T} | \mathbf{x}_0 \right) - \log p_\theta \left(\mathbf{x}_{0} | \mathbf{x}_1 \right)  \right] , \\
& =  \mathbbm{E} \left[  \log  \frac{q\left(\mathbf{x}_{T} | \mathbf{x}_0 \right)}{p \left(\mathbf{x}_T \right)} - \sum_{t=2}^T \log \frac{  p_\theta \left( \mathbf{x}_{t-1} | \mathbf{x}_t \right) }{ q\left(\mathbf{x}_{t-1} | \mathbf{x}_{t} , \mathbf{x}_0 \right)}    -  \log p_\theta \left(\mathbf{x}_{0} | \mathbf{x}_1 \right) \right] , \\
& =  \mathbbm{E} \left[ 
\underbrace{D_{KL} \left( q\left(\mathbf{x}_{T} | \mathbf{x}_0 \right) || p \left(\mathbf{x}_T \right) \right) }_{L_T}
+ \sum_{t=2}^T \underbrace{D_{KL} \left(  q\left(\mathbf{x}_{t-1} | \mathbf{x}_{t} , \mathbf{x}_0 \right)  || p_\theta \left( \mathbf{x}_{t-1} | \mathbf{x}_t \right) \right) }_{L_{t-1}}
\underbrace{- \log p_\theta \left(\mathbf{x}_{0} | \mathbf{x}_1 \right)}_{L_0}
\right] . \\
\end{split}
\end{equation}

Therefore, the overall loss function of minimizing the negative log-likelihood in Equation~\eqref{eq:diffusion_objective2} is decomposed into several losses, $L_T$, $L_{t-1}$, and $L_0$. Here, $L_T$ is constant since both $q\left(\mathbf{x}_{T} | \mathbf{x}_0 \right) $ and $p \left(\mathbf{x}_T \right) $ are fixed, and therefore, we can ignore this term. Also, in \cite{ho2016generative}, $L_0$ is explicitly defined by using the characteristics of the image generation problem, and as a result, $L_0$ can be interpreted as a reconstruction loss of a problem-specific decoder. As a result, the actual learning process of the diffusion model is related to $L_{t-1}$.

$L_{t-1}$ measures the KL-divergence of $q\left(\mathbf{x}_{t-1} | \mathbf{x}_{t} , \mathbf{x}_0 \right)$ from $p_\theta \left( \mathbf{x}_{t-1} | \mathbf{x}_t \right)$. The diffusion process, $q$, represents the process of adding small noise to the data; i.e., given a less noisy data $\mathbf{x}_{t-1}$, the distribution of a more noisy data $\mathbf{x}_{t}$. The first term, $q\left(\mathbf{x}_{t-1} | \mathbf{x}_{t}, \mathbf{x}_0 \right)$, represents the true denoising process which is derived from the definition of $q$ given the true data without noise, $\mathbf{x}_0$. What the diffusion models try to learn is the denoising process $p_\theta \left( \mathbf{x}_{t-1} | \mathbf{x}_t \right)$; i.e., given a more noisy data $\mathbf{x}_{t}$, the distribution of a less noisy data $\mathbf{x}_{t-1}$. As a result, $L_{t-1}$ captures the distributional difference between the true denoising process (given the true data) and the approximated denoising process (without the true data).

Since the diffusion process follows Gaussian distribution, the true reverse process,  $q\left(\mathbf{x}_{t-1} | \mathbf{x}_{t}, \mathbf{x}_0 \right)$, can be assumed to follow a Gaussian distribution if $T$ is sufficiently large, or $T\rightarrow \infty$. Let $q\left(\mathbf{x}_{t-1} | \mathbf{x}_{t}, \mathbf{x}_0 \right) = \mathcal{N} (\mathbf{x}_{t-1} ; \Tilde{\boldsymbol{\mu}}_t (\mathbf{x}_t, \mathbf{x}_0) , \Tilde{\beta}_t \mathbbm{I})$. To derive explicit form of $\Tilde{\boldsymbol{\mu}}_t (\mathbf{x}_t, \mathbf{x}_0)$ and $\Tilde{\beta}_t$, first, we should derive a closed-form equation for sampling $\mathbf{x}_t$ at an arbitrary timestep $t$ from Equation~\eqref{eq:diffusion_forward} as follows:
\begin{equation}
\begin{split}
\mathbf{x}_t 
& = \sqrt{1-\beta_t} \mathbf{x}_{t-1} + \sqrt{\beta_t} \boldsymbol{\epsilon}_t ,&& \\
& = \sqrt{\alpha_t} \mathbf{x}_{t-1} + \sqrt{1-\alpha_t} \boldsymbol{\epsilon}_t ,&& \Rightarrow \text{Let } \alpha_t = 1-\beta_t\\
& = \sqrt{\alpha_t} \left(  \sqrt{\alpha_{t-1}} \mathbf{x}_{t-2} + \sqrt{1-\alpha_{t-1}} \boldsymbol{\epsilon}_{t-2} \right) + \sqrt{1-\alpha_t} \boldsymbol{\epsilon}_t ,&& \\
& = \sqrt{\alpha_t \alpha_{t-1}} \mathbf{x}_{t-2} + \sqrt{1-\alpha_t \alpha_{t-1}} \Bar{\boldsymbol{\epsilon}}_t ,&& \Rightarrow \Bar{\boldsymbol{\epsilon}}_t \text{ merges two Gaussians } (\boldsymbol{\epsilon}_t \text{ and } \boldsymbol{\epsilon}_{t-1}) \\
& = \cdots \\
& = \sqrt{\alpha_t \alpha_{t-1} \cdots \alpha_{1}} \mathbf{x}_{0} + \sqrt{1- \alpha_t \alpha_{t-1} \cdots \alpha_{1} } \boldsymbol{\epsilon} ,&& \\
& = \sqrt{\Bar{\alpha}_t} \mathbf{x}_{0} + \sqrt{1- \Bar{\alpha}_t } \boldsymbol{\epsilon} ,&& \Rightarrow \text{reparameterize } \Bar{\alpha}_t = \prod_{i=1}^t \alpha_t \\
\end{split}
\end{equation}
therefore,
\begin{equation}
    q(\mathbf{x}_t | \mathbf{x}_0) = \mathcal{N} (\mathbf{x}_t ; \sqrt{\Tilde{\alpha}}_t \mathbf{x}_0 , (1-\Tilde{\alpha_t})\mathbbm{I} ) . 
\end{equation}

Then,
\begin{equation}\label{eq:diffusion_qder}
\begin{split}
q\left(\mathbf{x}_{t-1} | \mathbf{x}_{t}, \mathbf{x}_0 \right) 
&= q\left(\mathbf{x}_{t} | \mathbf{x}_{t-1}, \mathbf{x}_0 \right) \frac{q\left(\mathbf{x}_{t-1} |  \mathbf{x}_0 \right)}{q\left(\mathbf{x}_{t} |  \mathbf{x}_0 \right)} , \\
& \propto \exp\left( - \frac{1}{2} \left( \frac{(\mathbf{x}_t - \sqrt{\alpha_t} \mathbf{x}_{t-1})^2 }{\beta_t} + \frac{(\mathbf{x}_{t-1} - \sqrt{\Bar{\alpha}_{t-1}} \mathbf{x}_{0})^2 }{1-\Bar{\alpha}_{t-1}} - \frac{(\mathbf{x}_t - \sqrt{\Bar{\alpha}_t} \mathbf{x}_{0})^2 }{1-\Bar{\alpha}_{t}} \right) \right) , \\
& =  \exp\left( - \frac{1}{2} \left( 
\left(\frac{\alpha_t}{\beta_t} + \frac{1}{1-\Bar{\alpha}_{t-1}}\right) \mathbf{x}_{t-1}^2 - \left( \frac{2\sqrt{\alpha_t}\mathbf{x}_{t}  }{\beta_t} + \frac{2\sqrt{\Bar{\alpha}_{t-1} \mathbf{x}_0 }}{1-\Bar{\alpha}_{t-1}}  \right) \mathbf{x}_{t-1} + \cdots
\right) \right) ,
\\
\end{split}
\end{equation}
where $\cdots$ include the terms irrelevant to $\mathbf{x}_{t-1}$. Since $q\left(\mathbf{x}_{t-1} | \mathbf{x}_{t}, \mathbf{x}_0 \right) $ is a Gaussian distribution, we can find the mean and the variance from Equation~\eqref{eq:diffusion_qder}:
\begin{equation}
\begin{split}
& \Tilde{\beta}_t = \frac{1}{\frac{\alpha_t}{\beta_t} + \frac{1}{1-\Bar{\alpha}_{t-1}}} = \frac{1-\Bar{\alpha}_{t-1}}{ 1-\Bar{\alpha}_{t}} \cdot \beta_t , \\ 
\end{split}
\end{equation}
and
\begin{equation}
\begin{split}
 \Tilde{\boldsymbol{\mu}}_t  
&= \left( \frac{\sqrt{\alpha_t}\mathbf{x}_{t}  }{\beta_t} + \frac{\sqrt{\Bar{\alpha}_{t-1}  } \mathbf{x}_0 }{1-\Bar{\alpha}_{t-1}} \right) / \left( \frac{\alpha_t}{\beta_t} + \frac{1}{1-\Bar{\alpha}_{t-1}} \right) , \\ 
& = \left( \frac{\sqrt{\alpha_t}\mathbf{x}_{t}  }{\beta_t} + \frac{\sqrt{\Bar{\alpha}_{t-1}  } \mathbf{x}_0}{1-\Bar{\alpha}_{t-1}} \right) \cdot \left( \frac{1-\Bar{\alpha}_{t-1}}{ 1-\Bar{\alpha}_{t}} \cdot \beta_t \right) , \\
& =  \frac{(1-\Bar{\alpha}_{t-1}) \sqrt{\alpha_t} }{ 1-\Bar{\alpha}_{t}} \mathbf{x}_t + \frac{\sqrt{\Bar{\alpha}_{t-1}  } \beta_t }{1-\Bar{\alpha}_{t}}   \mathbf{x}_0 , \\
& =  \frac{(1-\Bar{\alpha}_{t-1}) \sqrt{\alpha_t} }{ 1-\Bar{\alpha}_{t}} \mathbf{x}_t + \frac{\sqrt{\Bar{\alpha}_{t-1}  } \beta_t }{1-\Bar{\alpha}_{t}}  \left( \frac{1}{\sqrt{\Bar{\alpha}_t}} (\mathbf{x}_t - \sqrt{1-\Bar{\alpha}_t} \boldsymbol{\epsilon}_t ) \right) , \\
& = \frac{1}{\sqrt{\alpha_t}} \left( \mathbf{x}_t - \frac{1-\alpha_t}{\sqrt{1-\Bar{\alpha}_t} } \boldsymbol{\epsilon}_t  \right) .\\
\end{split}
\end{equation}

Therefore, we can rewrite the $L_{t-1}$ term in Equation~\eqref{eq:diffusion_loss} using the KL-divergence between two Gaussian distributions as:
\begin{equation}
L_{t-1} = D_{KL} \left(  q\left(\mathbf{x}_{t-1} | \mathbf{x}_{t} , \mathbf{x}_0 \right)  || p_\theta \left( \mathbf{x}_{t-1} | \mathbf{x}_t \right) \right) = \mathbbm{E}_q \left[ \frac{1}{2\sigma_t^2} \| \Tilde{\boldsymbol{\mu}}_t (\mathbf{x}_t , \mathbf{x}_0 ) - \boldsymbol{\mu}_\theta (\mathbf{x}_t , t) \|^2 \right] + C . 
\end{equation}

Essentially, $\mu_{\theta} (\mathbf{x}_t, t )$ is a $\theta$-parameterized function that predicts the distribution mean given $\mathbf{x}_t$ and $t$. We can reparameterize it as:
\begin{equation}
\begin{split}
\mu_\theta (\mathbf{x}_t , t) = \frac{1}{\sqrt{\alpha_t}} \left( \mathbf{x}_t - \frac{1-\alpha_t}{\sqrt{1-\Bar{\alpha}_t} } \boldsymbol{\epsilon}_{\theta} (\mathbf{x}_t, t)  \right) , 
\end{split}
\end{equation}
which then 
\begin{equation}
\begin{split}
L_{t-1} & = \mathbbm{E}_q \left[ \frac{1}{2\sigma_t^2} \norm{\Tilde{\boldsymbol{\mu}}_t (\mathbf{x}_t , \mathbf{x}_0 ) - \boldsymbol{\mu}_\theta (\mathbf{x}_t , t) }^2 \right] + C ,\\
& = \mathbbm{E}_q \left[ \frac{(1-\alpha_t)^2}{2 \alpha_t (1-\alpha_t) \sigma_t^2} \norm{ \boldsymbol{\epsilon}_t - \boldsymbol{\epsilon}_{\theta} (\mathbf{x}_t , t)}^2 \right] + C , \\
& = \mathbbm{E}_{t, \mathbf{x}_0 , \boldsymbol{\epsilon}} \left[ \frac{(1-\alpha_t)^2}{2 \alpha_t (1-\alpha_t) \sigma_t^2} \norm{\boldsymbol{\epsilon} - \boldsymbol{\epsilon}_{\theta} \left(\sqrt{\Bar{\alpha}}_t \mathbf{x}_0 + \sqrt{1-\Bar{\alpha}_t }\boldsymbol{\epsilon} , t \right)}^2 \right] + C .\\
\end{split}
\end{equation}

\section{Derivation for Score-based Generative Model}\label{appendix:score_langevin}
Once the score network is trained, data generation can be accomplished through an iterative method known as \textit{Langevin dynamics}, or Langevin Monte Carlo \citep{parisi1981correlation, grenander1994representations}. Langevin dynamics is originally formulated to describe the behavior of molecular systems, such as Brownian motion. The fundamental principle of Langevin dynamics in the context of Brownian motion is captured by the Langevin Equation, which is expressed as follows:
\begin{equation}
    \label{eq:brownian}
    \begin{split}
        m \ddot{x} = -\lambda \dot{x} + \eta
    \end{split},
\end{equation}
where $m$ is the mass of the particle, $\lambda$ is the damping coefficient, and $\eta \sim \mathcal{N}(0, 2 \sigma^2)$ is a Gaussian noise. 

The dynamics of Brownian motion as described in Equation~\eqref{eq:brownian} can be further derived by considering the potential energy in the system. Specifically, the force acting on a particle can be expressed in terms of the gradient of potential energy; i.e., $\frac{\partial V(x)}{\partial x} = \nabla V(x)$. This leads to a reformulation of the equation as:
\begin{equation}
\nabla V(x) = -\lambda \dot{x} + \eta,
\end{equation}
or equivalently,
\begin{equation}
\lambda \dot{x} = - \nabla V(x) + \eta.
\end{equation}
This relationship allows us to interpret the motion of the particle in terms of potential energy changes, where $-\nabla V(x)$ represents the deterministic force derived from the potential energy landscape and $\eta$ accounts for stochastic thermal fluctuations.
To capture these dynamics in a more mathematically rigorous framework suitable for simulation and analysis, we can transform this physical concept into a stochastic differential equation (SDE):
\begin{equation}\label{eq:brown_sde}
dx= - \nabla V(x) dt + \sqrt{2}\sigma dW,
\end{equation}
where the term $- \nabla V(x)$ is the drift term, which guides the particle towards lower potential energy states, thus simulating deterministic motion under the influence of forces. The term $\sqrt{2}\sigma dW$ represents the stochastic component of the motion, with $\sigma$ being the volatility (akin to the temperature in physical systems) and $dW$ denoting the derivative of a standard Wiener process, which models the random thermal fluctuations.

In the context of probability distributions, the stochastic process described in Equation~\eqref{eq:brown_sde} corresponds to a Fokker-Planck equation \citep{fokker1914mittlere,planck1917satz,kadanoff2000statistical} for the probability density function $p(x,t)$ of finding the system in state $x$ at time $t$:
\begin{equation}
    \frac{\partial}{ \partial t} p(x,t) = \nabla \cdot \left[ \nabla V(x) p(x,t) + \sigma^2 \nabla p(x,t) \right]  . 
\end{equation}

At steady state, the time derivative of the probability density function becomes zero, implying:
\begin{equation}
    0 = \nabla \cdot \left[\nabla V(x) p(x,t) + \sigma^2 \nabla p(x,t) \right] ,
\end{equation}
which simplifies to:
\begin{equation}
\nabla V(x) p(x) = - \sigma^2 \nabla p(x,t) . 
\end{equation}

Dividing both sides by $p(x)$ and rearranging gives:
\begin{equation}
\nabla V(x) \propto - \frac{\nabla p(x,t)}{p(x)} = - \nabla \log p(x) , 
\end{equation}
indicating that, at equilibrium, the gradient of the potential energy is proportional to the gradient of the log of the probability distribution. 
\textbf{This equilibrium condition underlies the principle that in a score-based generative model, the score function }$\nabla_{\mathbf{x}} \log p(\mathbf{x})$ \textbf{can be interpreted as akin to a force derived from a potential energy landscape,} guiding the generation process towards high-probability regions of the data distribution.

Therefore, Equation~\eqref{eq:brown_sde} can be re-written as:
\begin{equation}\label{eq:brown_sde2}
dx= \nabla \log p(x) dt + \sqrt{2}\sigma dW,
\end{equation}

To implement this in a discrete setting for numerical simulation or data generation, the SDE can be approximated as follows:
\begin{equation}
\mathbf{x}_{i+1} \leftarrow \mathbf{x}_{i} +  \nabla_{x} \log p(\mathbf{x}) \epsilon + \sqrt{2} \sigma \Delta W ,
\end{equation}
where $\epsilon$ is a small timestep, and $\Delta W \sim \mathcal{N} (0, \epsilon)$ represents a discrete approximation of the Wiener process increment. Thus, it can be further simplified as:
\begin{equation}\label{eq:score_wiener}
\mathbf{x}_{i+1} \leftarrow \mathbf{x}_{i} +  \nabla_{x} \log p(\mathbf{x}) \cdot \epsilon + \sqrt{2\epsilon} \cdot \mathbf{z}_i , 
\end{equation}
where $\mathbf{z}_i \sim \mathcal{N}(0,I)$.

\section{Tutorials for Generating Household Travel Survey Data}\label{appendix:tutorials_hts}
\subsection{Variational Autoencoder}\label{sec:E.1}

The VAE loss function consists of two primary components as shown in Equation~\eqref{eq:vae_elbo}: the \textit{reconstruction loss} and the \textit{regularization loss}. The reconstruction loss measures how well the model can reproduce the original data. Since the encoder and decoder are both deterministic functions when input is given, the reconstruction loss term, $\mathbb{E}_{\mathbf{z}\sim q_{\phi} (\mathbf{z} | \mathbf{x})} \left[ \log p_\theta (\mathbf{x} | \mathbf{z}) \right]$, can be re-written as $\mathbb{E}_{\mathbf{x}} \left[ \log p_\theta (\mathbf{\hat{x}} | \mathbf{x}) \right]$. If we assume that this distribution has a Gaussian form:
\begin{equation}
     \log p_\theta (\mathbf{\hat{\mathbf{x}}} | \mathbf{\mathbf{x}}) \propto \log e^{-|\mathbf{x}-\hat{\mathbf{x}}|^2} = -|\mathbf{x}-\hat{\mathbf{x}}|^2 ,
\end{equation}
then maximizing $\mathbb{E}_{\mathbf{x}} \left[ \log p_\theta (\mathbf{\hat{x}} | \mathbf{x}) \right]$ corresponds to minimizing the square loss between $\mathbf{x}$ and $\mathbf{\hat{x}}$.
The regularization loss, on the other hand, ensures that the induced latent space from the encoder (i.e., the posterior distribution) follows the same distribution as the prior distribution. However, in practical implementations, the regularization term often requires simplifying assumptions. In this case, we assume that both the posterior distribution $q_{\phi}(\mathbf{z}|\mathbf{x})$ and the prior distribution $p(\mathbf{z})$ are Gaussian distribution. Under these assumptions, the Kullback-Leibler (KL) divergence between the two Gaussian distributions, considering D-dimensional vector $\mathbf{z}$ can be simplified into a more computationally manageable form as:
\begin{equation}
\label{eqn:vae_der_1}
    \begin{split}
    D_{KL} \left( q_{\phi} \left(\mathbf{z}|\mathbf{x} \right) || p (\mathbf{z})  \right) &= D_{KL} \left( \mathcal{N}(\boldsymbol{\mu}_1, \boldsymbol{\Sigma}_1) || \mathcal{N}(\boldsymbol{\mu}_2, \boldsymbol{\Sigma}_2) \right) \\
    &= \frac{1}{2} \left( \mathrm{Tr}(\boldsymbol{\Sigma}_2^{-1} \boldsymbol{\Sigma}_1) + (\boldsymbol{\mu}_2 - \boldsymbol{\mu}_1)^\top \boldsymbol{\Sigma}_2^{-1} (\boldsymbol{\mu}_2 - \boldsymbol{\mu}_1) - D + \log \frac{\det \boldsymbol{\Sigma}_2}{\det \boldsymbol{\Sigma}_1} \right).\\
    \end{split}
\end{equation}
where $(\boldsymbol{\mu}_1, \mathbf{\Sigma}_1)$, and $(\boldsymbol{\mu}_2, \mathbf{\Sigma}_2 )$ are the mean vectors and covariance matrices for $q_{\phi} (\mathbf{z}|\mathbf{x})$ and $p(\mathbf{z})$, respectively. We can further simplify the equations by setting the prior distribution, $p(\mathbf{x})$, to follow standard Gaussian distribution, i.e., $\boldsymbol{\mu}_2=0$, $\boldsymbol{\Sigma}_2 = \mathbf{I}$, and the posterior distribution, $q_\phi (\mathbf{z}|\mathbf{x})$, follows zero-mean Gaussian distribution with diagonal covariance matrix, i.e., $\boldsymbol{\mu}_1=0$, and $\boldsymbol{\Sigma}_1 = \boldsymbol{\sigma}_1^2 \mathbf{I}$:
\begin{equation}
\label{eqn:vae_der_4}
    \begin{split}
    D_{KL} \left( q_{\phi} \left(\mathbf{z}|\mathbf{x} \right) || p (\mathbf{z})  \right) &= \frac{1}{2} \left( \mathrm{Tr}(\boldsymbol{\Sigma}_1) + \boldsymbol{\mu}_1^\top \boldsymbol{\mu}_1 - D - \log \det \boldsymbol{\Sigma}_1 \right)\\
    & = \frac{1}{2} \sum_{i=1}^{D} \left( \sigma_{1,i}^2 + \mu_{1,i}^2 - 1 - \log \sigma_{1,i}^2 \right),\\
    \end{split}
\end{equation}
where $\mu_{1,i}$ is $i$-th element in $\boldsymbol{\mu}_1$ and $\sigma_{1,i}$ is $i$-th element in $\boldsymbol{\sigma}_{1}$. The corresponding code implementation of the VAE loss is stated below. 
We use $\log \boldsymbol{\sigma}_1$ instead of $\boldsymbol{\sigma}_1$ for numerical stability, and the variable \texttt{logvar} corresponds to $2\log \boldsymbol{\sigma}_1$. The variable \texttt{mu} corresponds to $\boldsymbol{\mu}_1$. Figure~\ref{fig:VAE_nn_disc} illustrates the structure of the neural network for this section. Figure~\ref{fig:VAE_nn_disc} (a) shows the structure of the encoder which encodes the inputs to $D$-dimensional vectors for both Mean and Variance and we used $D=64$, and Figure~\ref{fig:VAE_nn_disc} (b) shows the structure of the decoder which generates the sample from the latent vector. 
The Pytorch code implementation is as follows:

\begin{figure}[t]
    \centering
    \begin{subfigure}[b]{\textwidth}
        \centering
        \begin{tikzpicture}
        \tikzstyle{block0} = [rectangle, text centered, minimum height=1cm, minimum width=1cm]
        \tikzstyle{block} = [rectangle, draw, fill=gray!20, text centered, minimum height=1cm, minimum width=2.5cm, rotate=90]
        \tikzstyle{relu} = [rectangle, draw, fill=orange!30, text centered, minimum height=1cm, minimum width=2.5cm, rotate=90]
        \tikzstyle{arrow} = [thick,->,>=stealth]

        \node[block0] (input) at (0, 0) {\shortstack{Input \\ (\tt{x})}};
        \node[block] (linear1) at (2, 0) {\shortstack{Linear \\ (4 $\to$ 512)}};
        \node[relu] (relu1) at (3, 0) {ReLU};
        \node[block] (linear2) at (5, 0) {\shortstack{Linear \\ (512 $\to$ 512)}};
        \node[relu] (relu2) at (6, 0) {ReLU};
        \node[block] (linear3) at (8, 2) {\shortstack{Linear \\ (512 $\to$ 64) \\(\tt{mu\_layer}) }};
        \node[block] (linear4) at (8, -2) {\shortstack{Linear \\ (512 $\to$ 64)  \\(\tt{log\_var\_layer}) }};
        
        \node[block0] (Mean) at (10, 2){\shortstack{Mean\\ (\tt{mu})}};
        \node[block0] (Variance) at (10, -2){\shortstack{Variance \\ (\tt{logvar})}};

        \draw[arrow] (input.east) -- (linear1.north);
        \draw[arrow] (relu1.south) -- (linear2.north);
        \draw[arrow] (relu2.south) ++(0,0) -- ++(0.5,0) |- (linear3.north);
        \draw[arrow] (relu2.south) ++(0,0) -- ++(0.5,0) |- (linear4.north);
        \draw[arrow] (linear3.south) -- (Mean.west);
        \draw[arrow] (linear4.south) -- (Variance.west);

        \end{tikzpicture}
        \caption{Network structure of encoder}
        \label{fig:VAE_latent}
    \end{subfigure}
    
    \begin{subfigure}[b]{\textwidth}
        \centering
        \begin{tikzpicture}
        \tikzstyle{block0} = [rectangle, text centered, minimum height=1cm, minimum width=1cm]
        \tikzstyle{block1} = [rectangle,  minimum height=1cm, minimum width=1cm]        \tikzstyle{block} = [rectangle, draw, fill=gray!20, text centered, minimum height=1cm, minimum width=2.5cm, rotate=90]
        \tikzstyle{relu} = [rectangle, draw, fill=orange!30, text centered, minimum height=1cm, minimum width=2.5cm, rotate=90]
        \tikzstyle{arrow} = [thick,->,>=stealth]

        \node[block0] (input) at (0, 0) {\shortstack{Latent\\Vector\\(\tt{z})}};
        \node[block] (linear1) at (2, 0) {\shortstack{Linear \\ (64 $\to$ 512)}};
        \node[relu] (relu1) at (3, 0) {ReLU};
        \node[block] (linear2) at (5, 0) {\shortstack{Linear \\ (512 $\to$ 512)}};
        \node[relu] (relu2) at (6, 0) {ReLU};
        \node[block] (linear3) at (8, 0) {\shortstack{Linear \\ (512 $\to$ 4)}};
        \node[block1] (output) at (11, 0) {\shortstack{Sample \\ (\tt{x\_reconstructed})}};

        \draw[arrow] (input.east) -- (linear1.north);
        \draw[arrow] (relu1.south) -- (linear2.north);
        \draw[arrow] (relu2.south) -- (linear3.north);
        \draw[arrow] (linear3.south) -- (output.west);
        \end{tikzpicture}
        \caption{Network structure of decoder}
        \label{fig:VAE_out}
    \end{subfigure}
    
    \caption{Neural network structures of VAE in HTS data}
    \label{fig:VAE_nn_disc}
\end{figure}


\begin{lstlisting}[language=Python]
    class VAE(nn.Module):
        def __init__(self, input_dim, hidden_dim, z_dim):
            super(VAE, self).__init__()
            # Encoder
            self.encoder = nn.Sequential(
                nn.Linear(input_dim, hidden_dim),
                nn.ReLU(),
                nn.Linear(hidden_dim, hidden_dim),
                nn.ReLU()
            )
            self.mu_layer = nn.Linear(hidden_dim, z_dim)
            self.log_var_layer = nn.Linear(hidden_dim, z_dim)
    
            # Decoder
            self.decoder = nn.Sequential(
                nn.Linear(z_dim, hidden_dim),
                nn.ReLU(),
                nn.Linear(hidden_dim, hidden_dim),
                nn.ReLU(),
                nn.Linear(hidden_dim, input_dim)
            )
            
        def encode(self, x):
            h = self.encoder(x)
            mu = self.mu_layer(h)
            log_var = self.log_var_layer(h)
            return mu, log_var
    
        def reparameterize(self, mu, log_var):
            std = torch.exp(0.5 * log_var)
            eps = torch.randn_like(std)
            return mu + eps * std
    
        def decode(self, z):
            return self.decoder(z)
    
        def forward(self, x):
            mu, log_var = self.encode(x)
            z = self.reparameterize(mu, log_var)
            x_reconstructed = self.decode(z)
            return x_reconstructed, mu, log_var
    
    def compute_loss(x, x_reconstructed, mu, log_var):
        """
        MSE based reconstruction loss and KL divergence loss for VAE.
        """
        # Reconstruction loss
        recon_loss = nn.functional.mse_loss(x_reconstructed, x, reduction='sum')
        # KL Divergence
        kl_loss = -0.5 * torch.sum(1 + log_var - mu.pow(2) - log_var.exp())
        # total loss
        loss = recon_loss + kl_loss
        return loss
    
\end{lstlisting}

\subsection{Generative Adversarial Networks}\label{sec:E.2}

Unlike VAEs, GANs do not require complex derivations of the loss function. The code implementation of the Equations~\eqref{eqn:gan_1} and \eqref{eqn:gan_2} can be written intuitively. In practice, this involves using binary cross-entropy loss. The discriminator tries to identify if the given input is from the real data set or the generated data set. On the contrary, the generator tries to overcome the identification from the discriminator. 
Similar to the VAE, we constructed the 3-layer neural network for (a) generator and (b) discriminator as shown in Figure~\ref{fig:gan_nn_disc}. The noise vector (latent vector) is $D$-dimensional vector where $D=64$.
The Pytorch code implementation is as follows:
\begin{lstlisting}[language=Python]
    class Generator(nn.Module):
        def __init__(self, input_dim, output_dim, hidden_dim):
            super(Generator, self).__init__()
            self.net = nn.Sequential(
                nn.Linear(input_dim, hidden_dim),
                nn.ReLU(),
                nn.Linear(hidden_dim, hidden_dim),
                nn.ReLU(),
                nn.Linear(hidden_dim, output_dim)
            )
        
        def forward(self, z):
            return self.net(z)
    
    class Discriminator(nn.Module):
        def __init__(self, input_dim, hidden_dim):
            super(Discriminator, self).__init__()
            self.net = nn.Sequential(
                nn.Linear(input_dim, hidden_dim),
                nn.ReLU(),
                nn.Linear(hidden_dim, hidden_dim),
                nn.ReLU(),
                nn.Linear(hidden_dim, 1),
                nn.Sigmoid()
            )
        
        def forward(self, x):
            return self.net(x)

    def compute_discriminator_loss(discriminator, real_data, fake_data, criterion, real_label, fake_label):
        batch_size = real_data.size(0)
        # loss for real data
        real_targets = torch.full((batch_size, 1), real_label, device=real_data.device)
        d_real = discriminator(real_data)
        d_loss_real = criterion(d_real, real_targets)
    
        # loss for generated data
        fake_targets = torch.full((batch_size, 1), fake_label, device=real_data.device)
        d_fake = discriminator(fake_data)
        d_loss_fake = criterion(d_fake, fake_targets)
    
        d_loss = d_loss_real + d_loss_fake
        return d_loss
    
    def compute_generator_loss(discriminator, fake_data, criterion, real_label):
        batch_size = fake_data.size(0)
        real_targets = torch.full((batch_size, 1), real_label, device=fake_data.device)
        d_fake = discriminator(fake_data)
        g_loss = criterion(d_fake, real_targets)
        return g_loss

\end{lstlisting}


\begin{figure}[t]
    \centering
    
    \begin{subfigure}[b]{\textwidth}
        \centering
        \begin{tikzpicture}
        \tikzstyle{block0} = [rectangle, text centered, minimum height=1cm, minimum width=1cm]
        \tikzstyle{block} = [rectangle, draw, fill=gray!20, text centered, minimum height=1cm, minimum width=2.5cm, rotate=90]
        \tikzstyle{relu} = [rectangle, draw, fill=orange!30, text centered, minimum height=1cm, minimum width=2.5cm, rotate=90]
        \tikzstyle{arrow} = [thick,->,>=stealth]

        \node[block0] (input) at (0, 0) {\shortstack{Inputs\\(\tt{real\_data}\\\tt{fake\_data}) }};
        \node[block] (linear1) at (2, 0) {\shortstack{Linear \\ (4 $\to$ 512)}};
        \node[relu] (relu1) at (3, 0) {ReLU};
        \node[block] (linear2) at (5, 0) {\shortstack{Linear \\ (512 $\to$ 512)}};
        \node[relu] (relu2) at (6, 0) {ReLU};
        \node[block] (linear3) at (8, 0) {\shortstack{Linear \\ (512 $\to$ 4)}};
        \node[relu] (Sigmoid) at (9, 0) {Sigmoid};
        \node[block0] (output) at (12.5, 0){\shortstack{Decision \\ (\tt{d\_real}, \\ \tt{d\_fake}, \\\tt{g\_fake})}};

        \draw[arrow] (input.east) -- (linear1.north);
        \draw[arrow] (relu1.south) -- (linear2.north);
        \draw[arrow] (relu2.south) -- (linear3.north);
        \draw[arrow] (Sigmoid.south) -- (output.west);
        \end{tikzpicture}
        \caption{Network structure of discriminator}
        \label{fig:GAN_D}
    \end{subfigure}
    
    \begin{subfigure}[b]{\textwidth}
        \centering
        \begin{tikzpicture}
        \tikzstyle{block0} = [rectangle, text centered, minimum height=1cm, minimum width=1cm]
        \tikzstyle{block} = [rectangle, draw, fill=gray!20, text centered, minimum height=1cm, minimum width=2.5cm, rotate=90]
        \tikzstyle{relu} = [rectangle, draw, fill=orange!30, text centered, minimum height=1cm, minimum width=2.5cm, rotate=90]
        \tikzstyle{arrow} = [thick,->,>=stealth]

        \node[block0] (input) at (0, 0) {\shortstack{Latent \\ Vector\\(\tt{z})}};
        \node[block] (linear1) at (2, 0) {\shortstack{Linear \\ (64 $\to$ 512)}};
        \node[relu] (relu1) at (3, 0) {ReLU};
        \node[block] (linear2) at (5, 0) {\shortstack{Linear \\ (512 $\to$ 512)}};
        \node[relu] (relu2) at (6, 0) {ReLU};
        \node[block] (linear3) at (8, 0) {\shortstack{Linear \\ (512 $\to$ 4)}};
        \node[block0] (output) at (10, 0) {\shortstack{Sample\\ (\tt{fake\_data})}};

        \draw[arrow] (input.east) -- (linear1.north);
        \draw[arrow] (relu1.south) -- (linear2.north);
        \draw[arrow] (relu2.south) -- (linear3.north);
        \draw[arrow] (linear3.south) -- (output.west);
        \end{tikzpicture}
        \caption{Network structure of generator}
        \label{fig:GAN_G}
    \end{subfigure}
    
    \caption{Neural network structures of GAN in HTS data}
    \label{fig:gan_nn_disc}
\end{figure}

\subsection{Normalizing Flows (Flow-based Generative Models)}\label{sec:E.3}
We use a flow-based generative model inspired by RealNVP \citep{dinh2016density}, which leverages affine coupling layers and alternating masking strategies to transform complex data distributions into a standard normal distribution. For simplicity, and since the data dimensionality is not too large, the model in this tutorial does not include permutation layers between coupling layers and batch normalization that were proposed in the original RealNVP paper. A more detailed implementation of RealNVP will be discussed in Section~\ref{sec:E.3}. Despite these simplifications, our model maintains the core idea of using invertible affine transformations to compute the log-likelihood and optimize the data distribution. The training objective is to maximize the likelihood of the observed data under this transformation, which can be expressed as minimizing the negative log-likelihood. The original loss function for the normalizing flow is the same as stated in Equation~\eqref{eqn:nf_2}. For the affine coupling layers, we use the same equations as described from Equation~\eqref{eqn:nf_4} to Equation~\eqref{eqn:nf_7}. We use much simpler networks for scaling and translation network as shown in Figure~\ref{fig:NF_nn_disc}.
The code Pytorch implementation is as follows:


\begin{figure}[t]
    \centering
    
    \begin{subfigure}[b]{\textwidth}
        \centering
        \begin{tikzpicture}
        \tikzstyle{block0} = [rectangle, text centered, minimum height=1cm, minimum width=1cm]
        \tikzstyle{block} = [rectangle, draw, fill=gray!20, text centered, minimum height=1cm, minimum width=2.5cm, rotate=90]
        \tikzstyle{relu} = [rectangle, draw, fill=orange!30, text centered, minimum height=1cm, minimum width=2.5cm, rotate=90]
        \tikzstyle{arrow} = [thick,->,>=stealth]

        \node[block0] (input) at (0, 0) {\shortstack{Inputs \\(\texttt{x\_masked})}};
        \node[block] (linear1) at (2, 0) {\shortstack{Linear \\ (4 $\to$ 128)}};
        \node[relu] (relu1) at (3, 0) {ReLU};
        \node[block] (linear2) at (5, 0) {\shortstack{Linear \\ (128 $\to$ 128)}};
        \node[relu] (relu2) at (6, 0) {ReLU};
        \node[block] (linear3) at (8, 0) {\shortstack{Linear \\ (128 $\to$ 4)}};
        \node[relu] (Tanh) at (9, 0) {Tanh};
        \node[block0] (output) at (11, 0){\shortstack{Scale\\ (\texttt{s})}};

        \draw[arrow] (input.east) -- (linear1.north);
        \draw[arrow] (relu1.south) -- (linear2.north);
        \draw[arrow] (relu2.south) -- (linear3.north);
        \draw[arrow] (Tanh.south) -- (output.west);
        \end{tikzpicture}
        \caption{Network structure of scale net}
        \label{fig:NF_S}
    \end{subfigure}
    
    \begin{subfigure}[b]{\textwidth}
        \centering
        \begin{tikzpicture}
        \tikzstyle{block0} = [rectangle, text centered, minimum height=1cm, minimum width=1cm]
        \tikzstyle{block} = [rectangle, draw, fill=gray!20, text centered, minimum height=1cm, minimum width=2.5cm, rotate=90]
        \tikzstyle{relu} = [rectangle, draw, fill=orange!30, text centered, minimum height=1cm, minimum width=2.5cm, rotate=90]
        \tikzstyle{arrow} = [thick,->,>=stealth]

        \node[block0] (input) at (0, 0) {\shortstack{Inputs \\(\texttt{x\_masked})}};
        \node[block] (linear1) at (2, 0) {\shortstack{Linear \\ (4 $\to$ 128)}};
        \node[relu] (relu1) at (3, 0) {ReLU};
        \node[block] (linear2) at (5, 0) {\shortstack{Linear \\ (128 $\to$ 128)}};
        \node[relu] (relu2) at (6, 0) {ReLU};
        \node[block] (linear3) at (8, 0) {\shortstack{Linear \\ (128 $\to$ 4)}};
        \node[block0] (output) at (10.5, 0) {\shortstack{Translation\\ (\texttt{t})}};

        \draw[arrow] (input.east) -- (linear1.north);
        \draw[arrow] (relu1.south) -- (linear2.north);
        \draw[arrow] (relu2.south) -- (linear3.north);
        \draw[arrow] (linear3.south) -- (output.west);
        \end{tikzpicture}
        \caption{Network structure of translation net}
        \label{fig:NF_T}
    \end{subfigure}
    
    \caption{Neural network structures of normalizing flow in HTS data}
    \label{fig:NF_nn_disc}
\end{figure}

\begin{lstlisting}[language=Python]
    class AffineCoupling(nn.Module):
        [...]
        # provide calculation of log_det_jacobian using scale and translation functions
        def forward(self, x):
            x_masked = x * self.mask
            s = self.scale_net(x_masked) * (1 - self.mask)
            t = self.translate_net(x_masked) * (1 - self.mask)
            y = x_masked + (1 - self.mask) * (x * torch.exp(s) + t)
            log_det_jacobian = torch.sum(s, dim=1)
            return y, log_det_jacobian 
            [...]
        
    class NormalizingFlow(nn.Module):
        def __init__(self, input_dim, hidden_dim, num_layers):
            super(NormalizingFlow, self).__init__()
            self.layers = nn.ModuleList()
            mask = self.create_mask(input_dim, even=True)
            for i in range(num_layers):
                self.layers.append(AffineCoupling(input_dim, hidden_dim, mask))
                # Alternate the mask for the next layer
                mask = 1 - mask
        
        def forward(self, x):
            """
            Forward pass through the flow: transform x to z and compute log_prob.
            """
            log_det_jacobian = torch.zeros(x.size(0))
            for layer in self.layers:
                x, ldj = layer(x)
                log_det_jacobian += ldj
            # Compute log_prob under base distribution
            log_prob_z = self.base_log_prob(x)
            log_prob = log_prob_z + log_det_jacobian
            return log_prob
            
    def compute_nf_loss(model, x):
        log_prob = model(x)
        loss = -torch.mean(log_prob)
        return loss

\end{lstlisting}

\subsection{Score-based Generative Model}\label{sec:E.5}

The original objective function of the Score-based Model is defined as Equation~\eqref{eqn:score_1}. Here we show how the transformation to Equation~\eqref{eqn:score_2} proceeds when we are using the normal distribution for the noise $q_\sigma \left( \tilde{\mathbf{x}} | \mathbf{x} \right)$. In this case, the equation can be simplified as shown follows:
\begin{equation}
    \label{eqn:ncsn_der1}
    \begin{split}
    &  \left|\left|{ \nabla_{\mathbf{x}} \log q_\sigma \left( \tilde{\mathbf{x}} | \mathbf{x} \right) - s_{\theta} (\tilde{\mathbf{x}} , \sigma)} \right|\right|^2_2 
     = \left|\left|{ \frac{\left(\tilde{\mathbf{x}}-\mathbf{x}\right)}{\sigma^2} + s_{\theta} (\tilde{\mathbf{x}} , \sigma)} \right|\right|^2_2 , \\
    \end{split}
\end{equation}
since 
\begin{equation}
    \label{eqn:ncsn_der1}
    \begin{split}
    &q_\sigma \left( \tilde{\mathbf{x}} | \mathbf{x} \right) = \frac{1}{\sqrt{2 \pi \sigma^2}} \exp \left(-\frac{\left(\tilde{\mathbf{x}}-\mathbf{x} \right)^2}{2 \sigma^2}\right) , \\
    &\log q_\sigma \left( \tilde{\mathbf{x}} | \mathbf{x} \right) = -\frac{\left(\tilde{\mathbf{x}}-\mathbf{x}\right)^2}{2 \sigma^2}-\log \left(\sqrt{2 \pi \sigma^2}\right) , \\
    &\nabla_{\mathbf{x}} \log q_\sigma \left( \tilde{\mathbf{x}} | \mathbf{x} \right) = -\frac{\left(\tilde{\mathbf{x}}-\mathbf{x}\right)}{\sigma^2} .
    \end{split}
\end{equation}
where the term $\tilde{\mathbf{x}} - \mathbf{x}$ corresponds to the variable \texttt{noise}. 
We use a neural network with three linear layers and ReLU activations to obtain the estimated score as shown in Figure~\ref{fig:score_nn_disc}. In this tutorial, we train the NCSN using sequential sigma noise levels rather than randomly mixing them. This approach diverges from the typical method of random sampling of sigma values but was chosen to provide a clearer illustration of the overall training process. Given the relatively low dimension of the dataset, the results show that the sequential method is also feasible. The Pytorch code implementation is as follows:
\begin{lstlisting}[language=Python]
    class ScoreNetwork(nn.Module):
        def __init__(self, input_dim, hidden_dim, sigma_min, sigma_max, num_sigma):
            super(ScoreNetwork, self).__init__()
            [...]
            self.sigmas = torch.exp(torch.linspace(
                np.log(self.sigma_max), np.log(self.sigma_min), self.num_sigma))
    
            # Neural network architecture
            self.net = nn.Sequential(
                nn.Linear(input_dim + 1, hidden_dim),
                nn.ReLU(),
                nn.Linear(hidden_dim, hidden_dim),
                nn.ReLU(),
                nn.Linear(hidden_dim, input_dim)
            )
    
        def forward(self, x, sigma):
            # Concatenate sigma as an additional feature
            sigma_feature = sigma.view(-1, 1)
            x_sigma = torch.cat([x, sigma_feature], dim=1)
            score = self.net(x_sigma)
            return score
    
        def loss_fn(self, x):
            total_loss = 0.0
            for sigma in self.sigmas:
                sigma = sigma.to(device)
                sigma = sigma.expand(x.size(0), 1)
                noise = torch.randn_like(x) * sigma
                x_noisy = x + noise
                score = self.forward(x_noisy, sigma)
                loss = ((score + noise / (sigma ** 2)) ** 2).mean()
                total_loss += loss
            return total_loss

    [...]
    score_model = ScoreNetwork(input_dim, hidden_dim, sigma_min, sigma_max, num_sigma)

    [...] # in training loop
    loss = score_model.loss_fn(x) # Equation (4.6)
\end{lstlisting}


\begin{figure}
\centering
\begin{tikzpicture}
\tikzstyle{block0} = [rectangle, text centered, minimum height=1cm, minimum width=1cm]
\tikzstyle{block} = [rectangle, draw, fill=gray!20, text centered, minimum height=1cm, minimum width=2.5cm, rotate=90]
\tikzstyle{relu} = [rectangle, draw, fill=orange!30, text centered, minimum height=1cm, minimum width=2.5cm, rotate=90]
\tikzstyle{arrow} = [thick,->,>=stealth]

\node[block0] (input) at (-0.165, 1) {\shortstack{Input \\(\texttt{x\_noisy})}};
\node[block0] (sigma) at (0, -1) {\shortstack{Noise Level \\(\texttt{sigma})}};

\node[block] (linear1) at (2, 0) {\shortstack{Linear \\ (4+1 $\to$ 512)}};
\node[relu] (relu1) at (3, 0) {ReLU};
\node[block] (linear2) at (5, 0) {\shortstack{Linear \\ (512 $\to$ 512)}};
\node[relu] (relu2) at (6, 0) {ReLU};
\node[block] (linear3) at (8, 0) {\shortstack{Linear \\ (512 $\to$ 4)}};
\node[block0] (output) at (10, 0) {\shortstack{Score\\(\texttt{score})}};

\draw[arrow] (input.east) ++(0,0) -- ++(0.5,0) |- (linear1.north);
\draw[arrow] (sigma.east) ++(0,0) -- ++(0.29,0) |- (linear1.north);
\draw[arrow] (relu1.south) -- (linear2.north);
\draw[arrow] (relu2.south) -- (linear3.north);
\draw[arrow] (linear3.south) -- (output.west);

\end{tikzpicture}
\caption{Neural network structure of NCSN in HTS data}
\label{fig:score_nn_disc}
\end{figure}

\subsection{Diffusion Models}\label{sec:E.4}

In the tutorial code, we used the Denoising Diffusion Probabilistic Model (DDPM) as the reference for generating data \citep{ho2020denoising}. Instead of using the traditional U-Net architecture for noise prediction, we implemented a simpler neural network structure, as shown in Figure~\ref{fig:diff_nn_disc}. The U-Net architecture is commonly used in diffusion models due to its powerful ability to capture multi-scale features in image data. However, in this tutorial, we implemented a simpler neural network to reduce computational complexity and facilitate a focus on the fundamental concepts of diffusion models, such as stepwise noise estimation and progressive denoising. The loss function for training is calculated based on Equation~\eqref{eqn:diff_1} presented above, which measures the model's ability to predict noise accurately at each step of the diffusion process.
The Pytorch code implementation is as follows:


\begin{figure}[t]
\centering
\begin{tikzpicture}
\tikzstyle{block0} = [rectangle, text centered, minimum height=1cm, minimum width=1cm]
\tikzstyle{block} = [rectangle, draw, fill=gray!20, text centered, minimum height=1cm, minimum width=2.5cm, rotate=90]
\tikzstyle{block} = [rectangle, draw, fill=gray!20, text centered, minimum height=1cm, minimum width=2.5cm, rotate=90]
\tikzstyle{relu} = [rectangle, draw, fill=orange!30, text centered, minimum height=1cm, minimum width=2.5cm, rotate=90]
\tikzstyle{arrow} = [thick,->,>=stealth]

\node[block0] (input) at (0, 1) {\shortstack{Inputs \\(\texttt{x\_t})}};
\node[block0] (time) at (0, -1) {\shortstack{Inputs \\(\texttt{t})}};
\node[block] (linear1) at (2, 0) {\shortstack{Linear \\ (4+128 $\to$ 256)}};
\node[relu] (relu1) at (3, 0) {ReLU};
\node[block] (linear2) at (5, 0) {\shortstack{Linear \\ (256 $\to$ 256)}};
\node[relu] (relu2) at (6, 0) {ReLU};
\node[block] (linear3) at (8, 0) {\shortstack{Linear \\ (256 $\to$ 4)}};
\node[block0] (output) at (10.5, 0) {\shortstack{Noise \\ (\tt{noise\_pred})}};

\draw[arrow] (input.east) ++(0,0) -- ++(0.5,0) |- (linear1.north);
\draw[arrow] (time.east) ++(0,0) -- ++(0.5,0) |- (linear1.north);
\draw[arrow] (relu1.south) -- (linear2.north);
\draw[arrow] (relu2.south) -- (linear3.north);
\draw[arrow] (linear3.south) -- (output.west);

\end{tikzpicture}
\caption{Neural network structure of DDPM in HTS data}
\label{fig:diff_nn_disc}
\end{figure}

\begin{lstlisting}[language=Python]
    class DiffusionModel(nn.Module):
        [...]

        def forward_diffusion_sample(x0, t):
            noise = torch.randn_like(x0)
            sqrt_alpha_prod = sqrt_alphas_cumprod[t].view(-1, 1) 
            sqrt_one_minus_alpha_prod = sqrt_one_minus_alphas_cumprod[t].view(-1, 1) 
            x_t = sqrt_alpha_prod * x0 + sqrt_one_minus_alpha_prod * noise 
            return x_t, noise
        [...]
    
    def compute_diffusion_loss(model, x0, t, sqrt_alphas_cumprod, sqrt_one_minus_alphas_cumprod):
        """
        Do a forward diffusion with the given x0 and timestep t,
        compute the MSE between the noise predicted by the model and the actual noise.
        """
        x_t, noise = forward_diffusion_sample(x0, t, sqrt_alphas_cumprod, sqrt_one_minus_alphas_cumprod)
        noise_pred = model(x_t, t)
        loss = nn.functional.mse_loss(noise_pred, noise)
        return loss

\end{lstlisting}

\section{Tutorials for Generating Highway Traffic Speed Contour}\label{appendix:tutorials_sc}
\subsection{Variational Autoencoder}\label{sec:F.1}

\begin{figure}[t]
    \centering
    \includegraphics[width=0.60\textwidth]{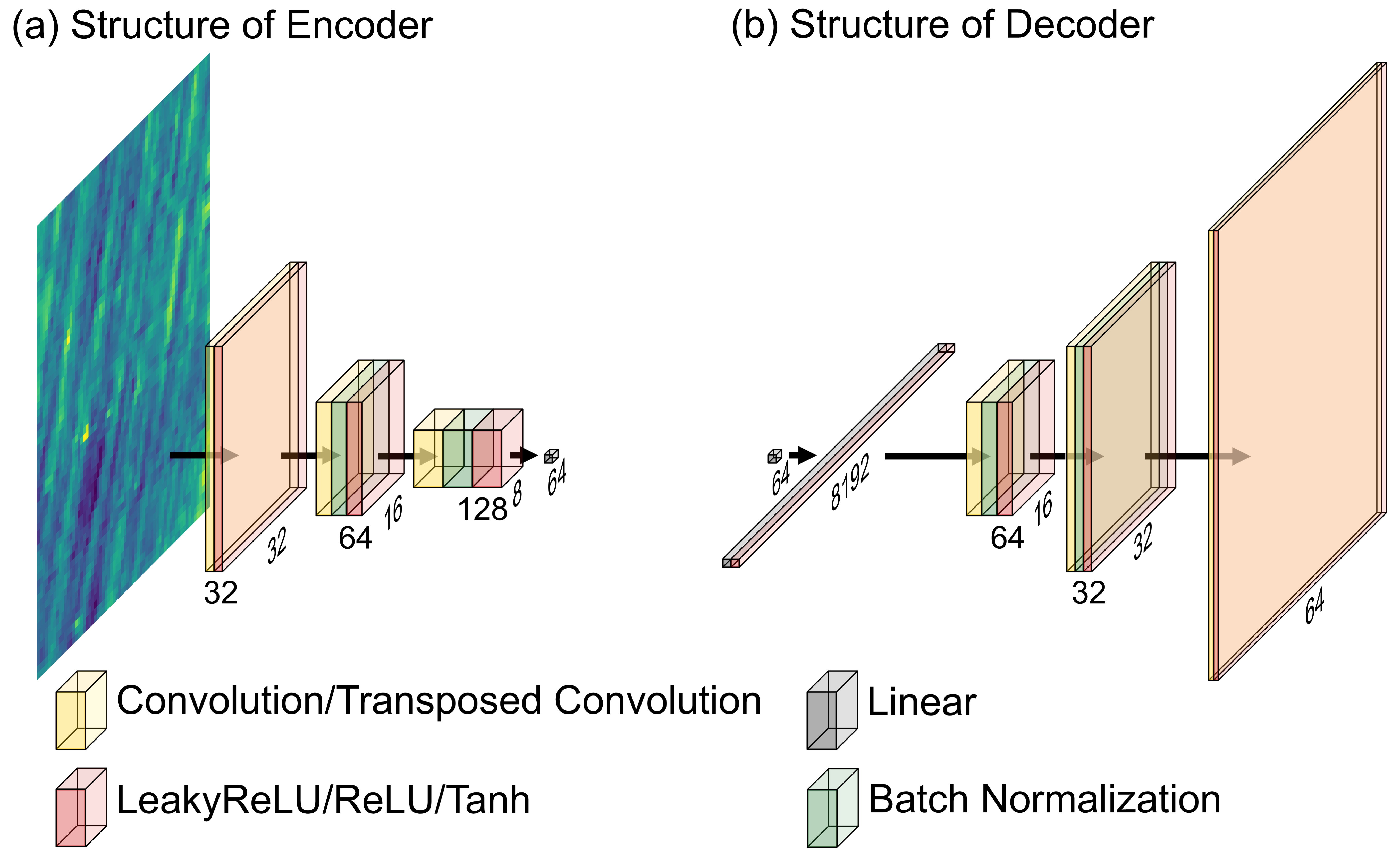}
    \caption{Neural network structure of VAE in highway traffic speed contour}
    \label{fig:vae_nn}
\end{figure}

We utilize the result of Equation~\eqref{eqn:vae_der_4} to compute the loss of the VAE. The code structure and implementation of the loss function are consistent with the approach presented in Section~\ref{sec:E.1}. Figure~\ref{fig:vae_nn} illustrates the structure of the neural network for traffic speed contour generation. Figure~\ref{fig:vae_nn} (a) illustrates the structure of the encoder. The encoder encodes the data to Mean and Variance with a $D$-dimensional vector. In this section, $D=64$ is applied. Figure~\ref{fig:vae_nn} (a) is the structure of the decoder that generates the image from the latent vector.

\subsection{Generative Adversarial Networks}\label{sec:F.2}
The loss function and model architecture of the GAN for traffic speed contour estimation follows the implementation outlined in Section~\ref{sec:E.2}. We employ binary cross-entropy loss for both the discriminator and the generator.

\begin{figure}[t]
    \centering
    \includegraphics[width=0.60\textwidth]{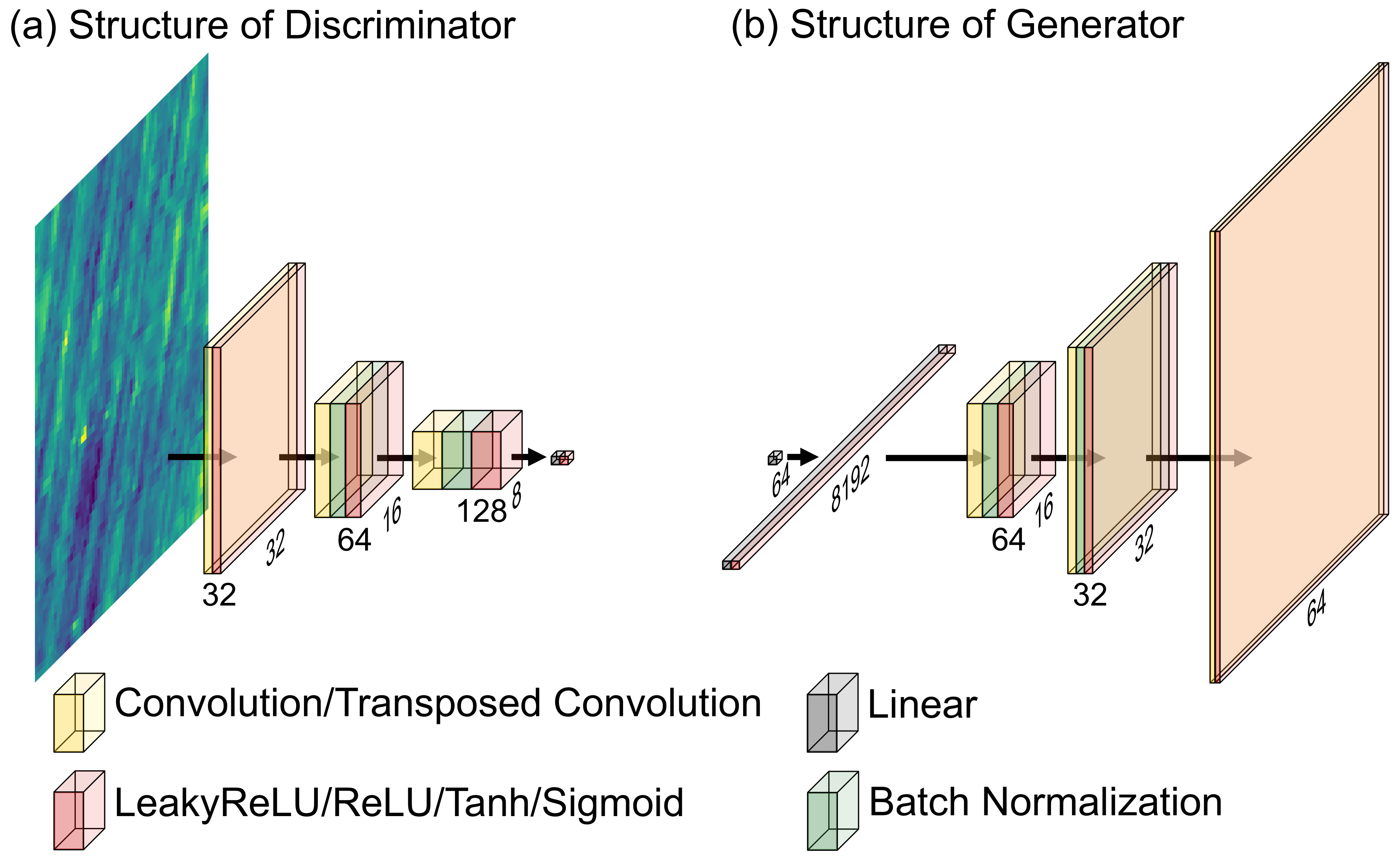}
    \caption{Neural network structure of GAN in highway traffic speed contour}
    \label{fig:gan_nn}
\end{figure}

Figure~\ref{fig:gan_nn} presents the detailed neural network architecture of the GAN, which closely resembles the VAE structure depicted in Figure~\ref{fig:vae_nn}. The discriminator shown in Figure~\ref{fig:gan_nn} (a) outputs the decision for the input if it is real or fake. The generator, shown in Figure~\ref{fig:gan_nn} (b) outputs the image from the $D$-dimensional noise vector. Here, $D=64$ is applied. This similarity of the neural network structure between GAN and VAE enables us to qualitatively compare the performance of these two models. The models discussed in subsequent sections, such as Normalizing Flows, Score-based Generative Models, and Diffusion Models, possess more intricate and distinctive structures. Consequently, evaluating these models within the same neural network framework is impractical. However, the architectures of the VAE and GAN can be designed to be similar, which fulfills our purpose of comparison.

\subsection{Normalizing Flows (Flow-based Generative Models)}\label{sec:F.3}

As we discussed in Section~\ref{sec:nf}, among various Flow-based Generative Models such as \cite{dinh2014nice,rezende2015variational,kingma2018glow,durkan2019neural}, we implemented RealNVP \citep{dinh2016density} because of its ability to generate high-dimensional distribution due to its flexibility. As we can see in Equation~\eqref{eqn:nf_2}, we need to calculate the $\log p_{0} (z_{0})$ in Equation~\eqref{eqn:nf_1} to get the likelihood of the data. In the implementation of RealNVP, we can simplify the log-likelihood $\log \left(p(\mathbf{z})\right)$ by assuming $z$ is sampled from the standard normal distribution. The other term in Equation~\eqref{eqn:nf_1}, the summation of the determinant of the Jacobian, is simplified by applying a coupling layer as described through Equation~\eqref{eqn:affine} to Equation~\eqref{eqn:nf_7}. Thus, the summation of the determinant of the Jacobian can be simplified as $\sum_j s(x_{1:d})_j$. In our code application, the regularization loss for batch normalization and scale parameters are applied following the contents in RealNVP \citep{dinh2016density}.

\cite{dinh2016density} multiplied $5\times10^{-5}$ for the regularization term of scale network. However, since the coefficients must be re-calibrated based on the data, we fine-tuned each coefficient for the loss term in our implementation. A general guideline is to prioritize log-likelihood over the sum of the determinant of the Jacobian. Additionally, the sum of the batch normalization also plays a critical role in the qualitative performance, which has been incorporated into the provided code.




\FloatBarrier

\begin{figure}
    \centering
    \includegraphics[width=0.70\textwidth]{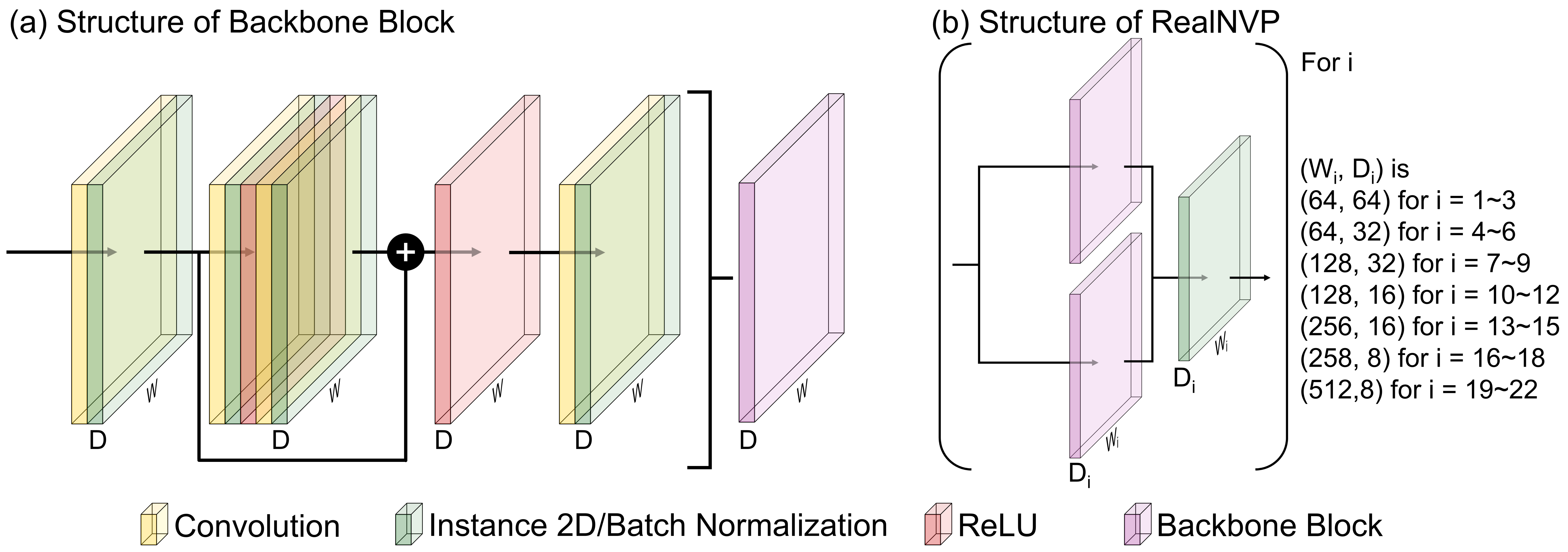}
    \caption{Neural network structure of RealNVP in highway traffic speed contour}
    \label{fig:realnvp_nn}
\end{figure}

The neural network structure of RealNVP is shown in Figure~\ref{fig:realnvp_nn}. The neural network is composed of multiple layers of its Backbone Block, which is shown in Figure~\ref{fig:realnvp_nn} (a). Each Backbone Block consists of convolution, Instance 2D Normalization, and ReLU layer. Each Backbone Block is the smallest unit for the scale and translation network. In the implementation in the current paper, we stacked 22 layers of scale and translation network. The dimensions and number of channels of the features change at each layer, as detailed in Figure~\ref{fig:realnvp_nn} (b). In the generation stage, the computation proceeds in an inverse way.

\subsection{Score-based Generative Models}\label{sec:F.5}
In the code implementation of NCSN in traffic speed contour, we had minor adjustments from the code from Section~\ref{sec:E.5}. By definition, $\tilde{\mathbf{x}}$ is defined as $\tilde{\mathbf{x}}=\mathbf{x}+\sigma \cdot \epsilon$. Then $\nabla_{\mathbf{x}} \log q_\sigma \left( \tilde{\mathbf{x}} | \mathbf{x} \right)$ in Equation~\eqref{eqn:ncsn_der1} is transformed as $\nicefrac{\epsilon}{\sigma}$. Practically, $\sigma^2$ is multiplied by the objective function as the weight of each stage. The detailed implementation can be found in the code.



\begin{figure}[t]
    \centering
    \includegraphics[width=0.80\textwidth]{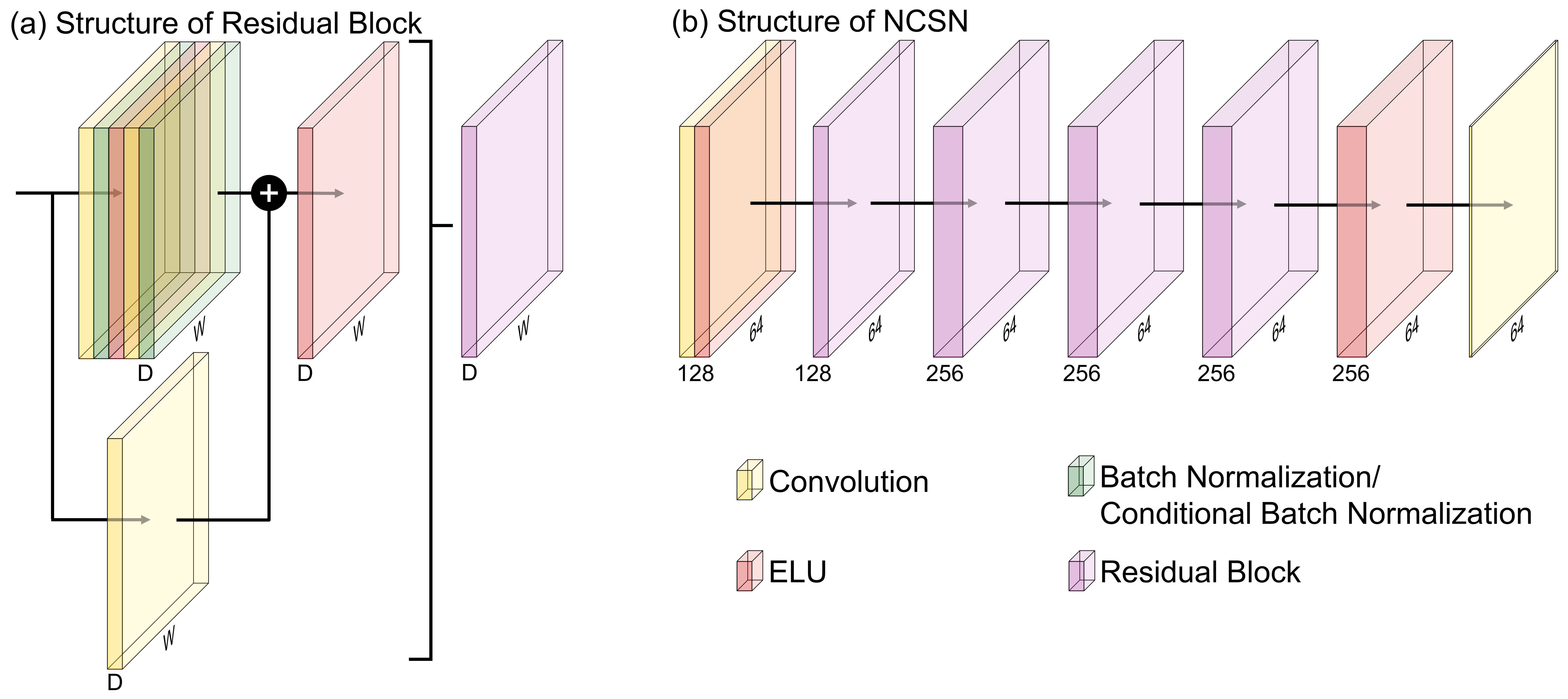}
    \caption{Neural network structure of NCSN in highway traffic speed contour}
    \label{fig:ncsn_nn}
\end{figure}

The neural network of NCSN is formed with a block-wise structure with a Residual Block. Figure~\ref{fig:ncsn_nn} (a) and Figure~\ref{fig:ncsn_nn} (b) illustrate the structure of the Residual Block and whole neural network for score function, respectively. Skip-connection is applied in each block to maintain the low-level feature information.

\subsection{Diffusion Models}\label{sec:F.4}
The loss function of DDPM is simplified as the difference between the predicted noise and the real noise as stated in Equation~\eqref{eqn:diff_1}. The criteria of the difference, i.e., the norm of difference, can be varied by the problem. Our code is also designed to select one of three types of loss, which is $L_1$, MSE loss, and Huber loss. For simplicity, we inserted the code block only including the MSE loss.


\begin{figure}
    \centering
    \includegraphics[width=0.70\linewidth]{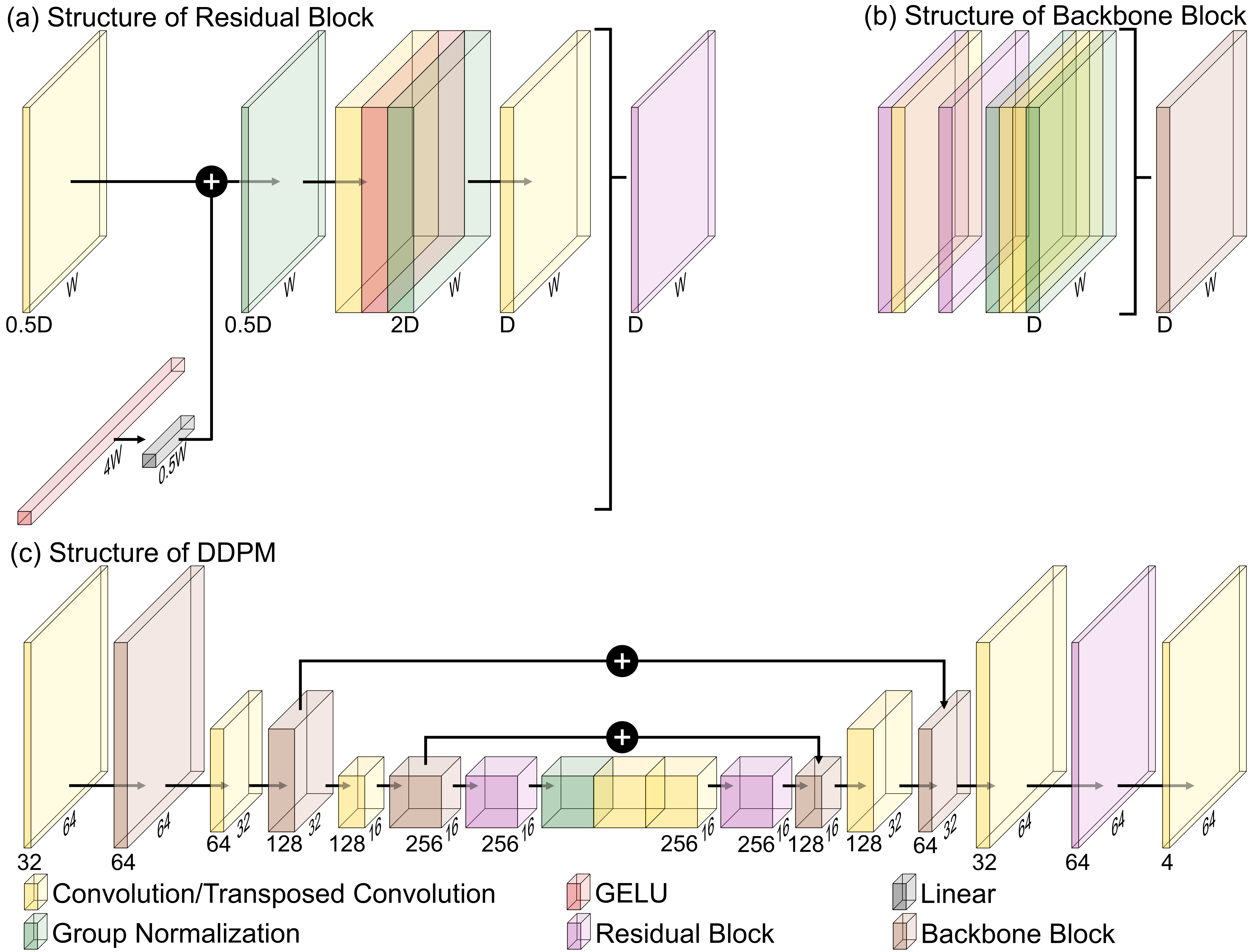}
    \caption{Neural network structure of DDPM in highway traffic speed contour}
    \label{fig:ddpm_nn}
\end{figure}

The neural network that is used for the DDPM is illustrated in Figure~\ref{fig:ddpm_nn}. Figure~\ref{fig:ddpm_nn} (a), Figure~\ref{fig:ddpm_nn} (b), and Figure~\ref{fig:ddpm_nn} (c) show the structure of the Residual Block, Backbone Block, and U-Net of DDPM, respectively. The U-Net structure that is applied in the neural network in DDPM enables one to learn the information from high-level features or features from previous layers. The residual block contains the time embedding as an input, which identifies the stage of the current generation process. A similar mechanism can also be found in the Score-based model, as the score-based model creates an image using the scheduled $\sigma$. The neural network in DDPM also includes the attention mechanism that captures the relationship between features. The attention mechanism was applied inside the Backbone block, between the Residual block and the group normalization.

\FloatBarrier

\section{Smoothing Process of Highway Traffic Speed Contour}\label{appendix:smoothing}

The detailed process of interpolation, which is based on Edie's definition of traffic flow dynamics, is outlined in Table~\ref{tab:table2}.

\begin{center}
\captionof{table}{Pseudocode for data preprocessing}\label{tab:table2}
\begin{tabular}{p{0.5cm}p{0.5cm}p{0.5cm}p{12.5cm}}
\hline
\multicolumn{4}{l}{Algorithm 1: Smoothing the speed using the Edie's definition} \\ \hline
1     & \multicolumn{3}{l}{\textbf{Given}}                                                         \\
2     & \multicolumn{3}{l}{$V(x, t)$: Velocity at $x, t$}                                          \\
3     & \multicolumn{3}{l}{$W, T$: Spatial and temporal width of the data}                         \\
4     & \multicolumn{3}{l}{$\sigma$: Spatial range of smoothing}                                   \\
5     & \multicolumn{3}{l}{$\tau$: Temporal range of smoothing}                                    \\
6     & \multicolumn{3}{l}{$c_{free}$: Propagation velocity of perturbation in free flow}          \\
7     & \multicolumn{3}{l}{$c_{cong}$: Propagation velocity of perturbation in congested flow}     \\
8     & \multicolumn{3}{l}{$V_c$: Velocity that transition occurs from free flow to congested flow}\\
9     & \multicolumn{3}{l}{$\Delta V$: Width of the transition region}                             \\
10    & \multicolumn{3}{l}{\textbf{for} $x$←$0$ to $W$, $t$←$0$ to $T$ \textbf{do}}                \\
11    &       & \multicolumn{2}{l}{\textbf{Initialize} $\phi_{free}(x', t'), \phi_{cong}(x', t')$} \\
12    &       & \multicolumn{2}{l}{\textbf{For} $x'$←$x-\nicefrac{\sigma}{2}$ to $x+\nicefrac{\sigma}{2}$, $t'$←$t-\nicefrac{\tau}{2}$ to $t+\nicefrac{\tau}{2}$ \textbf{do}}\\
13    &       &           & $\Delta x = x'-x$                                                      \\
14    &       &           & $\Delta t_{free} = (t'-t)-\nicefrac{(x'-x)}{c_{free}}$                 \\
15    &       &           & $\Delta t_{cong} = (t'-t)-\nicefrac{(x'-x)}{c_{cong}}$                 \\
16    &       &           & $\phi_{free}(x', t') = \exp(\nicefrac{|\Delta x|}{\sigma}-\nicefrac{|\Delta t_{free}|}{\tau})$ \\
17    &       &           & $\phi_{cong}(x', t') = \exp(\nicefrac{|\Delta x|}{\sigma}-\nicefrac{|\Delta t_{cong}|}{\tau})$ \\
18    &       & \multicolumn{2}{l}{\textbf{End for}}                                               \\
19    &       & \multicolumn{2}{l}{\textbf{Initialize} $V_{free}(x, t), V_{free}(x, t)$}           \\
20    &       & \multicolumn{2}{l}{$V_{free} (x, t)=\sum_{x'=x-\nicefrac{\sigma}{2}}^{x+\nicefrac{\sigma}{2}}\Bigl(\phi_{free}(x', t') \times V(x', t')\Bigr) \Big/\sum_{x'=x-\nicefrac{\sigma}{2}}^{x+\nicefrac{\sigma}{2}}\phi_{free}(x', t')$} \\
21    &       & \multicolumn{2}{l}{$V_{cong} (x, t)=\sum_{x'=x-\nicefrac{\sigma}{2}}^{x+\nicefrac{\sigma}{2}}\Bigl(\phi_{cong}(x', t') \times V(x', t')\Bigr) \Big/\sum_{x'=x-\nicefrac{\sigma}{2}}^{x+\nicefrac{\sigma}{2}}\phi_{cong}(x', t')$} \\
22    &       & \multicolumn{2}{l}{$V_{min}(x, t) = \min\Bigl(V_{free}(x, t), V_{cong}(x, t)\Bigr)$}       \\
23    &       & \multicolumn{2}{l}{$tanh(x, t)=\tanh\Bigl(\nicefrac{(V_c - V_{min}(x, t))}{\Delta V}\Bigr)$} \\
24    &       & \multicolumn{2}{l}{$w=0.5 \times \Bigl(1+\tanh(x, t)\Bigr)$} \\
25    &       & \multicolumn{2}{l}{$V(x, t)=(1-w) \times V_{free}(x, t) + w \times V_{cong}(x, t)$}\\
26    & \multicolumn{3}{l}{\textbf{End for}}         \\ \hline
\end{tabular}
\end{center}



\end{document}